%% file: main_content_ARR.tex
\pdfoutput=1

\documentclass[11pt]{article}

\usepackage[final]{acl}

\usepackage{times}
\usepackage{latexsym}

\usepackage[T1]{fontenc}

\usepackage[utf8]{inputenc}

\usepackage{microtype}

\usepackage{inconsolata}

\usepackage{graphicx}
\usepackage{hyperref}
\usepackage{booktabs}
\usepackage{multirow}
\usepackage{caption}
\usepackage{subcaption}
\usepackage[normalem]{ulem}
\useunder{\uline}{\ul}{}
\usepackage[most]{tcolorbox}
\newtcolorbox{RQBox}[2][]{colback=gray!20!white,colframe=gray!80!black,
                          fonttitle=\bfseries, title=#2,#1}
\input{math_commons}

\newcommand{\olmo}{OLMo-7b-Instruct\xspace}
\newcommand{\bloom}{BLOOM-7b1\xspace}
\newcommand{\llamaTwo}{Llama2-7b-chat\xspace}
\newcommand{\llamaTwoX}{Llama2-13b-chat\xspace}
\newcommand{\llamaTwoXX}{Llama2-70b-chat\xspace}
\newcommand{\llamaThree}{Llama3-8b-Instruct\xspace}
\newcommand{\llamaThreeXX}{Llama3-70b-Instruct\xspace}

\newcommand{\gptThree}{GPT-3\xspace}
\newcommand{\gptThreeFive}{GPT-3.5-turbo\xspace}
\newcommand{\gptFour}{GPT-4\xspace}
\newcommand{\gptFouro}{GPT-4o\xspace}

\newcommand{\dolma}{Dolma\xspace}
\newcommand{\roots}{ROOTS\xspace}
\newcommand{\conceptnet}{ConceptNet\xspace}
\newcommand{\causenet}{CauseNet\xspace}

\title{On the Reliability of Large Language Models for Causal Discovery}

\author{
 \textbf{Tao Feng\textsuperscript{1}},
 \textbf{Lizhen Qu\textsuperscript{1} \thanks{Corresponding author}},
 \textbf{Niket Tandon\textsuperscript{2}},
 \textbf{Zhuang Li\textsuperscript{1}},
\\
 \textbf{Xiaoxi Kang\textsuperscript{1}},
 \textbf{Gholamreza Haffari\textsuperscript{1}}
\\
 \textsuperscript{1}Monash University,
 \textsuperscript{2}Microsoft Research, India
\\
 \small{
   \textsuperscript{1} {{firstname.lastname}@monash.edu},
   \textsuperscript{2} {nikett@gmail.com}
 }
}

\begin{document}
\maketitle
\begin{abstract}
This study investigates the efficacy of Large Language Models (LLMs) in causal discovery. Using newly available open-source LLMs, \textsc{OLMo} and \textsc{BLOOM}, which provide access to their pre-training corpora, we investigate how LLMs address causal discovery through three research questions. We examine: (i) the impact of memorization for accurate causal relation prediction, (ii) the influence of incorrect causal relations in pre-training data, and (iii) the contextual nuances that influence LLMs' understanding of causal relations. Our findings indicate that while LLMs are effective in recognizing causal relations that occur frequently in pre-training data, their ability to generalize to new or rare causal relations is limited. Moreover, the presence of incorrect causal relations significantly undermines the confidence of LLMs in corresponding correct causal relations, and the contextual information critically affects the outcomes of LLMs to discern causal connections between random variables~\footnote{The code and data are available at \url{https://github.com/WilliamsToTo/causality_llm}}. 
\end{abstract}

\section{Introduction}
\label{sec:intro}
\input{sections/intro_new}

\section{Background}
\input{sections/background}

\section{Methodology}
\input{sections/methodology}

\section{Experimental Setup}
\input{sections/experimental_design}

\section{Experimental Results}

\input{sections/experimental_result_new}

\section{Related Work}
\input{sections/related_work}

\section{Conclusion}
\input{sections/conclusion}

\section*{Limitations}
\input{sections/limitations}

\section*{Ethics Statement}
\input{sections/ethics_statement}

\bibliography{custom}

\appendix

\section{Appendix}
\label{sec:appendix}
\input{sections/appendix}

\end{document}

%% file: math_commons.tex
\usepackage{amsthm,amsmath,amsfonts,bm,xspace}
\usepackage{color}

\usepackage[english]{babel}

\newtheorem{RQ}{Research Question}

\def\eg{{\em e.g.,}\xspace}
\def\ie{{\em i.e.,}\xspace}

\definecolor{mgreen}{rgb}{0,0.7,0}

\def\eqref#1{(\ref{#1})}

\def\1{\bm{1}}

\DeclareMathAlphabet{\mathsfit}{\encodingdefault}{\sfdefault}{m}{sl}
\SetMathAlphabet{\mathsfit}{bold}{\encodingdefault}{\sfdefault}{bx}{n}



%% file: sections/intro_new.tex
Identification and understanding of causal relations hold fundamental importance in human cognition and science, as those relations form the basis of causal models, which are utilized to answer observational, interventional, and counterfactual questions~\citep{zanga2022survey, wan2024bridging}. The task of identifying causal relations among a set of random variables is known as \textit{causal discovery}, where a random variable may refer to an event in daily life, a medical treatment, or a drug effect, etc.~\citep{pearl2009causality, 10.5555/3202377, nogueira2021causalDiscovery}. For decades, various statistical methods have been developed to identify causal relations from observational or interventional data \cite{heckerman1995learning, chickering2002optimal, 10.5555/2946645.2946677}. However, algorithms that can accurately recover true causal structures from observational data remain elusive \cite{neal2020causalitybook}.

With the rise of Large Language Models (LLMs), recent studies exploit the potential of LLMs for causal discovery by evaluating them on benchmark datasets~\cite{willig2022probing,ban2023query}. Proprietary LLMs, such as \gptThree and \gptFour, surpass the state-of-the-art (SOTA) statistical methods on several publicly available datasets~\citep{kıcıman2023causal}. However, \citet{romanou2023crab} notice both \gptThree and \gptFour have a performance drop on the causal relations involving real-world events occurring post-Jan 2022, compared to the ones before Jan 2022. \citet{kıcıman2023causal} find out that given part of a data table in the T{\"u}bingen cause-effect pairs dataset~\citep{mooij2016distinguishing}, \gptFour can recover 61\% of the remaining part. \citet{ze2023causal} conjecture that \textit{LLMs may just recall causal knowledge in their large pre-training corpora by acting as "causal parrots"}. However, there are no solid experiments to investigate in which cases LLMs' predictions are \textit{reliable}. Can we trust the predictive outcomes only if they are based on what LLMs memorize in the training data?  

Therefore, we conduct experiments to investigate the cases that LLMs make correct and wrong predictions respectively, leading to the following three research questions. \textit{RQ~\ref{rq:memorization_generalization}:  Under what conditions do LLMs \textit{reliably} and \textit{consistently} make \textit{accurate} predictions in causal discovery?} \textit{RQ~\ref{rq:anti_causal_occur}: How does the presence of incorrect causal relations affect LLMs' performance in causal discovery?} and \textit{RQ~\ref{rq:causal_context}: How does the contextual information of a causal relation influence LLMs' performance in causal discovery?}. To understand the effect of memorization, we employ the recently released open-source LLMs \textsc{OLMo} and \textsc{BLOOM}, which make their respective pre-training corpora \dolma and \roots publicly available \cite{groeneveld2024olmo, workshop2023bloom}. This provides the opportunity for us to investigate the correlations between model outputs and the frequency of relations mentioned in their pre-training corpora.

Our experiments reveal the following findings.
\begin{itemize}
    \item Although the evaluated LLMs are proficient at recognizing causal relations through memorization, their ability to generalize novel causal relations is limited. This limitation poses significant challenges for deploying LLM-based causal discovery methods in scenarios where causal relations are rarely or not included in their pre-training data.
    
    \item The presence of incorrect causal relations, such as the reversal of correct causal relations, adversely impacts LLMs' confidence in identifying correct causal relations. This finding highlights the necessity of minimizing conflicting causal information in pre-training datasets to enhance the performance of LLMs.

    \item The validity and strength of causal relations can vary significantly across different contexts. This variability suggests that LLM-based causal discovery methods should incorporate the context of causal relations as input to ensure accuracy, particularly to avoid misleading contexts that could substantially degrade performance.

\end{itemize}

%% file: sections/background.tex

Causal discovery aims to identify causal relations among a given set of random variables. For each pair of variables $X$ and $Y$, it identifies whether $X \leftarrow Y$, $Y \leftarrow X$, or there is no causal influence between them, where $\leftarrow$ denotes the direction of causality. 
The traditional algorithms for this task are statistical methods that perform causal discovery on tabular data, which are capable of unveiling previously unknown or uncertain causal relations that are not \textit{explicitly} mentioned anywhere in text (\eg "sea level pressure causally influences zonal wind at 10 m" \cite{huang2021benchmarking}). 
In contrast, prior NLP methods focus on either extracting mentions of known causal relations from documents~\cite{yang2022survey} or answering questions related to causality~\cite{oh2013question}. The gold standard for causal discovery is experimental approaches such as randomized controlled trials and A/B testing \cite{fisher:1935}. However, such experiments are often not feasible due to ethical or financial constraints, which necessitates the use of alternative methods that rely solely on statistics collected from observational data. 

The statistical causal discovery methods can be categorized into: constraint-based methods, such as Peter and Clark (PC) \cite{spirtes2000causation} and inductive causation (IC) \cite{pearl2009causality}; score-based methods \cite{heckerman1995learning, chickering2002optimal, koivisto2004exact, 10.5555/2946645.2946677}; and functional methods \cite{shimizu06a, hyvarinen10a}. Those methods rely on statistics calculated from tabular data to infer causal graphs, in which random variables are depicted as nodes and their causal relations are represented as edges. However, a significant drawback of these approaches is their dependency on extensive data collection to construct reliable tabular data, a process that can be both time-consuming and costly. Furthermore, a theoretical limitation of these statistical methods is their inability to precisely predict ground-truth causal graphs, unless strong assumptions are made. Instead, they typically yield an equivalence class of true causal graphs \cite{spirtes2000causation, pearl2009causality}. 

Recent advances of LLMs provide new opportunities to tackle the task without accessing tabular data by formulating it as a pairwise causal relation prediction task~\cite{kıcıman2023causal, ze2023causal, long2022can, zhang2023understanding}. Given a pair of variable names, an LLM is instructed to identify which is the cause and which is the effect using prompts \cite{kıcıman2023causal,ze2023causal}, by distilling such knowledge directly from the LLM. However, the reliability of such methods is under scrutiny. \citet{ze2023causal} argue that LLMs are \textit{"causal parrots"}, which may depend on \textit{memorization} to recall the causal relations present in their training data. In other words, LLMs may not \textit{generalize} well to detect causal relations that seldom or never occur in pre-training data. If this argument holds, LLMs may primarily excel at reproducing causal relations known in their training data rather than uncovering novel ones. However, there is no solid empirical justification of this argument because prior works employ either commercial LLMs or open-source LLMs that have no access to their training data. The current techniques for understanding and investigating memorization in LLMs are still in their infancy~\cite{speicher2024understanding}.

%% file: sections/methodology.tex

We empirically investigate the reliability of LLMs in causal discovery by addressing the three research questions introduced in Section \ref{sec:intro}. Our methodology starts with examining the evidence within pre-training data that supports accurate LLM predictions. We then systematically identify potential sources of prediction errors. Given the highly contextualized nature of LLM outputs, we thoroughly analyze how contextual variations influence their predictive performance in causal discovery.

\paragraph{RQ1.} This question aims to collect strong empirical evidence to verify the "causal parrots" hypothesis and investigate the cases, in which LLMs make accurate predictions. The "causal parrots" hypothesis states that LLMs predict correct causal relations just because they are explicitly mentioned in the training data. If the hypothesis is true, LLMs would only be suited to reproduce known causal relations and not infer new ones. However, prior studies on LLMs for causal discovery fail to provide solid empirical evidence in pre-training data.



To address this, we design experiments with \olmo and \bloom, which release their pre-training data \cite{groeneveld2024olmo, workshop2023bloom}, on both real-world and synthetic datasets. For real-world data, we collect mentions of given causal relations from pre-training data using causal relation templates, which contain keywords that indicate causal relations, such as "cause" and "lead to". We then compute the correlation between mention occurrence and LLMs' predictive accuracy. \textit{If an LLM relies solely on explicit linguistic cues to predict causal relations, we would expect a high correlation between mention occurrence and prediction accuracy.} To compute correlations, we stratify the occurrence range into intervals, ensuring that each interval contains a roughly equal number of relations. We then systematically evaluate LLMs' performance in each of these predefined occurrence intervals. Our assessment methodology involves transforming causal relations into yes-no questions, such as "does smoking cause lung cancer?", and measure the performance in terms of accuracy and F1-score. This experimental setup follows the approaches stated in \citet{razeghi-etal-2022-impact}.

As it is almost infeasible to collect all mentions of a causal relation from a dataset, we curate a synthetic causal relation dataset to further investigate RQ1 in a controlled environment. Herein, we use variables that do not exist in any pre-training data and fill them into our relation templates to curate the corresponding mentions with varying frequencies, such as "blaonge causes goloneke,". To simulate real-world data, we insert them into a random collection of documents as the synthetic training data for LLMs.


\paragraph{RQ2.} We conjecture that incorrect predictions primarily stem from the presence of semantically opposing or negating causal relations in the training data. Given a relation e.g. "smoking causes lung cancer.", we examine the extent to which the mention occurrence of "lung cancer causes smoking." or "smoking does not cause lung cancer." influences LLMs' predictive performance. To this end, we assess the confidence of LLMs in correct causal relations under varying frequencies of corresponding incorrect causal relations on both \textit{real-world data and synthetic data}. We hypothesize that a higher presence of incorrect causal relations diminishes the LLMs' confidence in correct causal relations. The confidence level of the LLMs is measured by the proportion of responses that affirm the correct causal relation out of a sample of generated responses for one query. Following the same procedure as RQ1, we create the synthetic dataset by inserting incorrect predictions with varying frequencies into the same document collection. 


\paragraph{RQ3.} While almost all statistical methods assume that causal graphs stay the same regardless their contexts, we observe that LLMs' predictions in causal discovery vary across different contexts. There is no quantitative study investigating how contexts influence LLMs' outcomes in causal discovery. However, in real-world scenarios, a causal relation is present only in certain contexts. 
For example, the causal relation "rain causes flooding" may be true during a heavy downpour in a city with poor drainage but may not be true during light rain in areas with good drainage systems. Therefore, we assess the performance of LLMs in various contexts. For each given causal relation from human-annotated datasets, we employ \gptFouro to generate five positive contexts that affirm the relation and five negative contexts. Both LLMs are instructed to provide the answers of the corresponding yes-no questions in those contexts. This study yields quantitative results to demonstrate the importance of context, which may serve as another source of prediction errors.

%% file: sections/experimental_design.tex

\begin{figure*}[ht]
    \centering
    \begin{subfigure}[t]{0.23\textwidth} 
        \includegraphics[width=\textwidth]{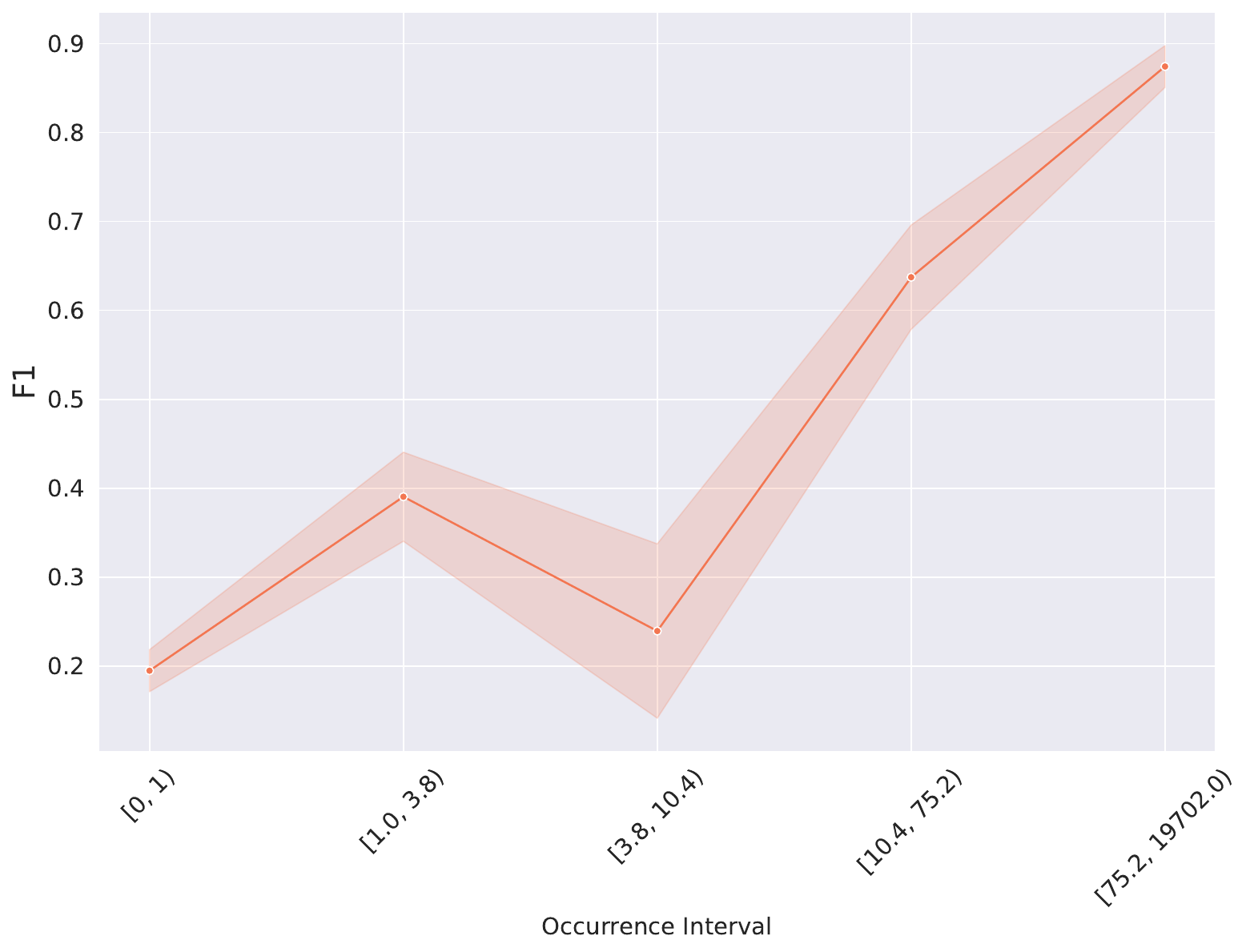} 
        \caption{F1 vs Occurrence; Pearson (r=0.80*); Spearman (r=0.9*)}
        \label{fig:occurrence_vs_f1_causal_discovery_dolma_exact_match}
    \end{subfigure}
    \hspace{0.01\textwidth}
    \begin{subfigure}[t]{0.23\textwidth} 
        \includegraphics[width=\textwidth]{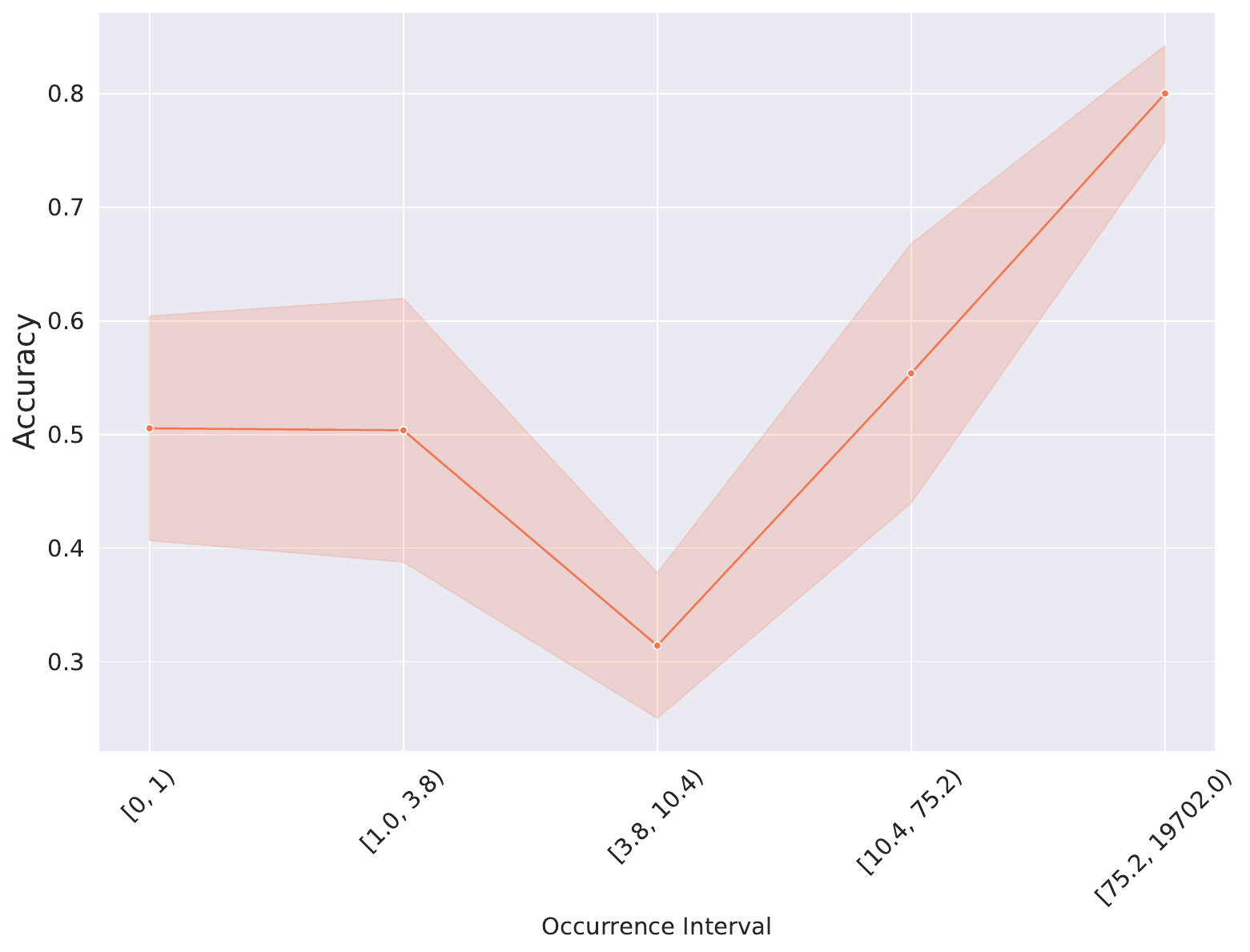} 
        \caption{Accuracy vs Occurrence; Pearson (r=0.85*); Spearman(r=0.6)}
        \label{fig:occurrence_vs_accuracy_causal_discovery_dolma_exact_match}
    \end{subfigure}
    \hspace{0.01\textwidth}
    \begin{subfigure}[t]{0.23\textwidth} 
        \includegraphics[width=\textwidth]{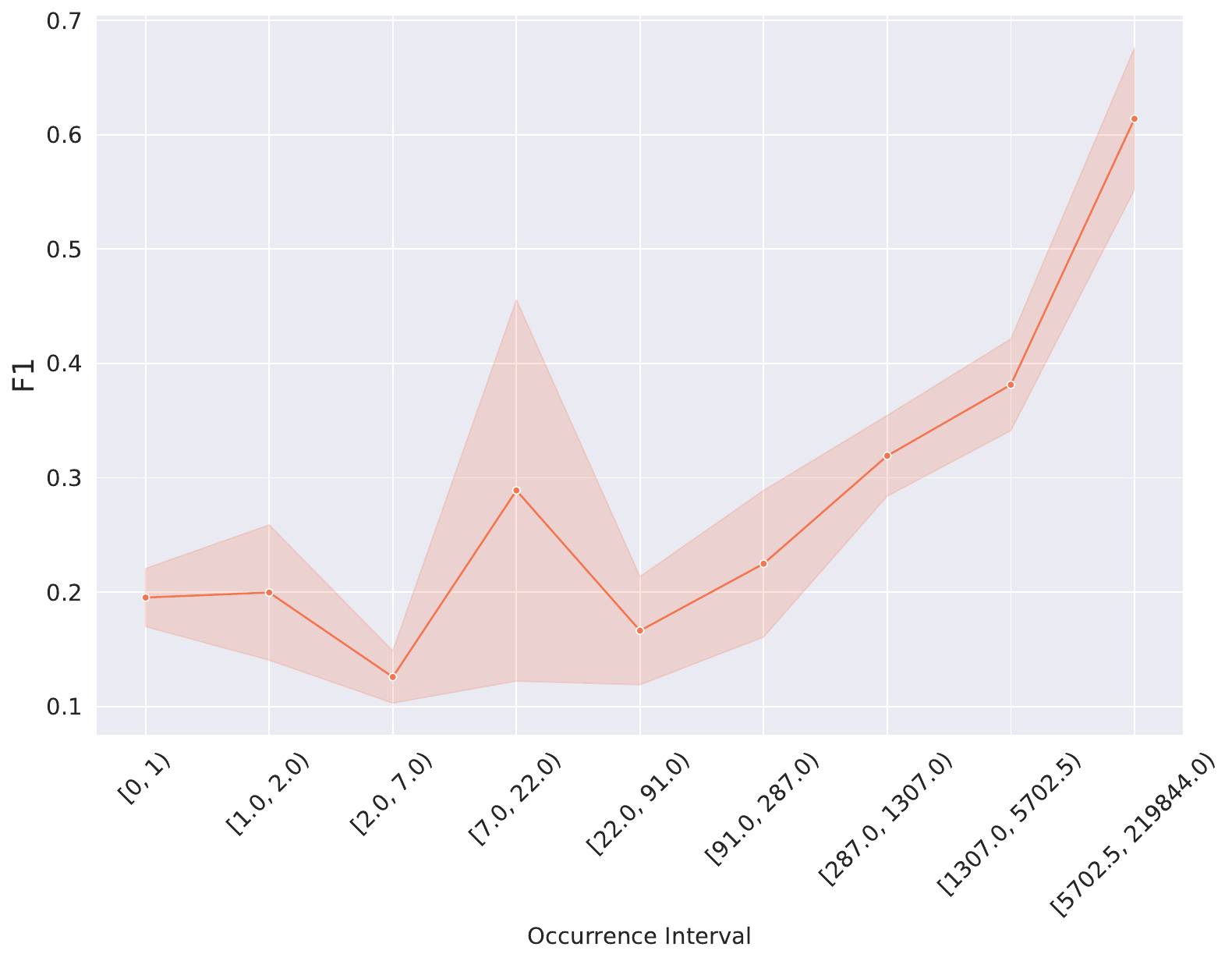} 
        \caption{F1 vs Occurrence; Pearson (r=0.86*); Spearman (r=0.78*)}
        \label{fig:occurrence_vs_f1_causal_discovery_dolma_causal_co_occur}
    \end{subfigure}
    \hspace{0.01\textwidth}
    \begin{subfigure}[t]{0.23\textwidth} 
        \includegraphics[width=\textwidth]{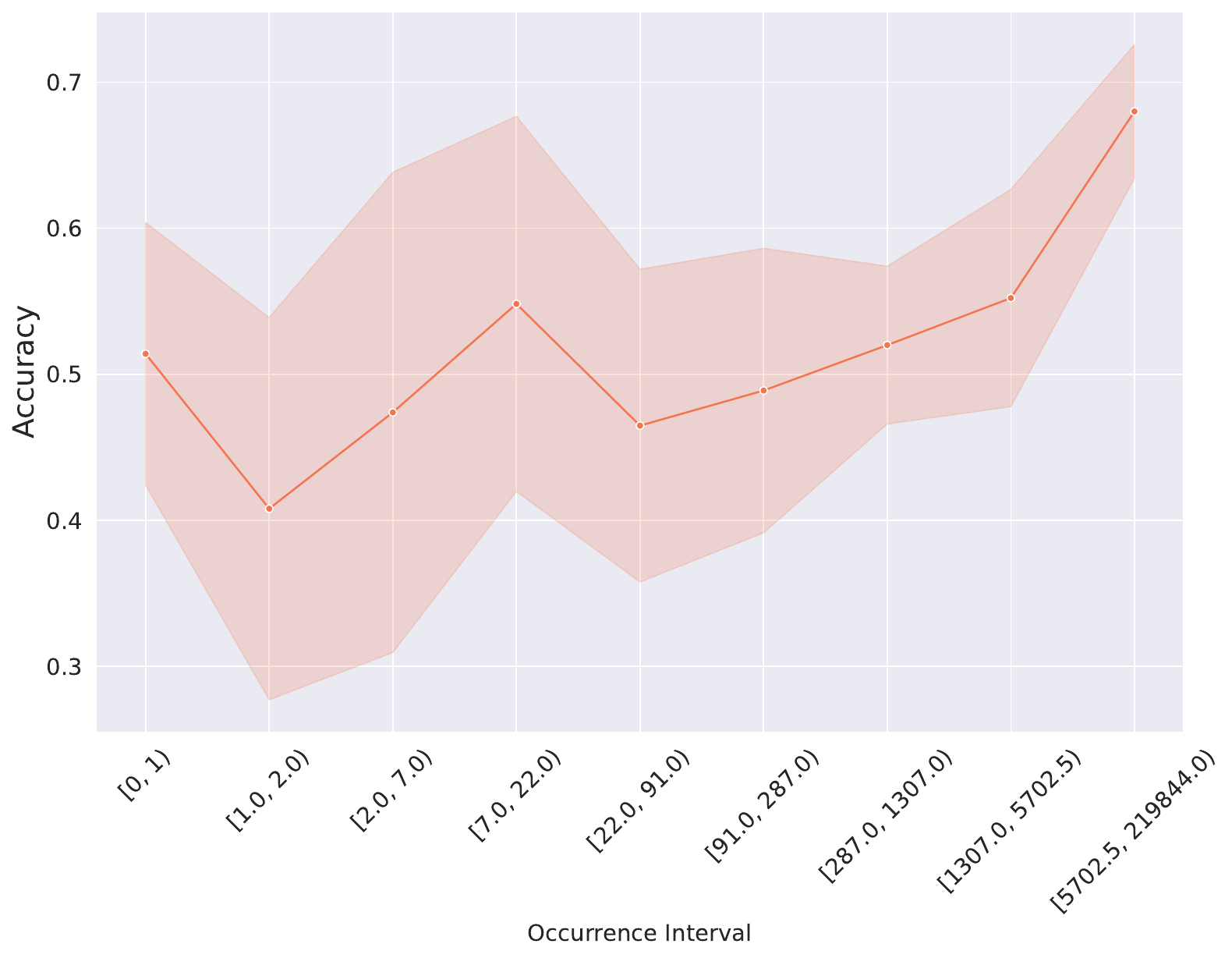} 
        \caption{Accuracy vs Occurrence; Pearson (r=0.82*); Spearman (r=0.67*)}
        \label{fig:occurrence_vs_accuracy_causal_discovery_dolma_causal_co_occur}
    \end{subfigure}
    \caption{The average F1 score and accuracy of \olmo by occurrence interval on full causal discovery tasks, where F1 and accuracy are computed from 0 to 4 ICL examples. The occurrence data of (a) and (b) are derived from the exact matching query, while the occurrence data of (c) and (d) are derived from the "event A" $\Rightarrow$ "causes" $\Rightarrow$ "event B" query. An asterisk (*) indicates a p-value < 0.05 for Pearson and Spearman correlation coefficients \cite{freedman2007statistics}.}
    \label{fig:causal_discovery_dolma_frequency}
\end{figure*}

\begin{figure}[ht]
    \centering
    \begin{subfigure}[t]{0.23\textwidth} 
        \includegraphics[width=\textwidth]{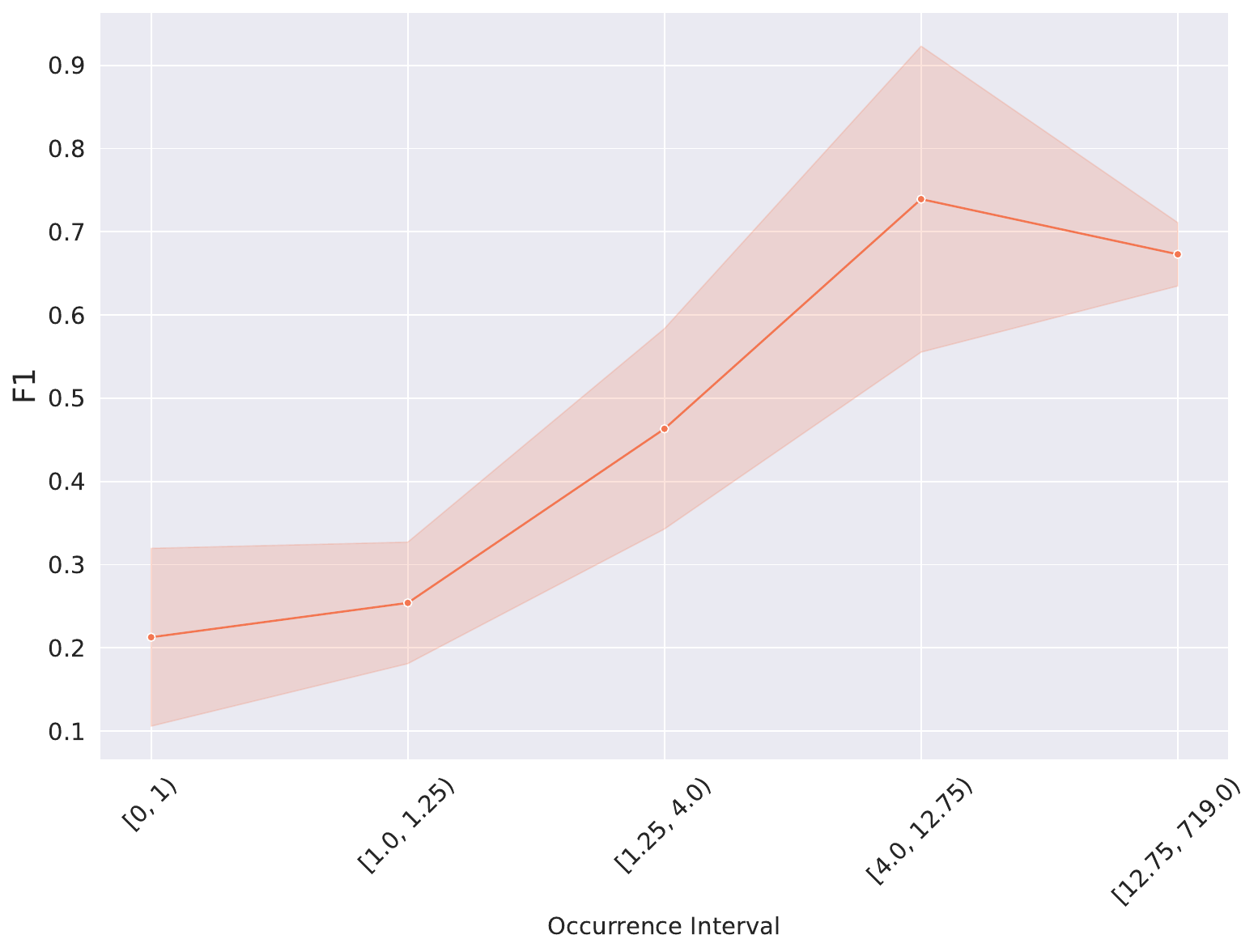} 
        \caption{F1 vs Occurrence; Pearson (r=0.5); Spearman (r=0.9*)}
        \label{fig:occurrence_vs_F1_bloom_xshot_causal_discovery_roots_causal_match_mean_std}
    \end{subfigure}
    \hspace{0.01\textwidth}
    \begin{subfigure}[t]{0.23\textwidth} 
        \includegraphics[width=\textwidth]{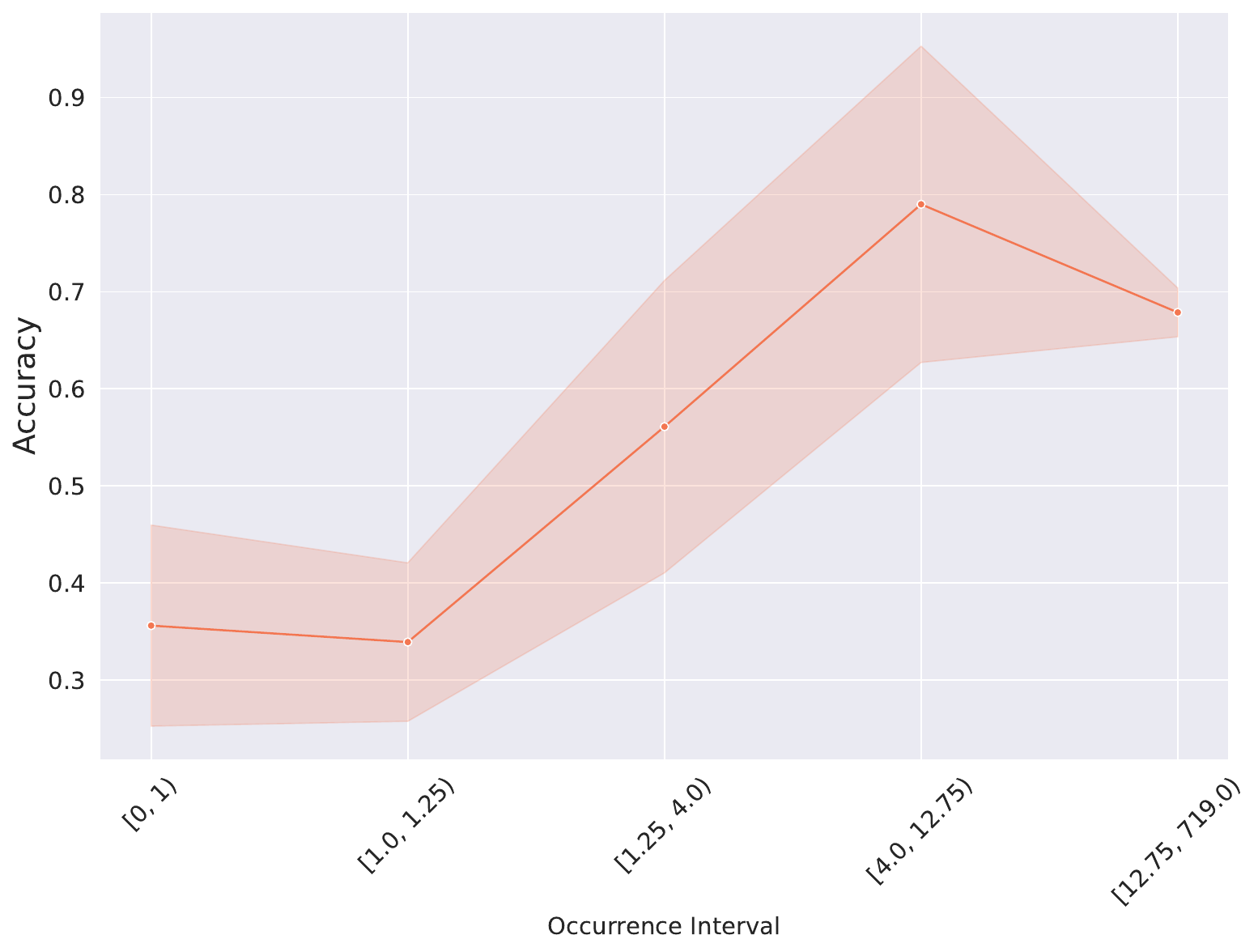} 
        \caption{Accuracy vs Occurrence; Pearson (r=0.4); Spearman(r=0.8)}
        \label{fig:occurrence_vs_Accuracy_bloom_xshot_causal_discovery_roots_causal_match_mean_std}
    \end{subfigure}
    \caption{The average F1 score and accuracy of \bloom by occurrence interval on full causal discovery, averaged 0-4 ICL examples. The occurrence data are derived from the exact matching query.}
    \label{fig:causal_discovery_roots_frequency}
\end{figure}

\begin{figure*}[ht]
    \centering
    \begin{subfigure}[t]{0.23\textwidth} 
        \includegraphics[width=\textwidth]{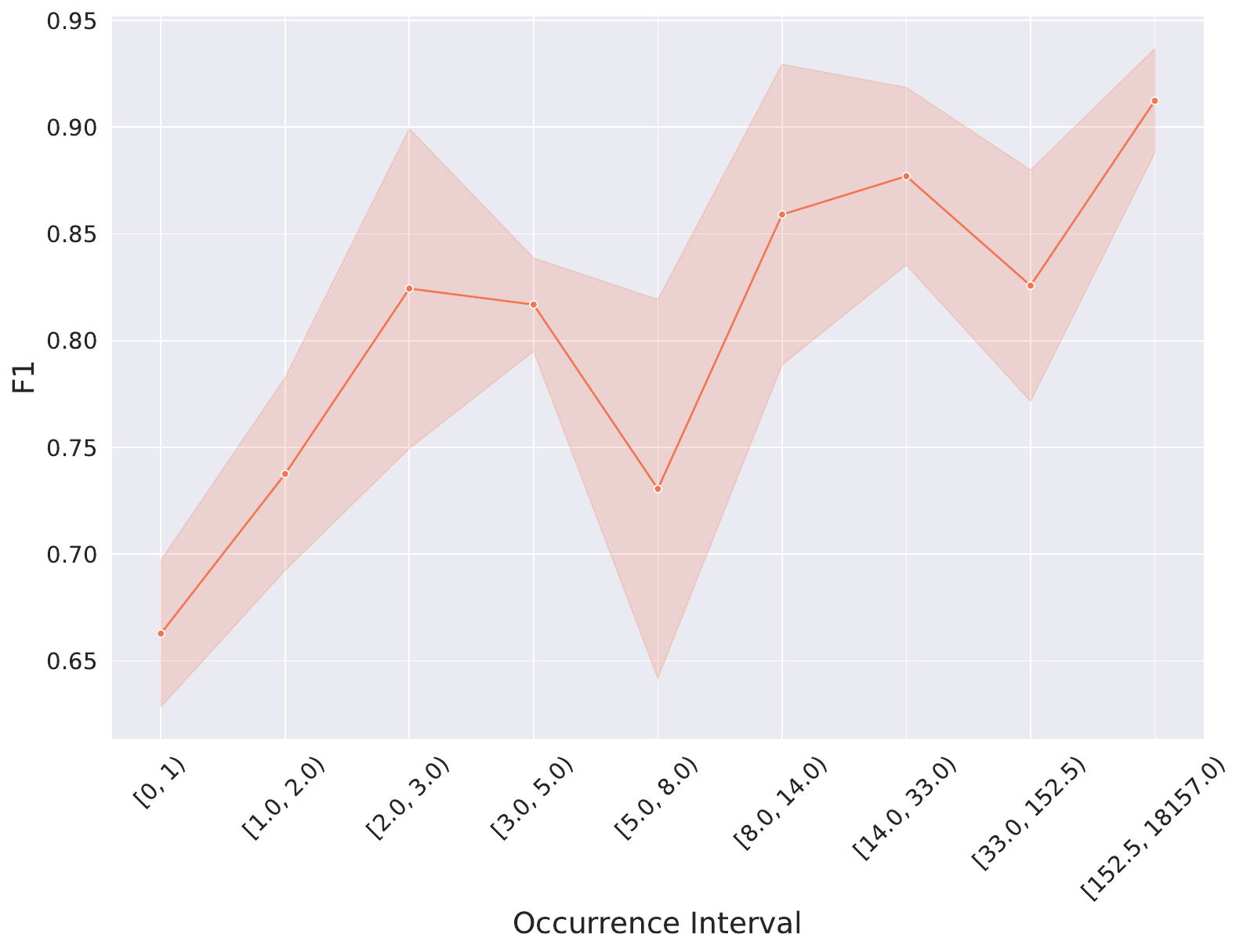} 
        \caption{F1 vs Occurrence at \conceptnet; Pearson (r=0.51); Spearman (r=0.83*)}
        \label{fig:occurrence_vs_f1_dolma_conceptnet_exact_match}
    \end{subfigure}
    \hspace{0.01\textwidth}
    \begin{subfigure}[t]{0.23\textwidth} 
        \includegraphics[width=\textwidth]{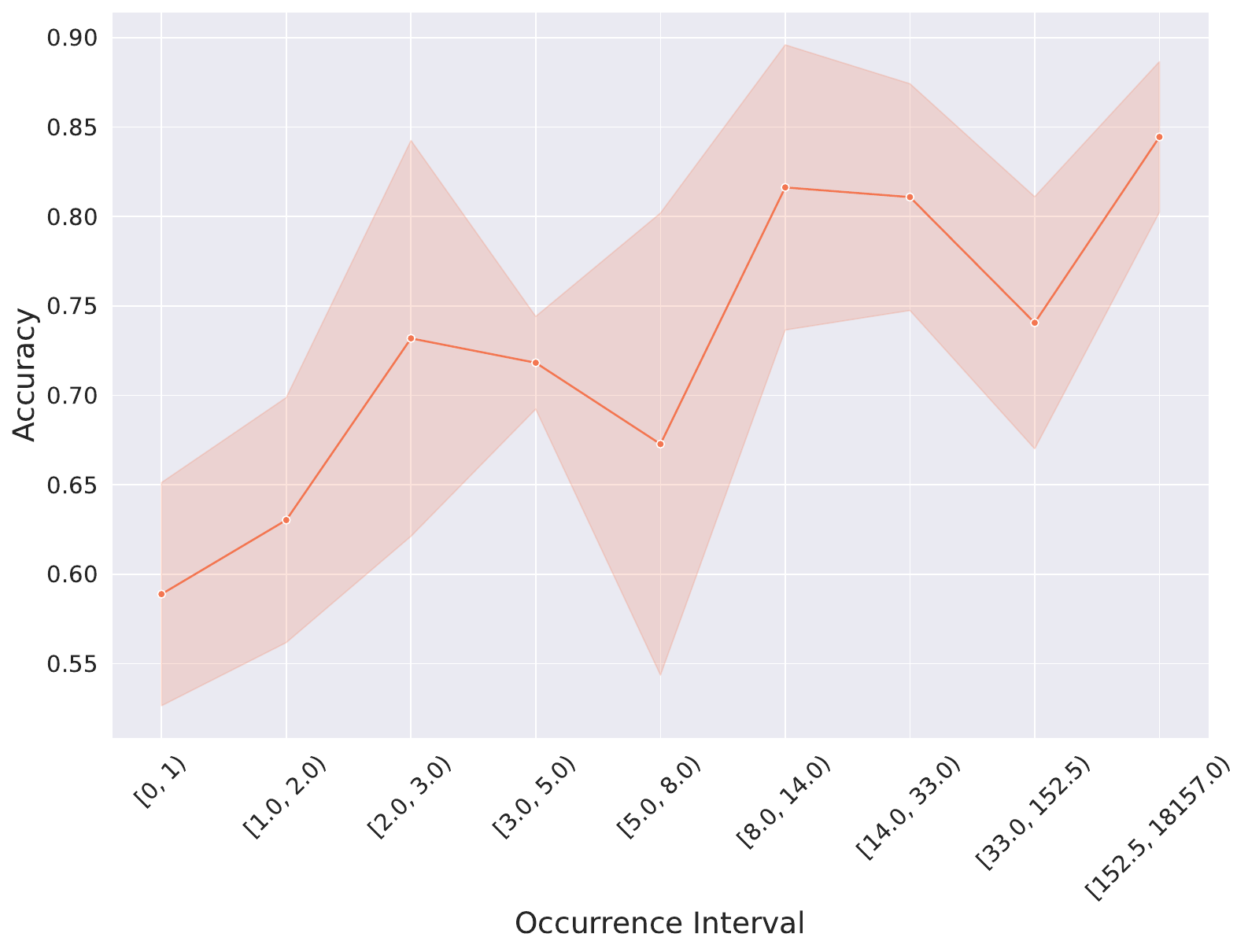} 
        \caption{Accuracy vs Occurrence at \conceptnet; Pearson (r=0.50); Spearman (r=0.87*)}
        \label{fig:occurrence_vs_accuracy_dolma_conceptnet_exact_match}
    \end{subfigure}
    \hspace{0.01\textwidth}
    \begin{subfigure}[t]{0.23\textwidth} 
        \includegraphics[width=\textwidth]{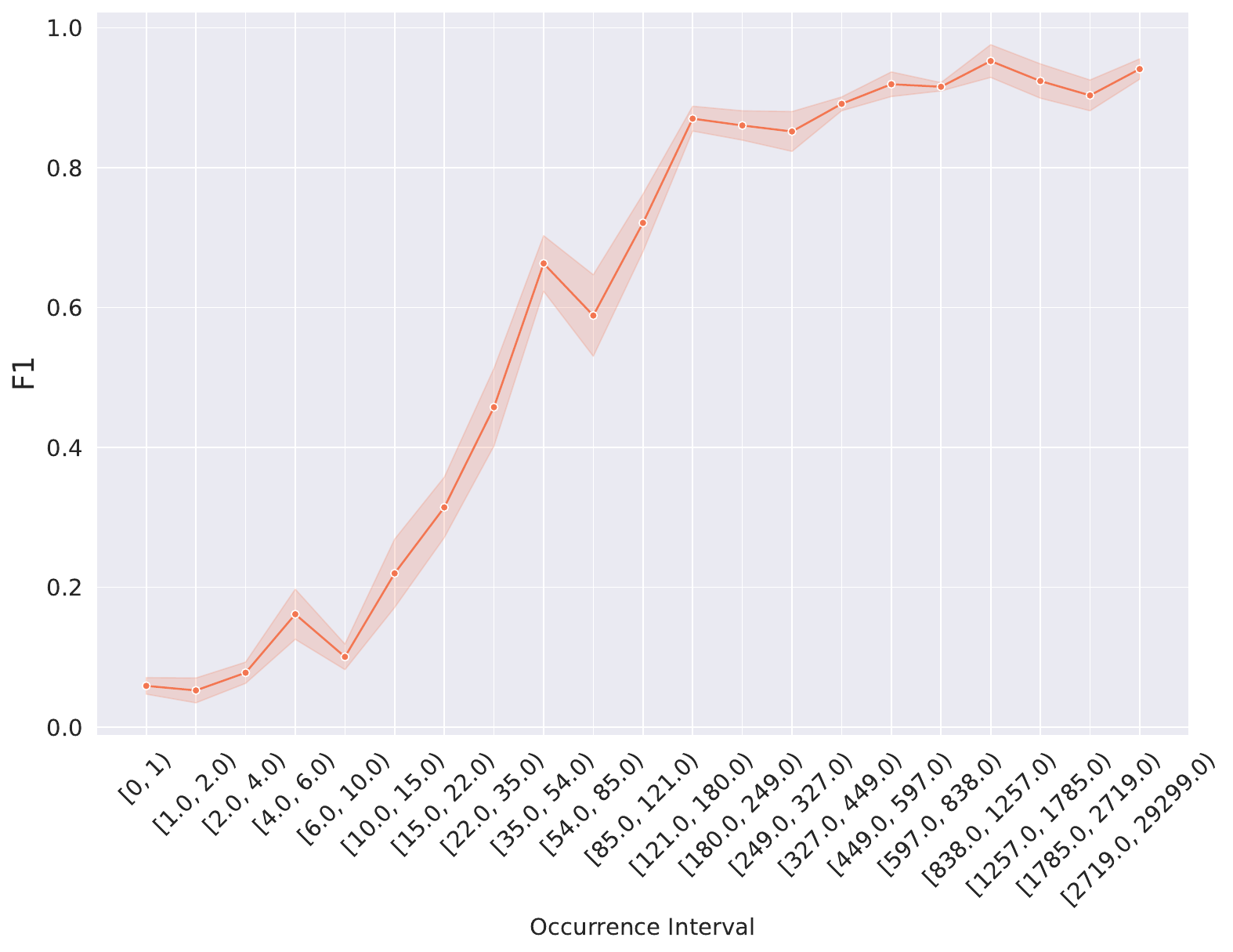} 
        \caption{F1 vs Occurrence at \causenet; Pearson (r=0.32); Spearman (r=0.97*)}
        \label{fig:occurrence_vs_f1_dolma_causenet_exact_match}
    \end{subfigure}
    \hspace{0.01\textwidth}
    \begin{subfigure}[t]{0.23\textwidth} 
        \includegraphics[width=\textwidth]{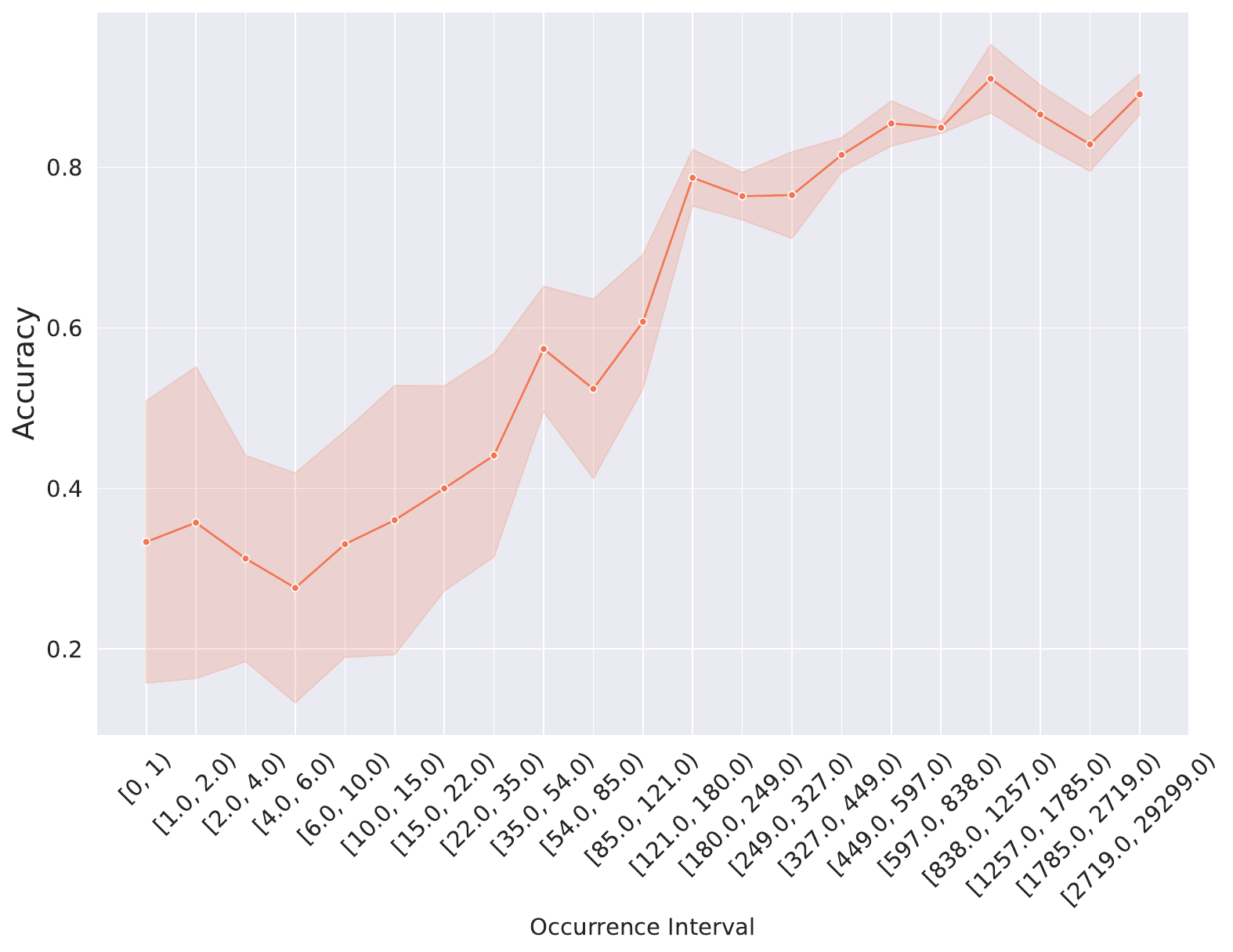} 
        \caption{Accuracy vs Occurrence at \causenet; Pearson (r=0.38); Spearman (r=0.96*)}
        \label{fig:occurrence_vs_accuracy_causenet_dolma_exact_match}
    \end{subfigure}
    \caption{The average F1 score and accuracy of \olmo by occurrence interval on causal direction identification task, averaged across 0 to 4 ICL examples. The occurrence data are derived from the exact matching query in the \dolma pre-training corpus. }
    \label{fig:dolma_exact_match}
\end{figure*}

\begin{figure*}[ht]
    \centering
    \begin{subfigure}[t]{0.23\textwidth} 
        \includegraphics[width=\textwidth]{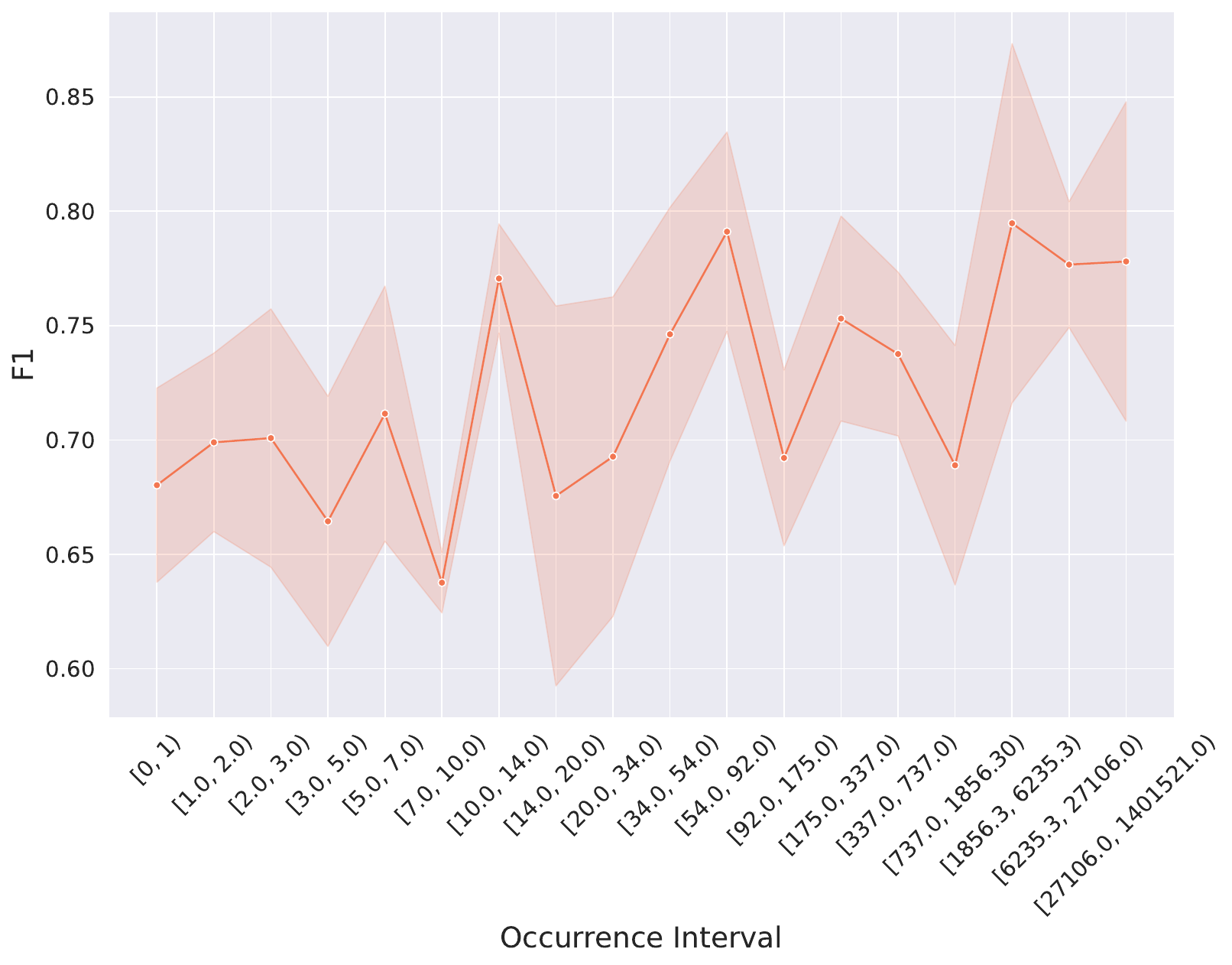} 
        \caption{F1 vs Occurrence at \conceptnet; Pearson (r=0.30); Spearman (r=0.58*)}
        \label{fig:occurrence_vs_f1_dolma_conceptnet_causal_co_occur}
    \end{subfigure}
    \hspace{0.01\textwidth}
    \begin{subfigure}[t]{0.23\textwidth} 
        \includegraphics[width=\textwidth]{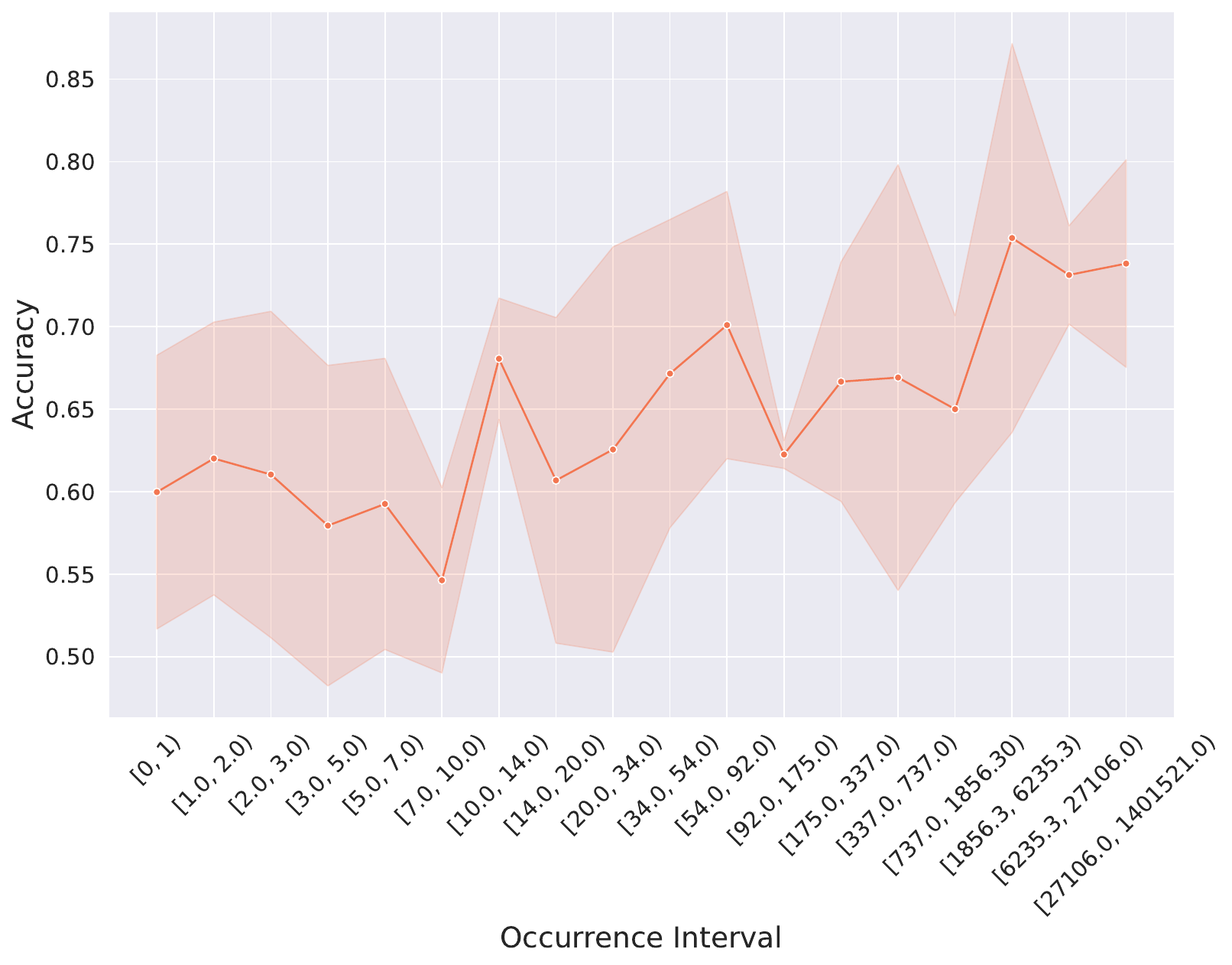} 
        \caption{Accuracy vs Occurrence at \conceptnet; Pearson (r=0.40); Spearman (r=0.78*)}
        \label{fig:occurrence_vs_accuracy_dolma_conceptnet_causal_co_occur}
    \end{subfigure}
    \hspace{0.01\textwidth}
    \begin{subfigure}[t]{0.23\textwidth} 
        \includegraphics[width=\textwidth]{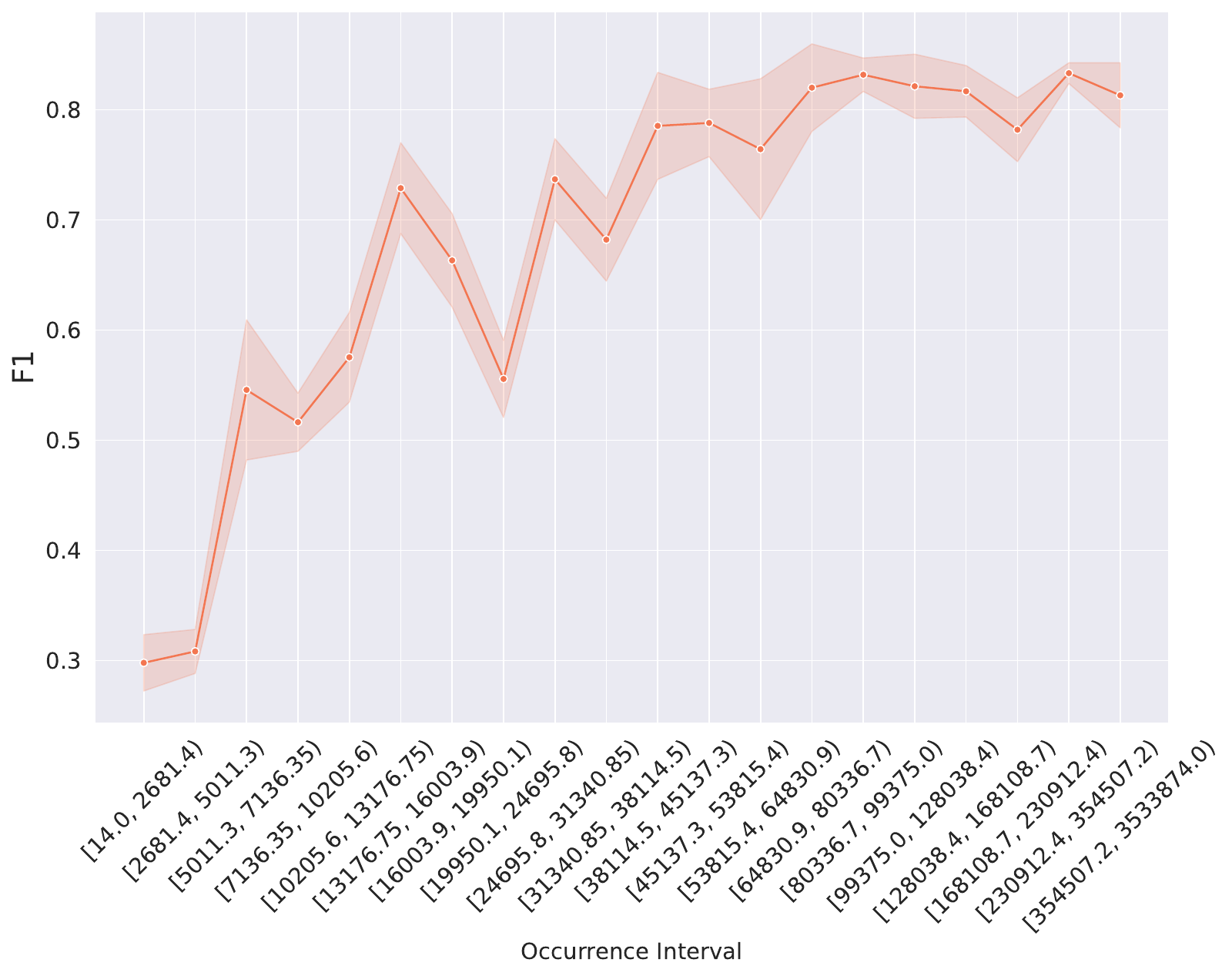} 
        \caption{F1 vs Occurrence at \causenet; Pearson (r=0.28); Spearman (r=0.90*)}
        \label{fig:occurrence_vs_f1_dolma_causenet_causal_co_occur}
    \end{subfigure}
    \hspace{0.01\textwidth}
    \begin{subfigure}[t]{0.23\textwidth} 
        \includegraphics[width=\textwidth]{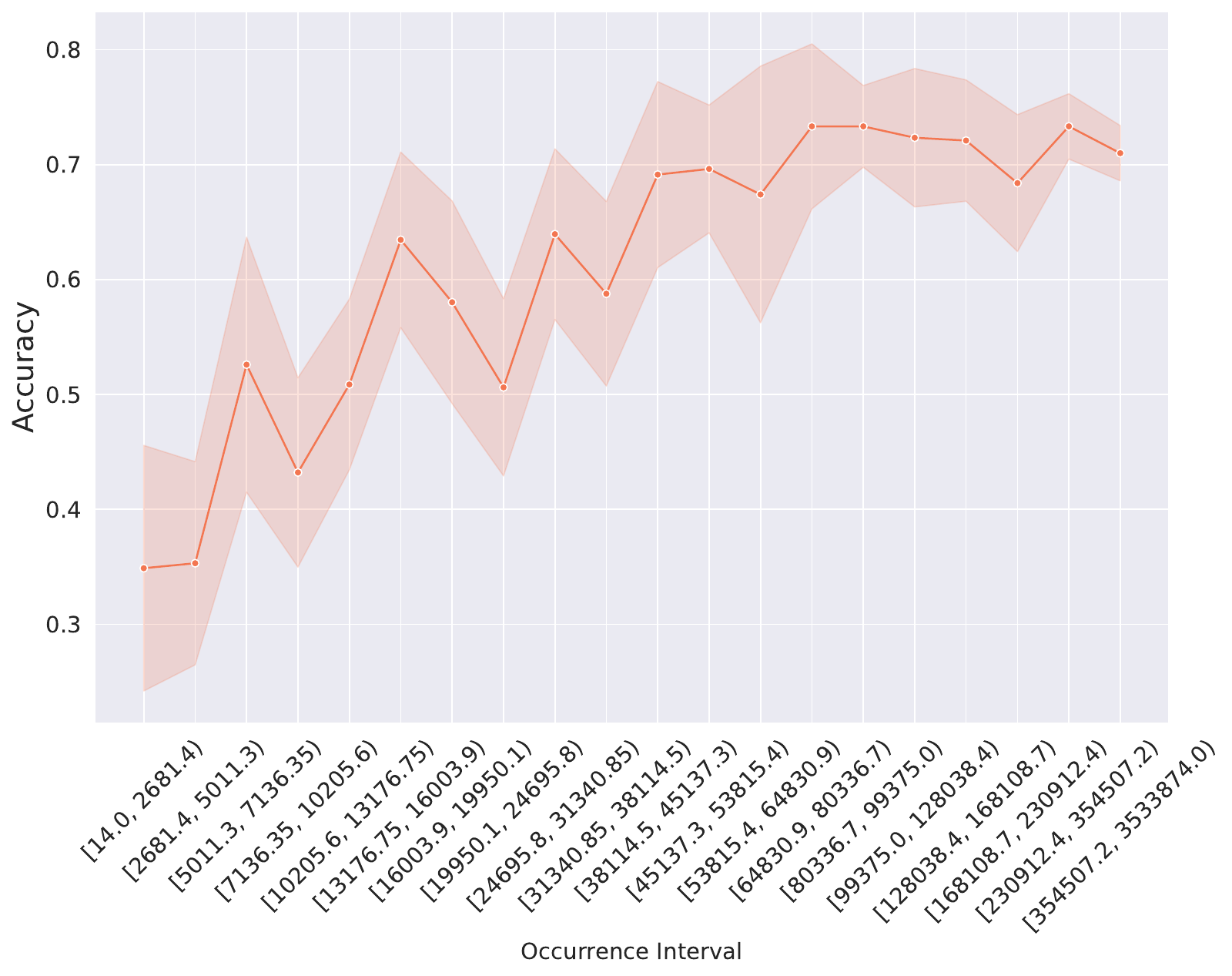} 
        \caption{Accuracy vs Occurrence at \causenet; Pearson (r=0.29); Spearman (r=0.88*)}
        \label{fig:occurrence_vs_accuracy_dolma_causenet_causal_co_occur}
    \end{subfigure}
    \caption{The average F1 score and accuracy of \olmo by occurrence interval on causal direction identification task, averaged across 0 to 4 ICL examples. The occurrence data are derived from the "event A" $\Rightarrow$ "causes" $\Rightarrow$ "event B" query in the \dolma pre-training corpus. }
    \label{fig:dolma_causal_co_occurrence}
\end{figure*}

\begin{figure*}[ht]
    \centering
    \begin{subfigure}[t]{0.23\textwidth} 
        \includegraphics[width=\textwidth]{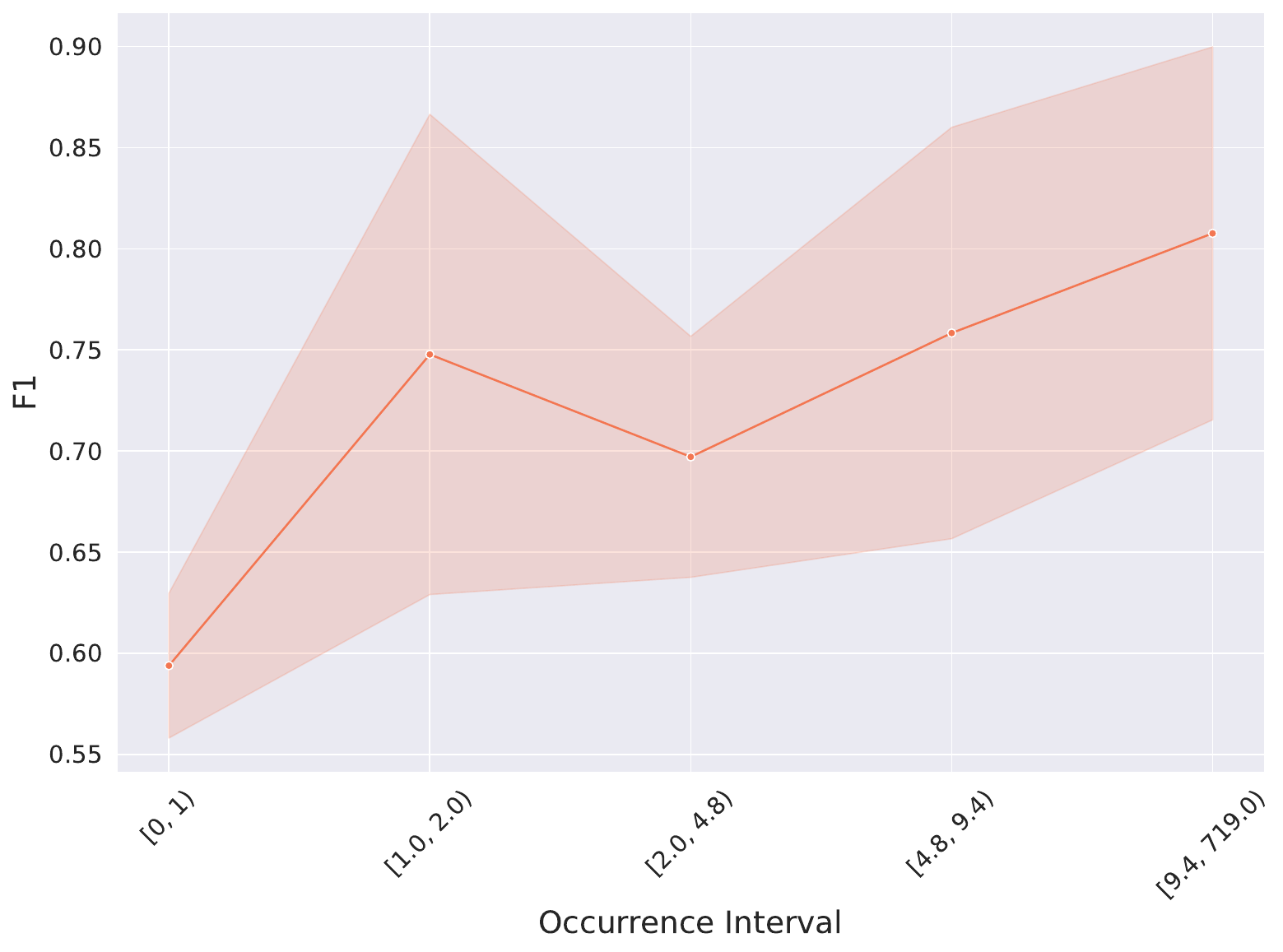} 
        \caption{F1 vs Occurrence at \conceptnet; Pearson (r=0.61); Spearman (r=0.90*)}
        \label{fig:occurrence_vs_f1_roots_conceptnet_exact_match}
    \end{subfigure}
    \hspace{0.01\textwidth}
    \begin{subfigure}[t]{0.23\textwidth} 
        \includegraphics[width=\textwidth]{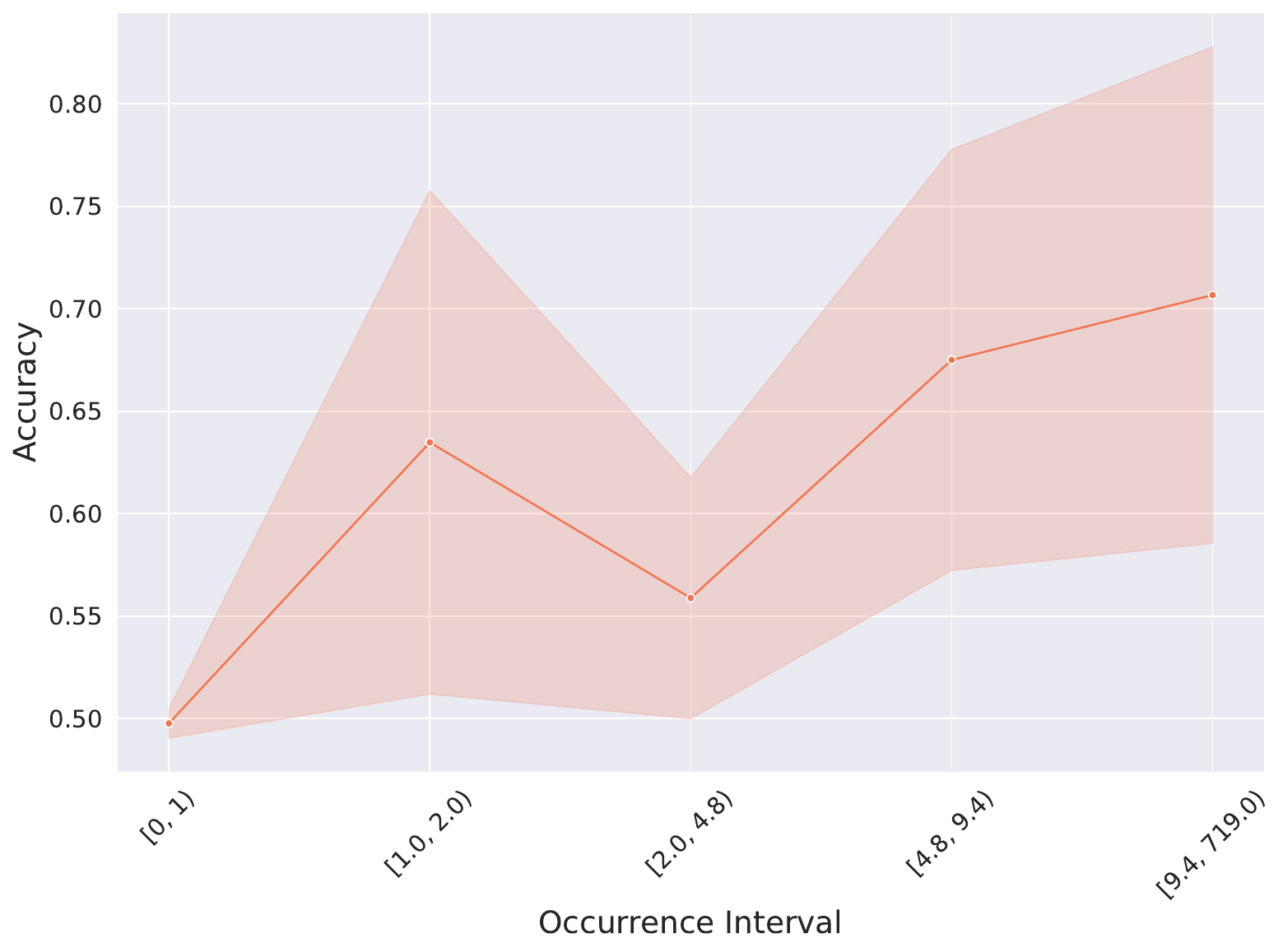} 
        \caption{Accuracy vs Occurrence at \conceptnet; Pearson (r=0.61); Spearman (r=0.90*)}
        \label{fig:occurrence_vs_accuracy_roots_conceptnet_exact_match}
    \end{subfigure}
    \hspace{0.01\textwidth}
    \begin{subfigure}[t]{0.23\textwidth} 
        \includegraphics[width=\textwidth]{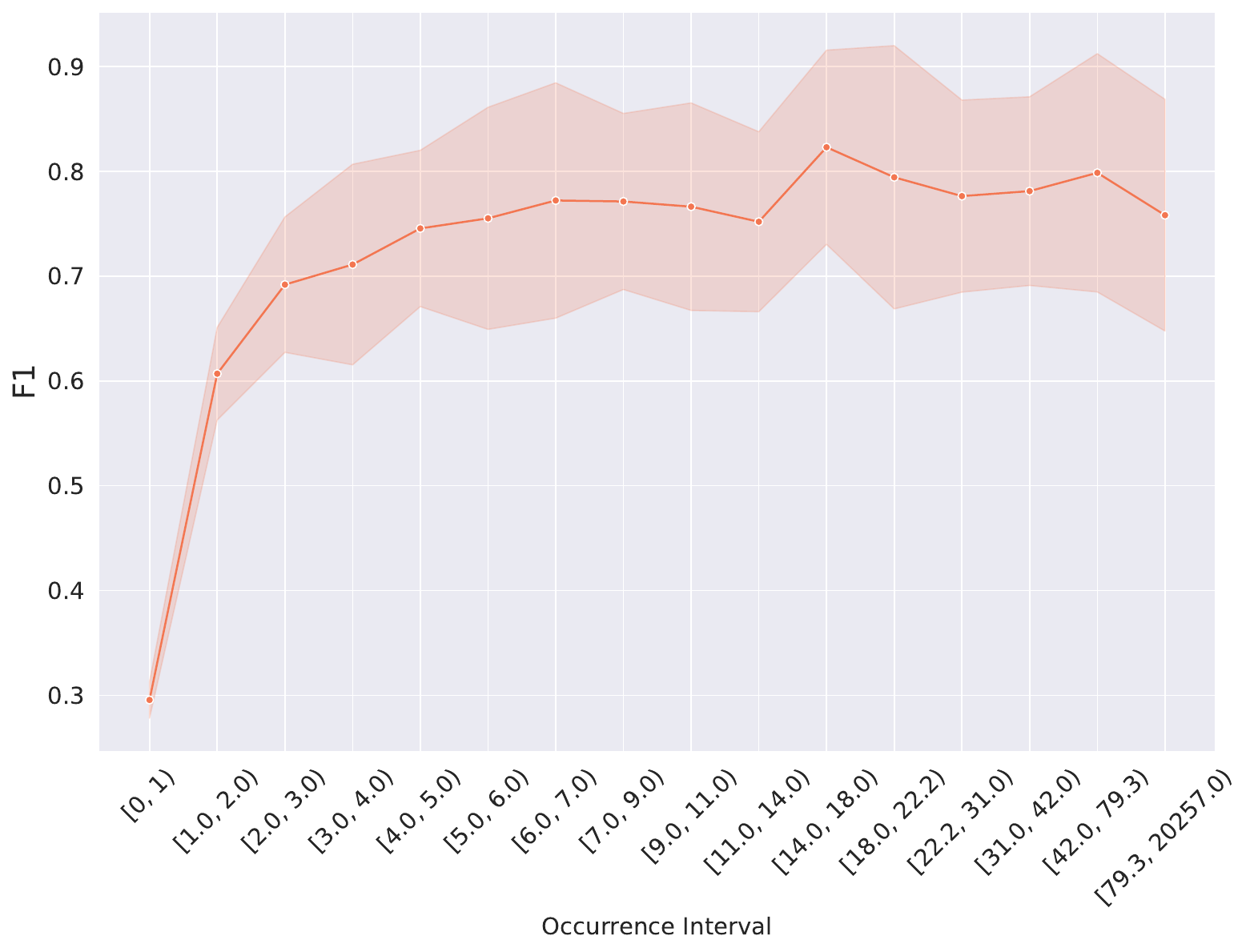} 
        \caption{F1 vs Occurrence at \causenet; Pearson (r=0.07); Spearman (r=0.81*)}
        \label{fig:occurrence_vs_f1_roots_causenet_exact_match}
    \end{subfigure}
    \hspace{0.01\textwidth}
    \begin{subfigure}[t]{0.23\textwidth} 
        \includegraphics[width=\textwidth]{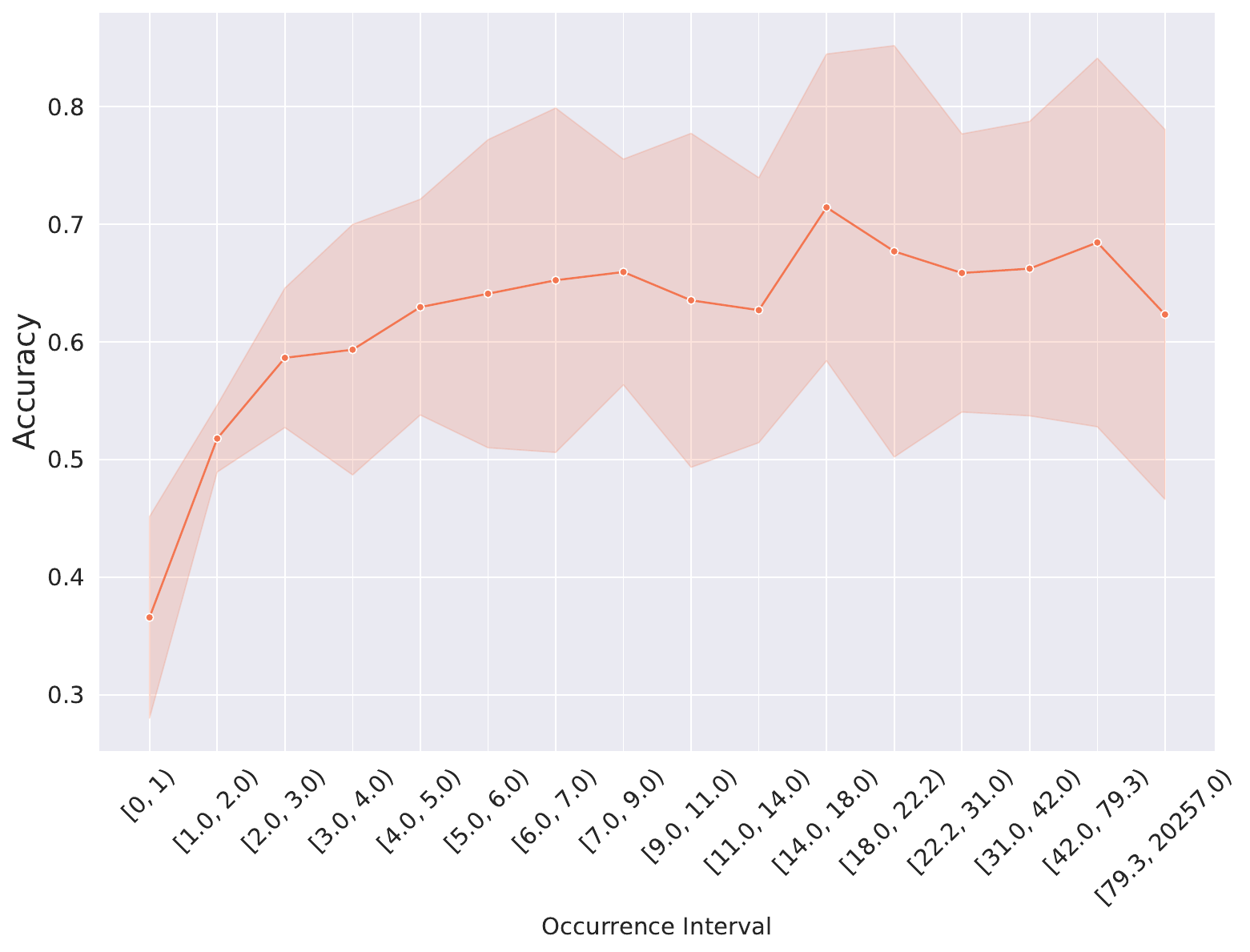} 
        \caption{Accuracy vs Occurrence at \causenet; Pearson (r=0.01); Spearman (r=0.69*)}
        \label{fig:occurrence_vs_accuracy_roots_causenet_exact_match}
    \end{subfigure}
    \caption{The average F1 score and accuracy of \bloom by occurrence interval on causal direction identification task, averaged across 0 to 4 ICL examples. The occurrence data are derived from the exact matching query in the \roots pre-training corpus. }
    \label{fig:roots_exact_match}
\end{figure*}

\begin{figure}[ht]
    \centering
    \begin{subfigure}[t]{0.23\textwidth} 
        \includegraphics[width=\textwidth]{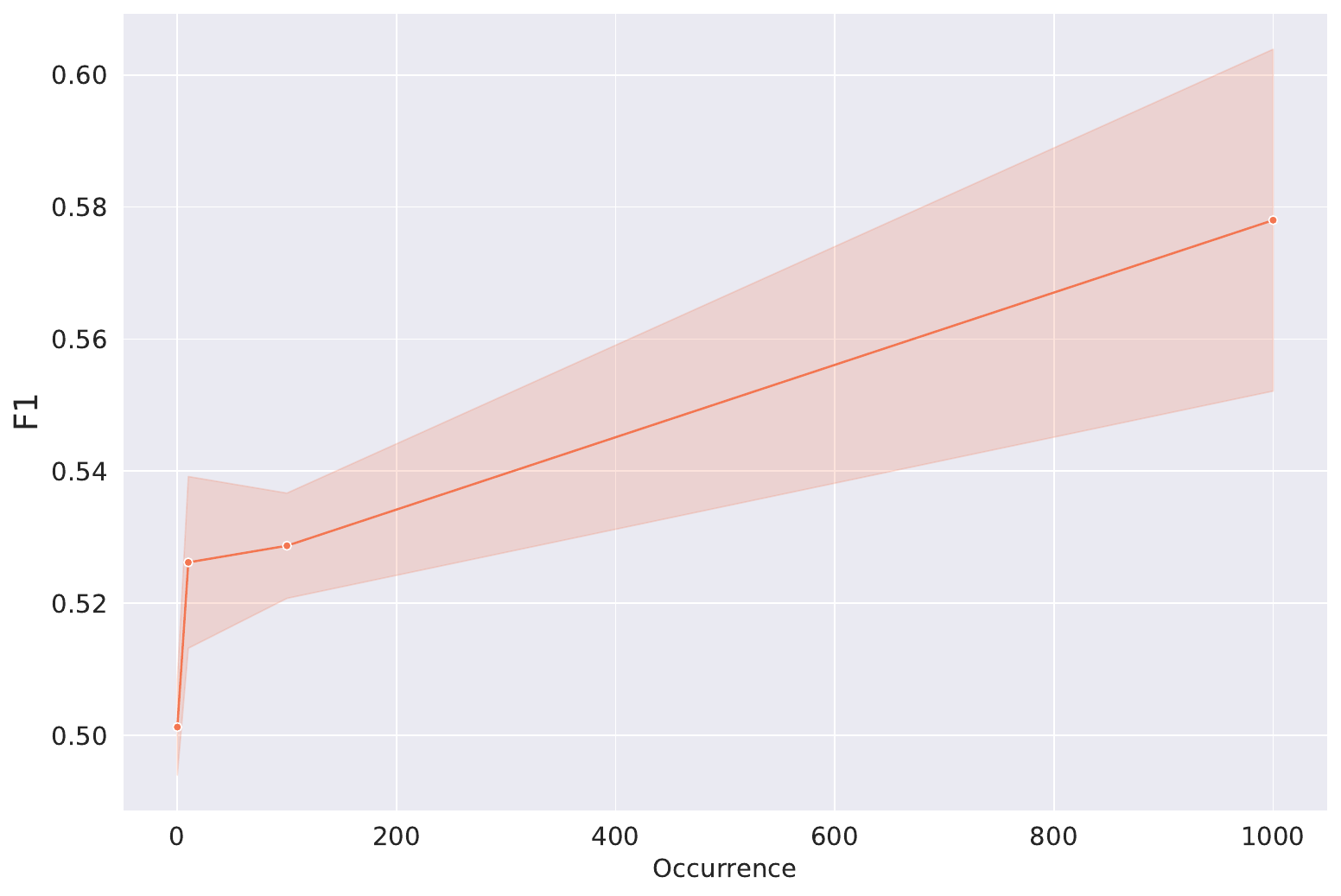} 
        \caption{F1 vs Occurrence; Pearson (r=0.94); Spearman (r=1.0*)}
        \label{fig:occurrence_vs_F1_olmo_xshot_correct_synthetic_relations_mean_std}
    \end{subfigure}
    \hspace{0.01\textwidth}
    \begin{subfigure}[t]{0.23\textwidth} 
        \includegraphics[width=\textwidth]{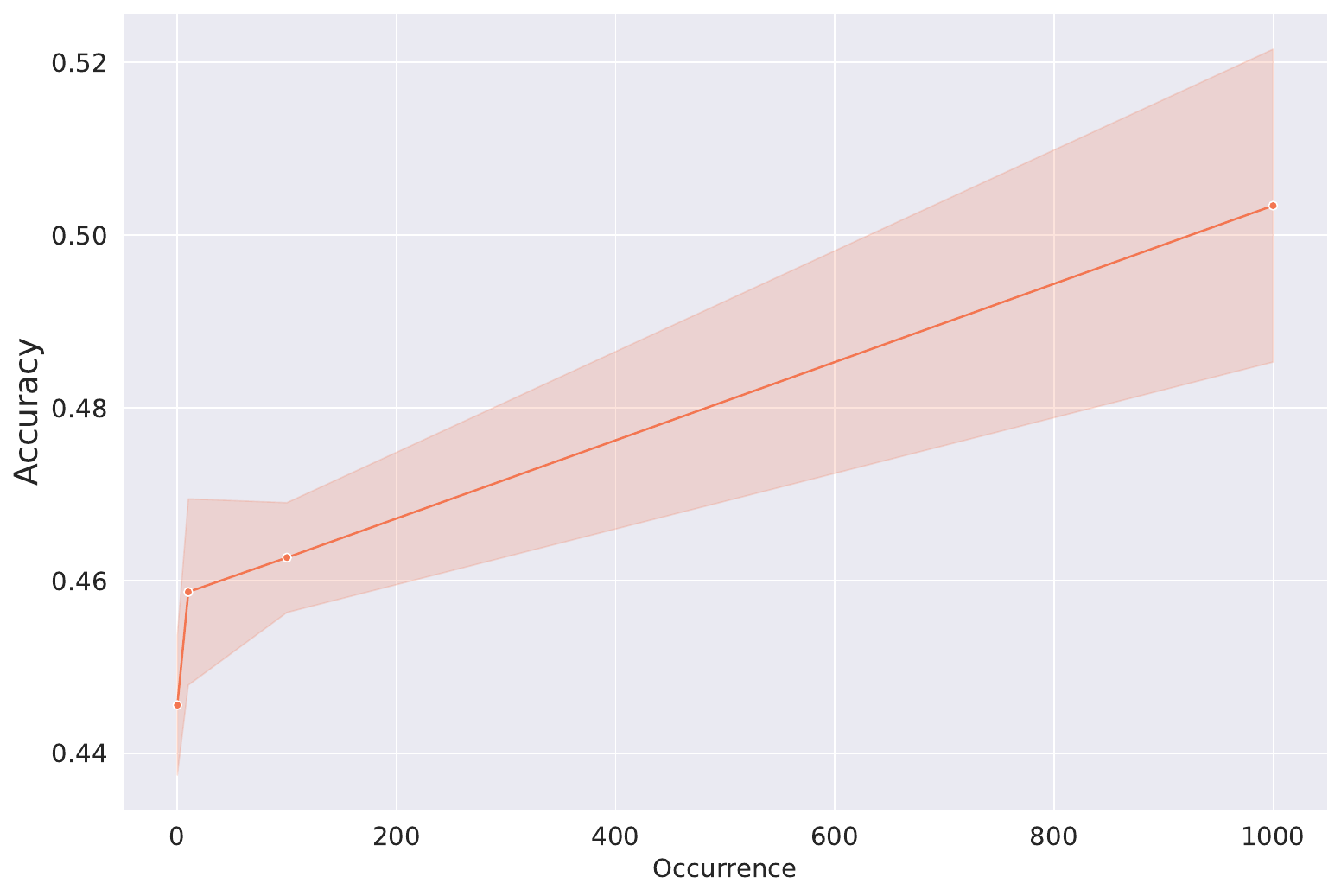} 
        \caption{Accuracy vs Occurrence; Pearson (r=0.97*); Spearman(r=1.0*)}
        \label{fig:occurrence_vs_Accuracy_olmo_xshot_correct_synthetic_relations_mean_std}
    \end{subfigure}
    \caption{The average F1 score and accuracy of fine-tuned \olmo by various occurrences on synthetic causal relations, averaged 0-4 ICL examples. }
    \label{fig:correct_synthetic_causal_relations_olmo_lora}
\end{figure}

\subsection{Datasets}
\paragraph{Tasks.} Following~\citep{kıcıman2023causal}, we consider the following two causal discovery tasks. \textit{Full Causal Discovery}. Given a set of random variables $\mathbf{X}$, for each possible pair of variables $(X_i, X_j)$, an LLM is instructed to identify whether: $X_i\rightarrow X_j$, $X_i\leftarrow X_j$, or no causal relation between $X_i$ and $X_j$. \textit{Causal Direction Identification}. Given two causally related variables $(X, Y)$, the causal direction identification task involves deciding whether $X\rightarrow Y$ or $X\leftarrow Y$ is true.  The full causal discovery and causal direction identification tasks can be treated as classification tasks. Therefore, we evaluate the results using F1 and accuracy. 
\subsubsection{Real-World Data}

\paragraph{Full Causal Discovery.}
 We consider six datasets for this task. We utilize four small causal graphs within the medical literature as our ground-truth causal graphs, which include \textbf{Alcohol}, \textbf{Cancer}, \textbf{Diabetes}, and \textbf{Obesity} (see Fig.~\ref{fig:four_small_graphs}) \cite{hernan2004structural, long2022can}. We also use a causal graph from atmospheric science, named \textbf{Arctic Sea Ice} \cite{huang2021benchmarking}. This causal graph explores the factors influencing arctic sea ice coverage. The Arctic Sea Ice is based on expert knowledge and consists of a causal graph with 12 variables and 46 edges, each edge derived from textbooks and peer-reviewed publications (see Fig.~\ref{fig:arctic_sea_ice}). Then, we employ a larger causal graph used for evaluating car \textbf{Insurance} risks \cite{binder1997adaptive}, which comprises 27 variables and 52 edges (see Fig.~\ref{fig:insurance}).  

\paragraph{Causal Direction Identification.} 
For this task, we consider two datasets derived from \textbf{\conceptnet} \cite{10.5555/3298023.3298212} and \textbf{\causenet} \cite{10.1145/3340531.3412763}. From \conceptnet, we select the top 1,900 causal relations based on confidence and generate an equal number of reverse-causal relations by swapping the cause and effect, resulting in 3,800 causal and reverse-causal relations. From \causenet, we select 814 high-confidence causal relations and create an equal number of reverse-causal relations, totaling 1,628 relations. These procedures are detailed in Appendix~\ref{apx:causal_direction_identification_task}.

\subsubsection{Synthetic Data}
\paragraph{Causal Direction Identification.}
We create a pre-training dataset including synthetic correct and incorrect causal relations that are absent in the original corpora. This dataset includes 100,000 documents randomly sampled from \dolma, with incorrect causal relations that either swap the positions of cause and effect or use negation templates such as "X does not cause Y." We generate 100 artificial causal relations using fictitious terms like 'blaonge' and 'goloneke'.  Utilizing predefined templates listed in Table~\ref{tab:imaginary_causal_relation_templates} in Appendix~\ref{apx:synthetic_causal_relations}, we craft mentions for both correct and incorrect causal relations. Then we create positive documents containing correct causal relations and negative documents containing incorrect causal relations by inserting these mentions between sentences within the documents. We adopt three approaches for the insertion of mentions.  \textbf{Correct Relation Scaling}: we vary the insertion of each correct causal relation from 0 to 1,000 occurrences. \textbf{Reverse Relation Scaling}: we first insert 1000 occurrences of each correct causal relation followed by inserting the corresponding reverse causal relations from 0 to 1,000 occurrences.  \textbf{Negated Relation Scaling}: After inserting 1,000 occurrences of each correct causal relation, we insert negations of these causal relations, from 0 to 1,000 occurrences. We then fine-tune \olmo utilizing LoRA \cite{hu2022lora} on synthetic datasets, with details provided in Appendix~\ref{apx:training_details}.

\subsection{Models}

\paragraph{Large Language Models.}
We conduct experiments using the following language models: \olmo \cite{groeneveld2024olmo}, \bloom \cite{workshop2023bloom}, \llamaTwo \cite{llama2}, \llamaThree \cite{llama3}, \gptThreeFive \cite{chatgpt} and \gptFouro \cite{gpt4o}. \olmo and \bloom provide access to both their pre-training corpora and model weights. \llamaTwo and \llamaThree have only released their model weights. \gptThreeFive and \gptFouro are closed-source models.  \olmo was pre-trained using the \dolma dataset \cite{soldaini2024dolma}, while \bloom utilized the \roots corpus \cite{laurenon2022the}. The release of corresponding search tools, WIMBD \cite{elazar2024whats} for \dolma and \roots Search \cite{piktus-etal-2023-roots} for \roots, enables the searching for causal relations.

\paragraph{In-Context Learning (ICL) and Prompt.}
For both the causal direction identification and the full causal discovery tasks, we utilize similar in-context learning demonstrations and prompts designed using the chain-of-thought strategy \cite{wei2022chain}, additional details provided in  Appendix~\ref{apx:in_context_learning_and_prompt}. When evaluating a pair of variables $(X, Y)$, we pose two questions to the LLMs: "Does X cause Y?" and "Does Y cause X?" The LLMs are expected to generate step-by-step explanations and provide a final response of either 'yes' or 'no'.

\subsection{Retrieval Query}

The pre-training corpus for \olmo is \dolma \cite{soldaini2024dolma}, which has a search tool named WIMBD \cite{elazar2024whats}. In our usage of WIMBD, we implement two search queries: an exact match for "event A causes event B"; an ordered phrase search for "event A" $\Rightarrow$ "causes" $\Rightarrow$ "event B". Here, $X \Rightarrow Y$ indicates that $X$ occurs before $Y$ within a predefined text window of 32 words. The search tool for \bloom pre-training corpus \roots \cite{laurenon2022the} is \roots Search \cite{piktus-etal-2023-roots}. Due to its limited search capability, we only utilize exact match in \roots Search. We also account for words with synonyms by identifying them using WordNet \cite{fellbaum1998wordnet}. In Table~\ref{tab:exact_match_query_wimbd}, ~\ref{tab:order_causal_query_wimbd} in Appendix~\ref{apx:query_for_search_engine}, we detail the methods used to create queries for retrieving causal relations.

%% file: sections/experimental_result_new.tex
\begin{RQ}
\label{rq:memorization_generalization}
Under what conditions do LLMs \textit{reliably} and \textit{consistently} make \textit{accurate} predictions in causal discovery?
\end{RQ}

Relations frequently occurring in pre-training data are likely memorized by LLMs. However, relations that are seldom or never present in pre-training data require LLMs to generalize these relations.

To address RQ \ref{rq:memorization_generalization}, we evaluate LLMs on causal relations across different occurrence intervals, which contain the similar number of causal relations. Causal relations with high occurrences are likely to be memorized by LLMs, whereas those with low occurrences suggest that an LLM is able to generalize \cite{carlini2023quantifying}. We then analyze the correlation between the occurrence of causal relations and the performance of LLMs on these causal relations.

\noindent \underline{\textit{Real-World Data}} We compute the average F1 and accuracy at each occurrence interval over various numbers of ICL examples (i.e., from 0-shot to 4-shot). The results are plotted with the x-axis representing occurrence intervals and the y-axis representing F1 or accuracy. Fig. ~\ref{fig:causal_discovery_dolma_frequency}-~\ref{fig:roots_exact_match} show that both F1 and accuracy exhibit a strong positive correlation with occurrence in the pre-training corpora. For instance, in the full causal discovery task, the Spearman correlation between F1 scores and occurrence rates is 0.9 using \olmo and its pre-training data.  Compared to highly frequent causal relations, LLMs exhibit significantly poorer performance when identifying low-frequency causal relations. For instance, in a full causal discovery task, \olmo achieves an F1 score of 0.88 in the highest occurrence interval, but only 0.2 in the lowest occurrence interval. In the causal direction identification task, \olmo reaches a 0.93 F1 score at the highest occurrence interval, compared to just 0.35 at the lowest. These results indicate that LLMs have a low generalization ability in causal discovery tasks.


\noindent \underline{\textit{Synthetic Data}} We train \olmo with Correct Relation Scaling. Fig.~\ref{fig:correct_synthetic_causal_relations_olmo_lora} demonstrates that both F1 and accuracy have a strong positive correlation with occurrence within the pre-training corpora, which aligns with real-world data.

\noindent \underline{\textit{Discussion}} These results demonstrate that while LLMs excel at recognizing causal relations through memorization, their capacity to generalize from less frequent or entirely novel data remains highly constrained. This limitation highlights the challenges in deploying LLMs in scenarios where causal relations are novel and absent from their pre-training data. Furthermore, this suggests the necessity of traditional statistical methods for causal discovery that rely solely on statistics to determine causal relations, irrespective of the novelty of causal relations. This insight suggests that future research might explore integrating traditional statistical methods with LLMs to enhance their generalization ability.


\begin{figure}[t]
    \centering
    \begin{subfigure}[t]{0.23\textwidth} 
        \includegraphics[width=\textwidth]{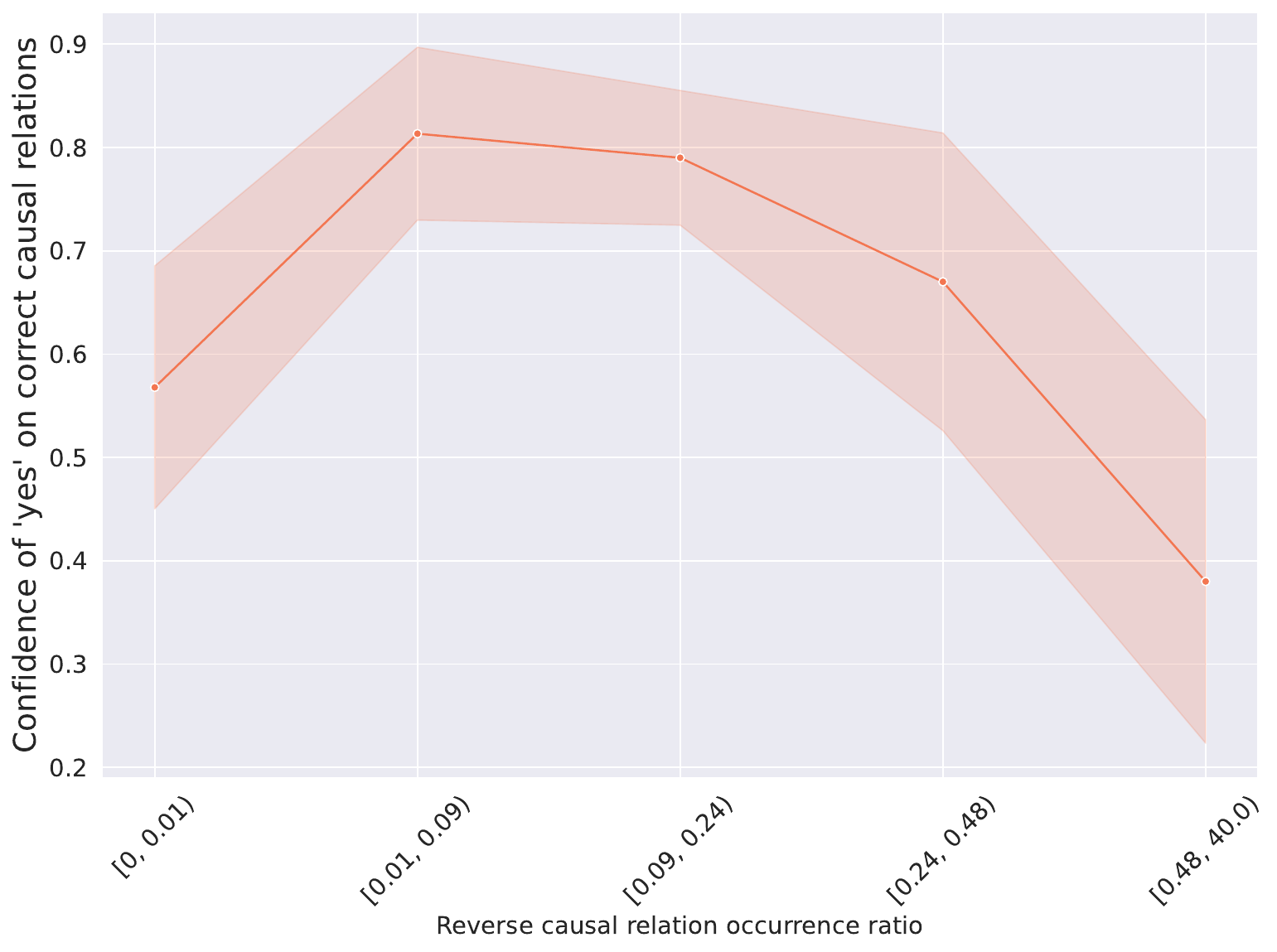} 
        \caption{Confidence vs Occurrence; Pearson (r=-0.83*), Spearman (r=-0.4)}
        \label{fig:negative_occurrence_vs_average_yes_confidence_olmo_xshot_causal_discovery_dolma_exact_match_mean_std}
    \end{subfigure}
    \hspace{0.01\textwidth}
    \begin{subfigure}[t]{0.23\textwidth} 
        \includegraphics[width=\textwidth]{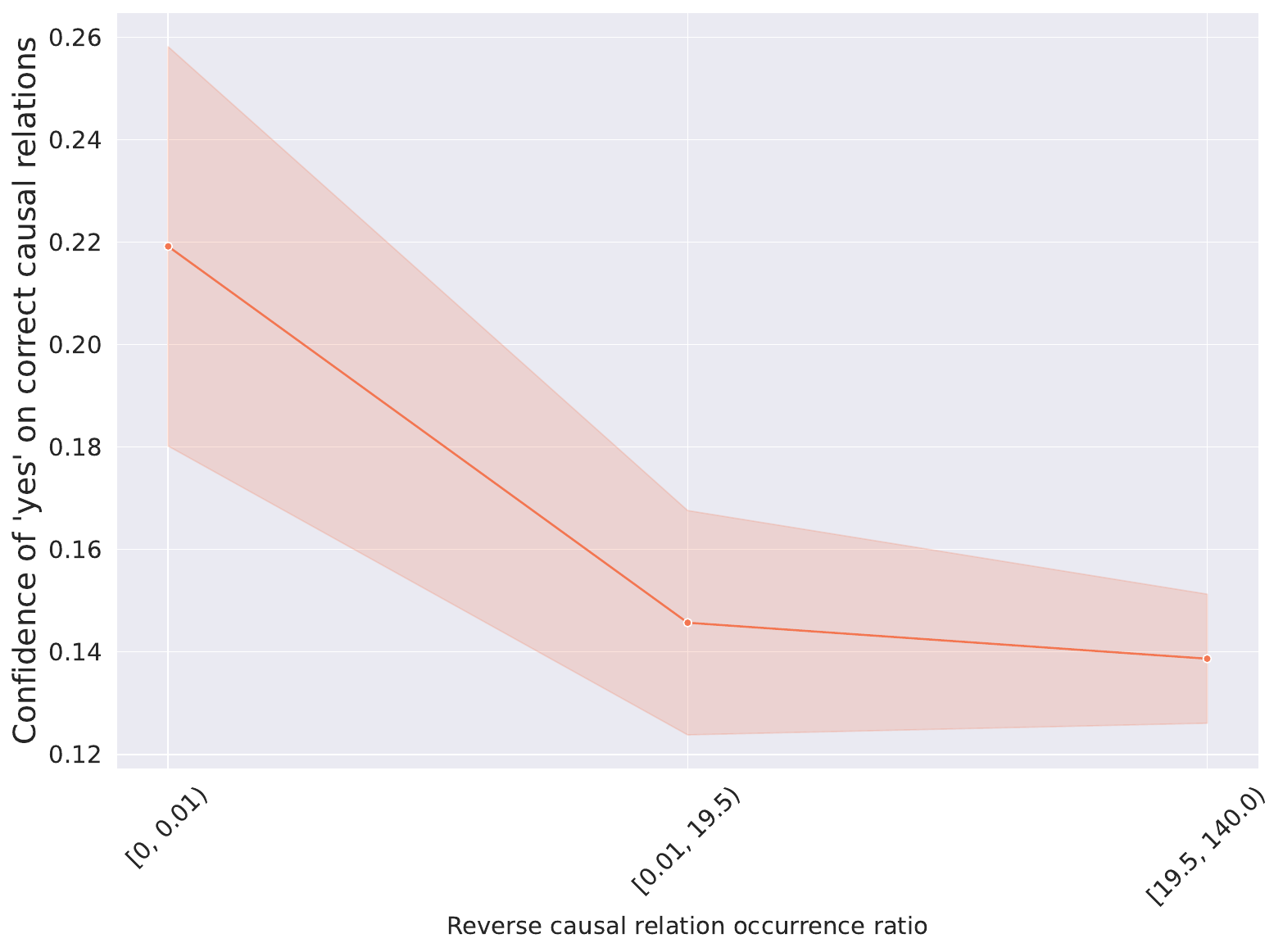} 
        \caption{Confidence vs Occurrence; Pearson (r=-0.66), Spearman (r=-1.0*)}
        \label{fig:negative_occurrence_vs_average_yes_confidence_bloom_xshot_causal_discovery_roots_exact_match_mean_std}
    \end{subfigure}
    \caption{The average confidence of correct causal relations on \olmo (a) and \bloom (b) by reverse casual relation occurrence ratio intervals on full causal discovery tasks. }
    \label{fig:confidence_negative_ratio_causal_discovery}
\vspace{-15pt}
\end{figure}

\begin{figure*}[ht]
    \centering
    \begin{subfigure}[t]{0.23\textwidth} 
        \includegraphics[width=\textwidth]{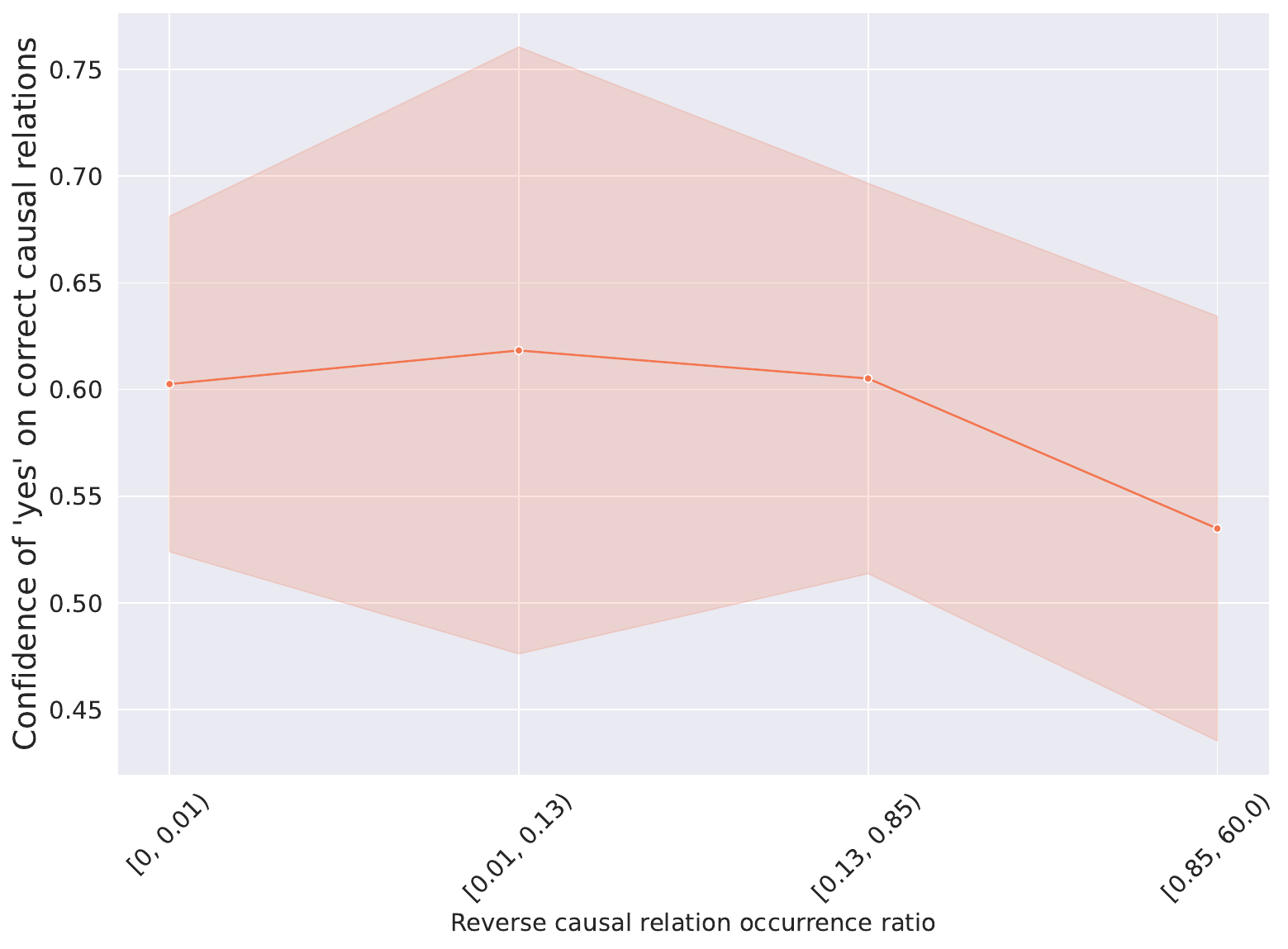} 
        \caption{Confidence vs Occurrence at \conceptnet; Pearson (r=-0.98*), Spearman (r=-0.4)}
        \label{fig:negative_occurrence_vs_average_yes_confidence_olmo_xshot_conceptnet_dolma_exact_match}
    \end{subfigure}
    \hspace{0.01\textwidth}
    \begin{subfigure}[t]{0.23\textwidth} 
        \includegraphics[width=\textwidth]{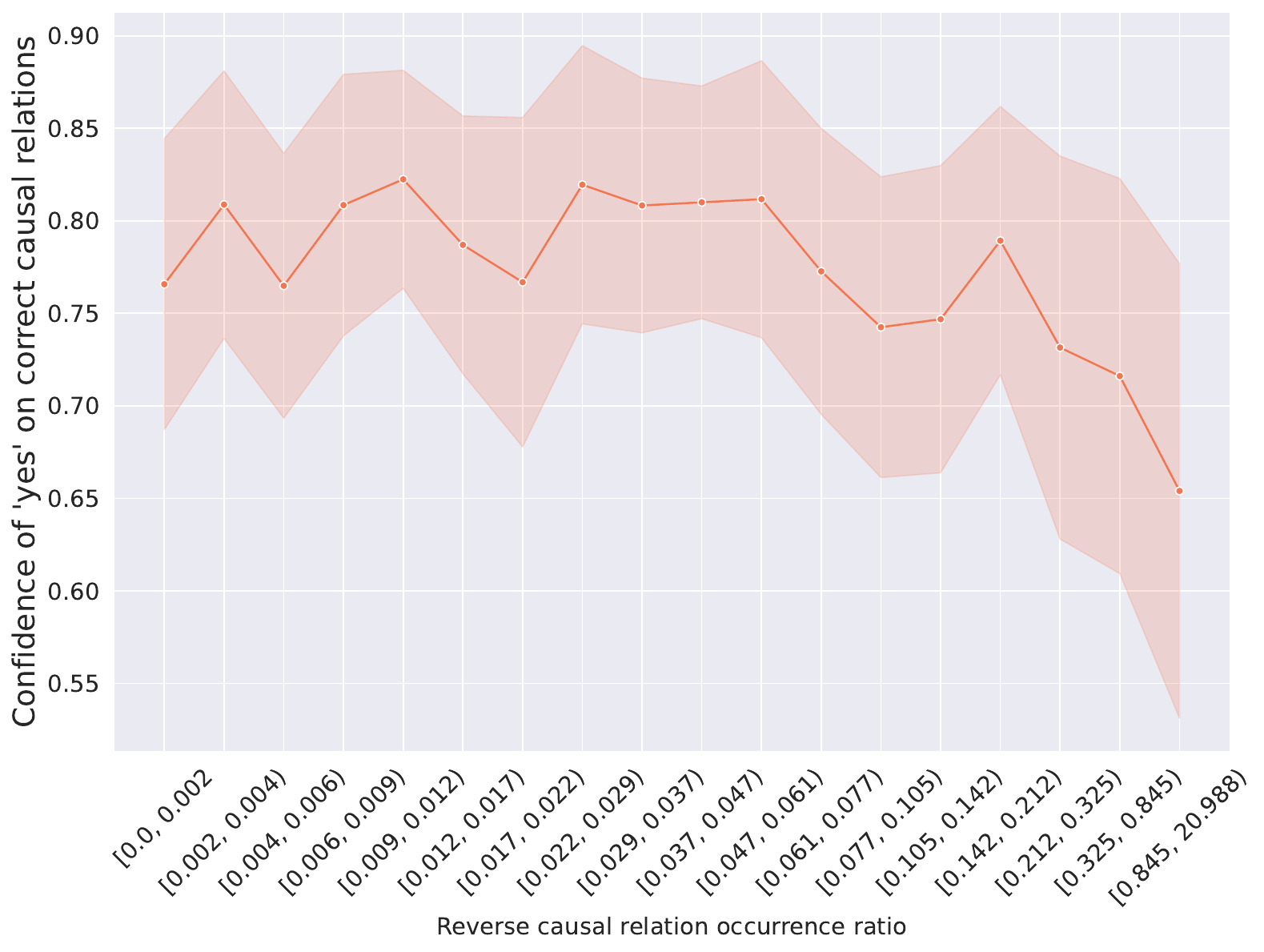} 
        \caption{Confidence vs Occurrence at \causenet; Pearson (r=-0.71*), Spearman (r=-0.51*)}
        \label{fig:negative_occurrence_vs_average_yes_confidence_olmo_xshot_causenet_dolma_exact_match}
    \end{subfigure}
    \hspace{0.01\textwidth}
    \begin{subfigure}[t]{0.23\textwidth} 
        \includegraphics[width=\textwidth]{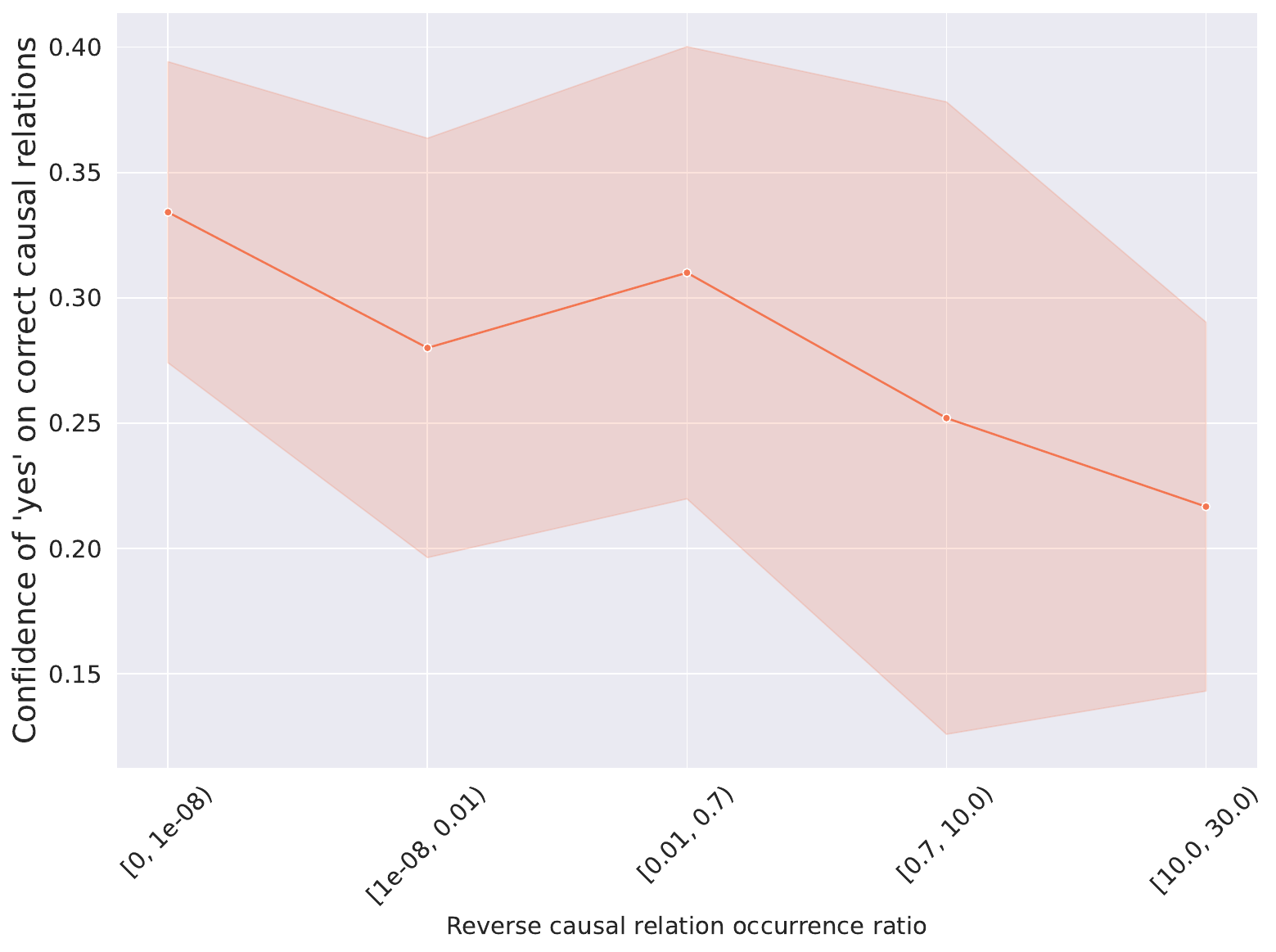} 
        \caption{Confidence vs Occurrence at \conceptnet; Pearson (r=-0.86), Spearman (r=-0.89*)}
        \label{fig:negative_occurrence_vs_average_yes_confidence_bloom_xshot_conceptnet_roots_exact_match_mean_std}
    \end{subfigure}
    \hspace{0.01\textwidth}
    \begin{subfigure}[t]{0.23\textwidth} 
        \includegraphics[width=\textwidth]{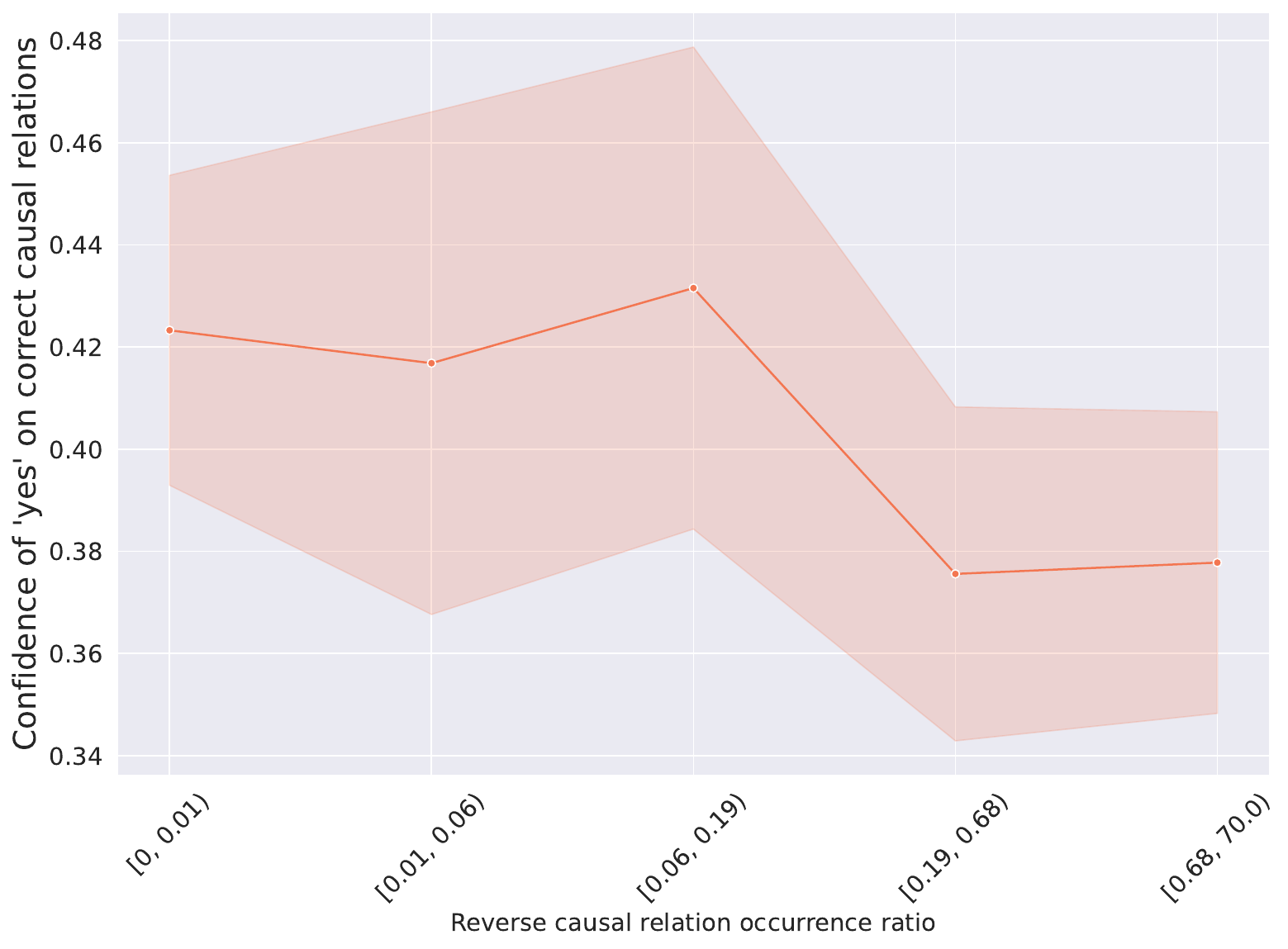} 
        \caption{Confidence vs Occurrence at \causenet; Pearson (r=-0.58), Spearman (r=-0.6)}
        \label{fig:negative_occurrence_vs_average_yes_confidence_bloom_xshot_causenet_roots_exact_match_mean_std}
    \end{subfigure}
    \caption{The average confidence of correct causal relations on \olmo (a,b) and \bloom (c,d) by reverse casual relation occurrence ratio intervals on causal direction identification, averaged 0-4 ICL examples. }
    \label{fig:confidence_negative_ratio_causal_direction}
\end{figure*}

\begin{figure}[ht]
    \centering
    \begin{subfigure}[t]{0.23\textwidth} 
        \includegraphics[width=\textwidth]{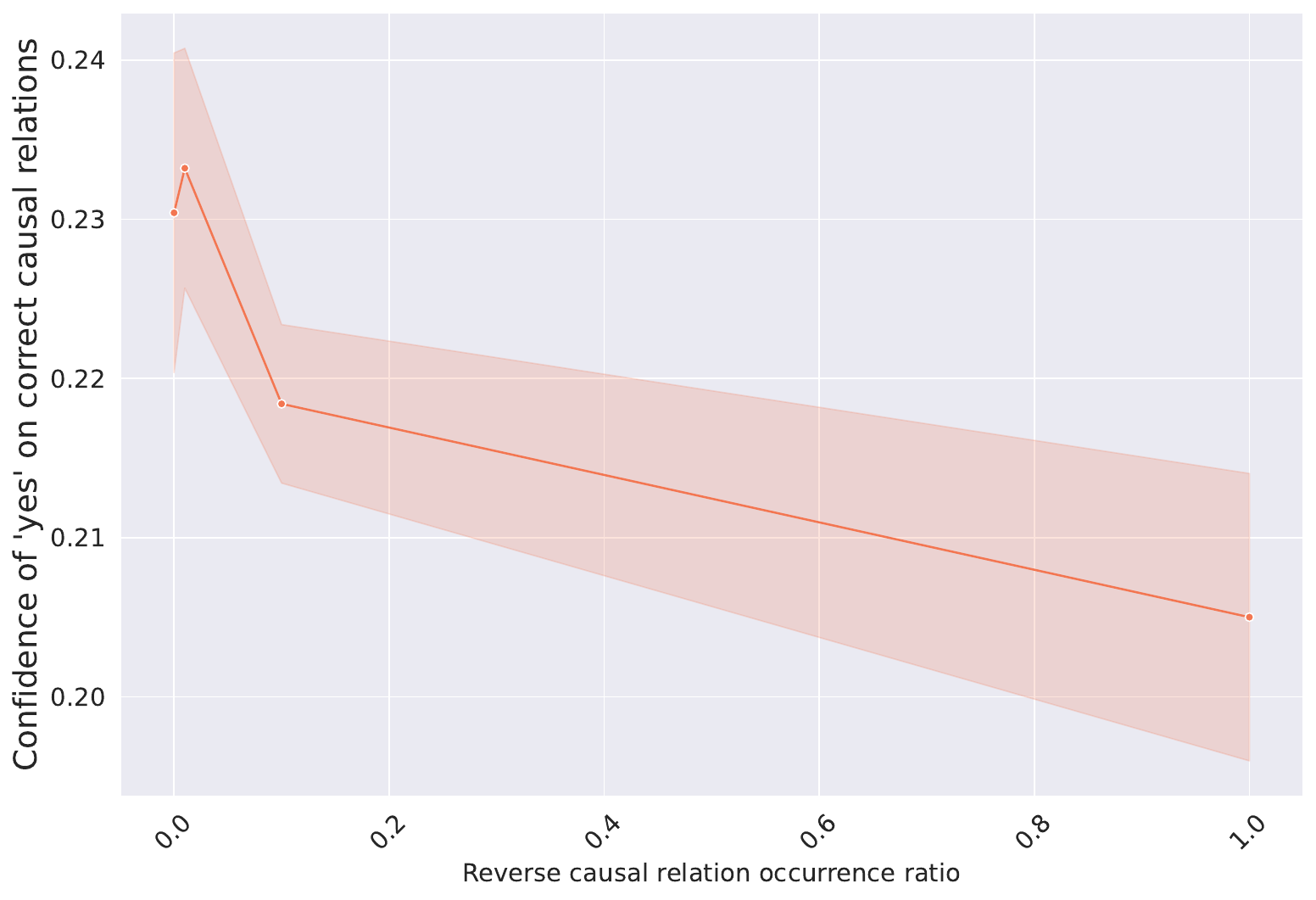} 
        \caption{Reverse Relation Scaling; Pearson (r=-0.91), Spearman (r=-0.8)}
        \label{fig:negative_occurrence_vs_average_yes_confidence_olmo_xshot_reverse_synthetic_relations_mean_std}
    \end{subfigure}
    \hspace{0.01\textwidth}
    \begin{subfigure}[t]{0.23\textwidth} 
        \includegraphics[width=\textwidth]{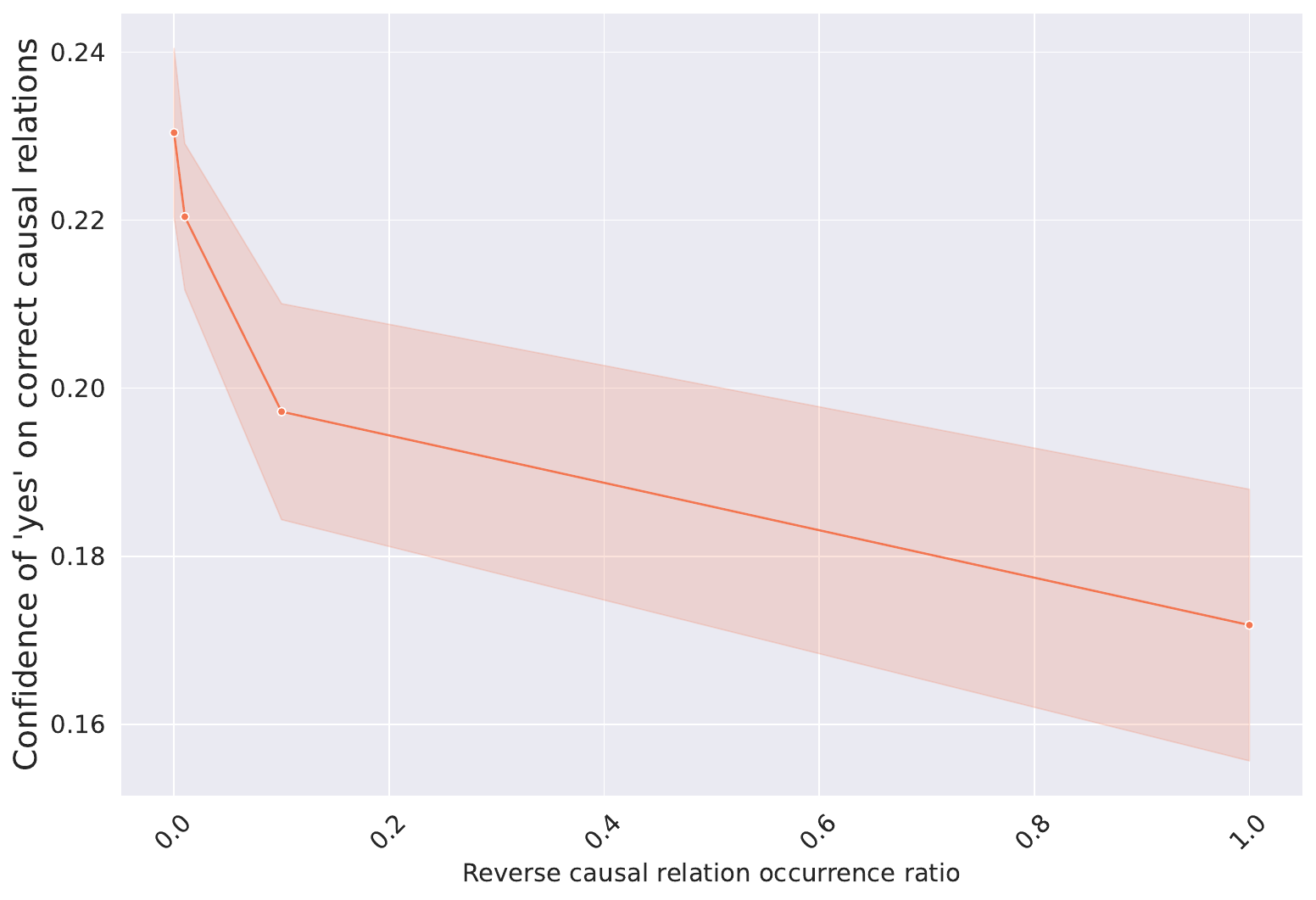} 
        \caption{Negated Relation Scaling; Pearson (r=-0.89), Spearman (r=-1.0*)}
        \label{fig:negative_occurrence_vs_average_yes_confidence_olmo_xshot_negation_synthetic_relations_mean_std}
    \end{subfigure}
    \caption{The average confidence of correct causal relations on fine-tuned \olmo by reverse casual relation occurrence ratio (a) and negation casual relation occurrence ratio (b) on synthetic causal relations, averaged across 0 to 4 ICL examples. }
    \label{fig:confidence_negative_ratio_synthetic_causal_relations}
\end{figure}

\begin{RQ}
\label{rq:anti_causal_occur}
How does the occurrence of incorrect causal relations affect LLMs in causal discovery tasks?
\end{RQ}

Incorrect causal relations include reversals of correct causal relations (e.g., lung cancer causes smoking) and negations of correct causal relations (e.g., smoking does not cause lung cancer).

We hypothesize that when both correct and incorrect causal relations are frequent, LLMs may struggle to discern the correct relations, thereby reducing their confidence in correct causal relations. To investigate this, we examine the correlation between the occurrence ratio of incorrect causal relations and LLMs' confidence in correct causal relations. The occurrence ratio is defined as the number of incorrect causal relations divided by the number of corresponding correct causal relations. For example, if the phrase "smoking causes lung cancer" appears 13,652 times and its reverse "lung cancer causes smoking" appears 99 times, the resulting occurrence ratio is approximately 0.007. Confidence in correct causal relations (\ie affirmative confidence) is measured by the proportion of affirmative responses among multiple generated responses, where a response is considered affirmative if it contains "yes" and negative if it contains "no". If neither "yes" nor "no" appears in an answer, we classify it as a 'fail'. The average proportion of 'fail' across all datasets is 0.03, indicating that most responses are either 'yes' or 'no'. If the query "Does smoking cause lung cancer?" results in affirmative responses in 8 out of 10 generation samples, the affirmative confidence for "smoking causes lung cancer" is 0.8. In this experiment, we sample 10 responses for each query.

\noindent \underline{\textit{Real-World Data}} We calculate and plot the correlation between different intervals of occurrence ratios of incorrect causal relations and affirmative confidence. The experiment results, shown in Fig. ~\ref{fig:confidence_negative_ratio_causal_discovery} and ~\ref{fig:confidence_negative_ratio_causal_direction}, indicate a negative correlation, showing that LLMs' confidence in correct causal relations decreases as the occurrence ratio of incorrect causal relations increases. 

\noindent \underline{\textit{Synthetic Data}} We fine-tune \olmo employing both Reverse Relation Scaling and Negated Relation Scaling.  Fig. ~\ref{fig:confidence_negative_ratio_synthetic_causal_relations} shows a similar negative correlation with real-world data: as the occurrence of incorrect causal relations increases, there is a decline in the LLMs' confidence in the corresponding correct causal relations.

\noindent \underline{\textit{Discussion}} 
This negative correlation suggests that while LLMs excel at memorizing frequently occurring information, they struggle to discern the correct relation when confronted with high frequencies of conflicting data. This inability leads to a loss of confidence in correct causal relations. This finding underscores the necessity of not only enhancing the presence of correct information but also of eliminating misinformation in pre-training data. Furthermore, these results pave the way for future research aimed at developing models that can manage conflicting information within their pre-training data.

\input{tables/LLMs_results_causal_relation_contexts_short}

\begin{RQ}
\label{rq:causal_context}
How does the context of a causal relation influence LLM performance in causal discovery tasks?
\end{RQ}

We hypothesize that the strength and validity of causal relations can vary across different contexts. Thus, when a causal discovery question is presented with different contexts, LLMs might provide different and sometimes opposite answers to the causal relation's validity. 

From \conceptnet and \causenet, we select 100 high-confidence correct causal relations from each. Since both \conceptnet and \causenet lack context information, for each causal relation, we use \gptFouro to generate five positive contexts that enhance it and five negative contexts that weaken it. Then we hire thirteen annotators to evaluate these causal relations under different contexts in three rounds. The prompt and evaluation details are presented in Appendix~\ref{sec:human_eval_causal_relation_contexts}. The agreement between annotators and \gptFouro is 0.76 using Krippendorff's Alpha \cite{castro-2017-fast-krippendorff}. We then assess the performance of LLMs on these causal relations within positive and negative contexts. The query format is similar to Table~\ref{tab:icl_prompt}, except we provide context information using the phrase "Given the scenario: \{description\}". We assess LLM performance on correct causal relations within various contexts using the affirmative ratio. This ratio is calculated by dividing the number of correct causal relations identified by the LLM by the total number of correct causal relations presented.

\noindent \underline{\textit{Observation}} From the results in Table~\ref{tab:LLMs_results_causal_relation_contexts_short}, we observe that all LLMs are more likely to identify causal relations in positive contexts compared to no context. In contrast, adding negative contexts significantly decreases LLMs' ability to identify causal relations compared to no context. These results indicate that the validity and strength of causal relations can vary in different contexts.

\noindent \underline{\textit{Discussion}} 
The significant variation in causal relation identification across positive and negative contexts indicates the context sensitivity of LLM-based causal discovery methods. This observation suggests that LLM-based algorithms should explicitly provide contextual information to enable LLMs to better understand the scenario and thereby make more accurate predictions. It is particularly crucial for these algorithms to avoid misleading contexts, as our results demonstrate that negative contexts can substantially impair LLM performance. Furthermore, investigating the underlying mechanisms of how different contexts influence the strength and validity of causal relations could be a promising direction for future research.

%% file: tables/LLMs_results_causal_relation_contexts_short.tex
\begin{table}[ht]
\centering
\resizebox{\linewidth}{!}{%
\begin{tabular}{lccc}
\hline
\multicolumn{4}{c}{Full Causal Discovery Task}             \\
                         & w/o Ctx    & P.Ctx    & N.Ctx   \\ \hline
\olmo (3 ICL)            & 0.65       & 0.87     & 0.42    \\
\bloom (3 ICL)           & 0.62       & 0.76     & 0.59    \\
\llamaTwo (3 ICL)        & 0.68       & 0.85     & 0.25    \\
\llamaThree (3 ICL)      & 0.67       & 0.73     & 0.20    \\
\gptThreeFive (3 ICL)    & 0.65       & 0.86     & 0.24    \\
\gptFouro (3 ICL)        & 0.69       & 0.92     & 0.27    \\
Avg.                     & 0.66       & 0.83*    & 0.33*   \\ \hline
\multicolumn{4}{c}{Causal Direction Identification Task}   \\
                         & \multicolumn{3}{c}{\conceptnet} \\
\olmo (3 ICL)            & 0.9        & 0.95     & 0.62    \\
\bloom (3 ICL)           & 0.79       & 0.81     & 0.70    \\
\llamaTwo (3 ICL)        & 0.79       & 0.95     & 0.31    \\
\llamaThree (3 ICL)      & 0.66       & 0.85     & 0.10    \\
\gptThreeFive (3 ICL)    & 0.77       & 0.90     & 0.33    \\
\gptFouro (3 ICL)        & 0.87       & 0.96     & 0.34    \\
                         & \multicolumn{3}{c}{\causenet}   \\
\olmo (3 ICL)            & 0.89       & 0.99     & 0.61    \\
\bloom (3 ICL)           & 0.72       & 0.78     & 0.63    \\
\llamaTwo (3 ICL)        & 0.92       & 0.99     & 0.47    \\
\llamaThree (3 ICL)      & 0.88       & 0.94     & 0.14    \\
\gptThreeFive (3 ICL)    & 0.93       & 0.98     & 0.67    \\
\gptFouro (3 ICL)        & 0.98       & 0.99     & 0.60    \\
Avg.                     & 0.84       & 0.92*    & 0.46*   \\ \hline
\end{tabular}%
}
\caption{Affirmative ratio of LLMs on causal relations across different contexts. An asterisk (*) indicates a statistically significant difference (p-value < 0.05) between the affirmative ratio in positive/negative contexts and the affirmative ratio without context using paired t-test.}
\label{tab:LLMs_results_causal_relation_contexts_short}
\end{table}

%% file: sections/related_work.tex
\paragraph{Causality with LLMs}
\citet{kıcıman2023causal, ze2023causal, long2022can, feng-etal-2023-less, feng-etal-2025-causalscore, answercausalquestion2024} explore the inference of causal relations by submitting pairwise queries about variable pairs to LLMs. These queries are either structured as option selection questions \cite{kıcıman2023causal} or yes-no questions \cite{long2022can, ze2023causal}. Results from these experiments demonstrate that the LLM-based approach surpasses traditional statistical algorithms in performance. Remarkably, the LLM-based method requires only the names of the variables, without needing their statistical data. However, the approach of pairwise queries may lead to inefficiencies in time and computation, as identifying all possible relations among a set $n$ of variables necessitates $\mathit{O}(n^2)$ queries. To address this, \citet{jiralerspong2024efficient} have proposed a breadth-first search strategy that significantly reduces the number of queries to a linear scale. Additionally, beyond exploring relationships among observable variables, \citet{liu2024discovery} has developed a framework capable of uncovering high-level hidden variables from unstructured data using LLMs, and subsequently inferring causal relationships. Beyond causal discovery task, \citet{jiang2024llmcausal, cai2024knowledgelargelanguagemodels} propose leveraging LLMs to tackle broader causal tasks. They suggest using LLMs to comprehend the task, execute the appropriate algorithm, and provide intuitive interpretations of the algorithm's output.

\paragraph{Influence of Pre-training Data on Language Models.}
Research conducted by \citet{kassner-etal-2020-pretrained} and \citet{wei-etal-2021-frequency} involving controlled variations in pretraining data sheds light on its impact on language models' (LM) capabilities to memorize factual information and understand syntactic rules. Their findings confirm that the frequency of data plays a crucial role in determining a model's ability to remember specific facts or grammatical structures about verb forms. Furthermore, \citet{sinha-etal-2021-masked, min-etal-2022-rethinking} show that altering the word order during pretraining barely affects the LMs' performance in subsequent tasks, and mixing up labels in in-context learning scenarios does not significantly affect the models' few-shot learning accuracy. These studies collectively indicate that the efficacy of LMs predominantly hinges on their capacity to process complex word co-occurrence patterns. Additionally, \citet{carlini2023quantifying, 10.5555/3361338.3361358, 10.1145/3292500.3330885} have identified that LMs can retain sensitive information from their training datasets, even when such instances are infrequent. The experiments of \citet{razeghi-etal-2022-impact} demonstrate that models are more accurate on numerical reasoning questions whose terms are more prevalent in pre-training data.

%% file: sections/conclusion.tex
In this study, we investigate the factors that impact the performance of LLMs in causal discovery tasks. Our results show that the frequency of causal relations within a model's pre-training data has a positive correlation with LLM performance, while the presence of incorrect causal relations can negatively affect the models' confidence in correct causal relations. Furthermore, our experiments reveal that the context of causal relations significantly affects the validity of causal relations. 

%% file: sections/limitations.tex
One limitation of our study is its focus exclusively on LLMs for which both pre-training data and model weights are openly available. This restricts our analysis, as we are unable to extend similar investigations to LLMs that release only model weights or to entirely closed-source models. Therefore, our findings may not fully represent the behavior of all LLMs.

Furthermore, most causal relations explored in this research are related to commonsense scenarios. The real world, however, often presents more dynamic and complex causal relations that may not be adequately captured by the datasets utilized in our study.

%% file: sections/ethics_statement.tex
A key finding of our research is that LLMs tend to memorize high-frequency patterns presented in their pre-training data. While this characteristic can enhance model performance, it also poses a risk of harmful biases that are embedded in the training corpora. Such biases, when unaddressed, have the potential to lead to erroneous causal relations, which is of particular concern in critical domains like healthcare, legal systems, and public policy.

In conducting this research, we adhered to ethical guidelines ensuring that all data and models used were appropriately licensed, and any potentially identifiable information was anonymized to prevent misuse of the data and protect individual privacy.

%% file: sections/appendix.tex

\subsection{Ground-Truth Causal Graphs}
\label{apx:ground_truth_causal_graphs}
\begin{figure*}
    \centering
    \includegraphics[width=0.8\linewidth]{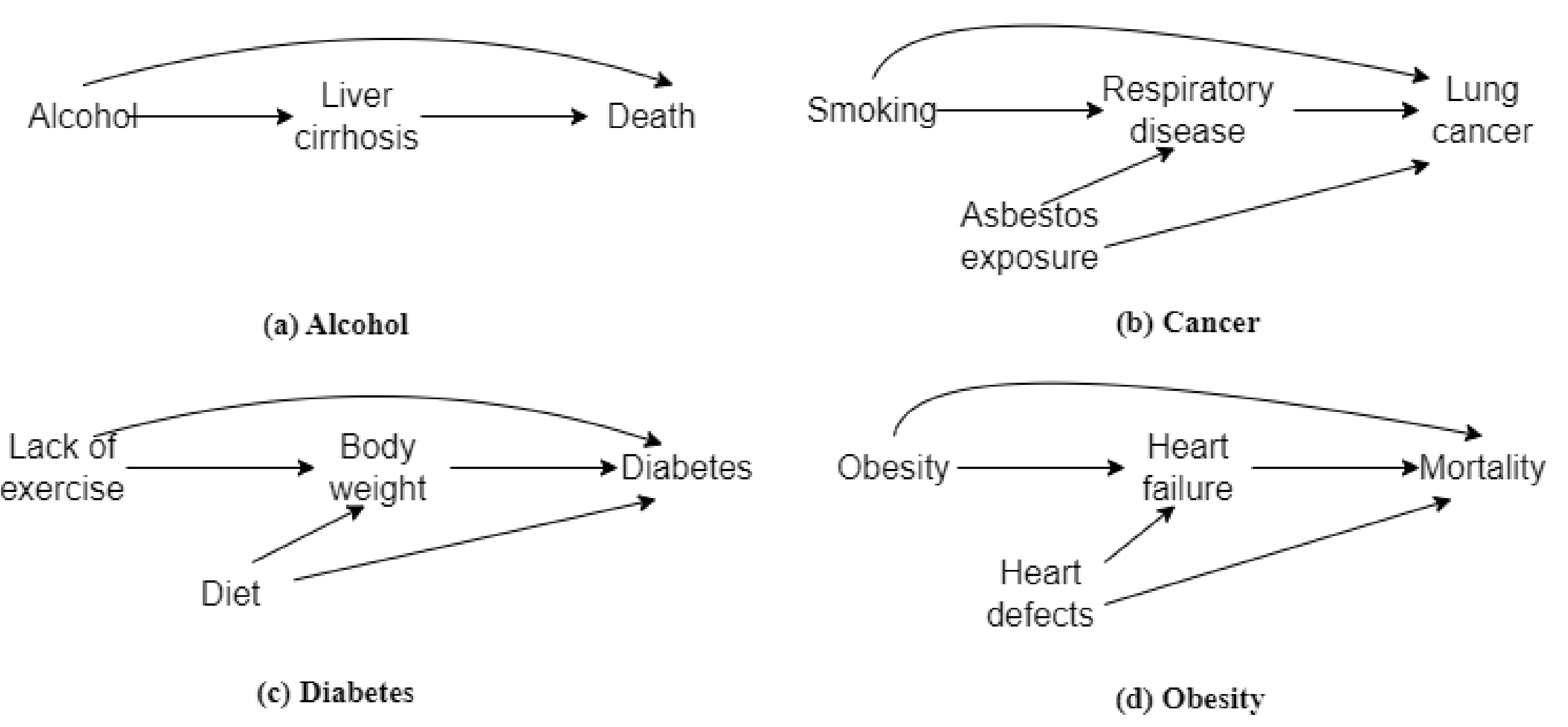}
    \caption{Four causal graphs illustrating well-known exposure-outcome effects in the medical literature. This figure is from \citet{long2022can}.}
    \label{fig:four_small_graphs}
\end{figure*}

\begin{figure}
    \centering
    \includegraphics[width=0.8\linewidth]{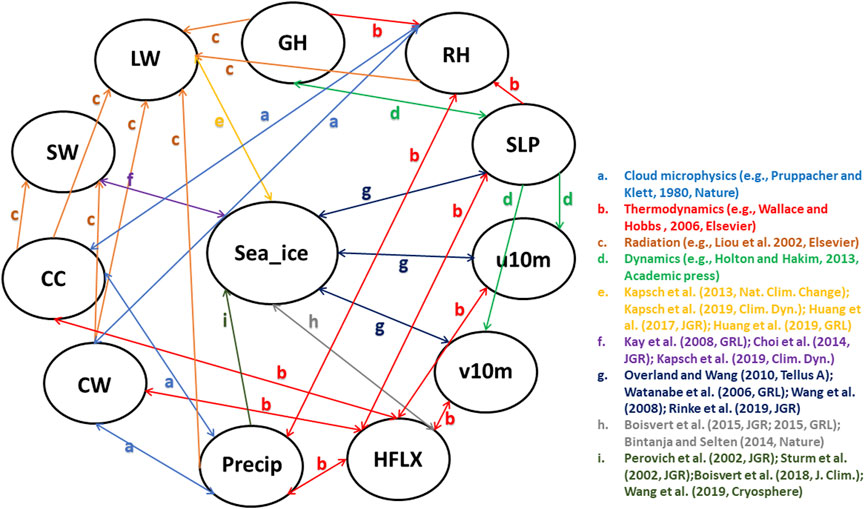}
    \caption{The causal graph between key atmospheric variables and sea ice over the Arctic based on literature review. This figure is from \citet{huang2021benchmarking}.}
    \label{fig:arctic_sea_ice}
\end{figure}

\begin{figure}
    \centering
    \includegraphics[width=0.8\linewidth]{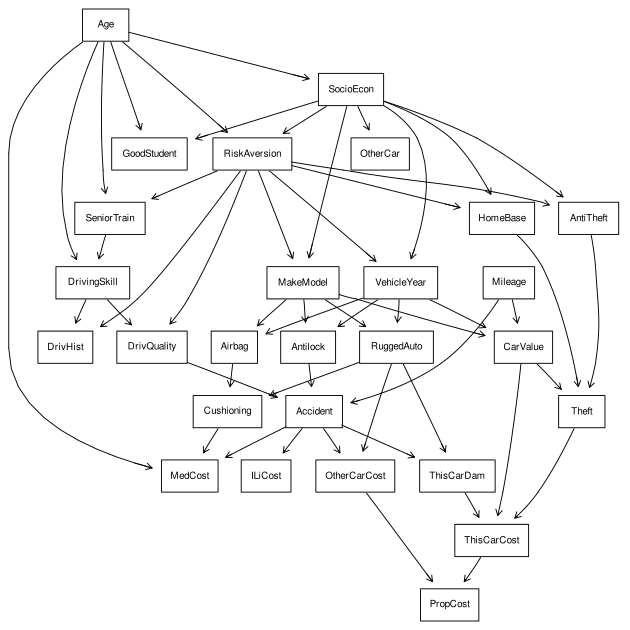}
    \caption{The causal graph for evaluating car insurance risks. This figure is sourced from  \citet{JSSv035i03}.}
    \label{fig:insurance}
\end{figure}

Figure~\ref{fig:four_small_graphs}, ~\ref{fig:arctic_sea_ice}, ~\ref{fig:insurance} demonstrate ground-truth causal graphs for the causal discovery task.

\subsection{Causal Direction Identification Task}
\label{apx:causal_direction_identification_task}
\conceptnet is a knowledge graph that connects natural language concepts via labeled edges. It includes the "[A, /r/Causes, B]" relation, indicating that event A causes event B. Each relation in \conceptnet also has a weight attribute, reflecting the confidence level of the relation; a higher weight suggests broader agreement across sources. From \conceptnet, we selected the top 1,900 causal relations by weight and generated an equal number of reverse-causal relations by swapping the cause and effect. This process yielded a total of 3,800 causal and reverse-causal relations.

\causenet is a large-scale knowledge base containing claimed causal relations between concepts. We extract 814 high-confidence causal relations from \causenet, each supported by at least 100 web sources and 10 extraction patterns. By swapping the cause and effect, we generate an equivalent number of reverse-causal relations. We then create a dataset containing 1,628 causal and reverse-causal relations.

\subsection{In-Context Learning and Prompt}
\label{apx:in_context_learning_and_prompt}
\input{tables/icl_prompt}
For the causal direction identification task and the causal discovery task, we employ similar in-context learning demonstrations and prompts, detailed in Table~\ref{tab:icl_prompt}. When presented with a pair of nodes $(A, B)$, we generate two questions: "Does A cause B?" and "Does B cause A?". 

In the causal direction identification task, the ground-truth instances are formatted as $(A\rightarrow B, true)$ and $(A\leftarrow B, false)$. These yes-no questions are directly transformed into such instances, aligning perfectly with the binary nature of the task. In the causal discovery task, the ground-truth instances are structured as $(A, B, l)$, where the label $l$ can take one of four possible values: $\leftarrow$, $\rightarrow$, $\times$, $\leftrightarrow$. Here, $\times$ denotes no causal relation, and $\leftrightarrow$ indicates a bi-directional causal relation. We include bi-directional causal relation because it exists in some ground-truth causal graphs such as Arctic Sea Ice. The conversion of yes-no responses to these four-way labels is handled as follows. If only one of the questions receives a 'yes' answer, it translates directly to the corresponding causal direction (\ie $\leftarrow$ or $\rightarrow$). If both questions are answered with 'no', this indicates no causal relation (\ie $\times$). If both questions receive a 'yes' response, this suggests a bi-directional relation (\ie $\leftrightarrow$). 

To determine the most confident answer, each LLM should generate ten distinct responses \cite{chen2023quantifying, geng2024survey}. We then extract 'yes' or 'no' from each output. If the count of 'yes' responses is greater than or equal to the count of 'no' responses, the final answer is 'yes'. If 'no' responses predominate, the final answer is 'no'. This methodology ensures a robust approach to determining causal relationships in both tasks.

The decoding hyperparameters are configured as follows: the top-p sampling parameter is set to 0.9, the repetition penalty is 1.25, the temperature is 0.8, and the maximum number of new tokens generated does not exceed the maximum input length. We employ the Hugging Face library to load LLMs and generate responses \cite{wolf-etal-2020-transformers}. All experiments were conducted on NVIDIA A100 GPUs.

\subsection{Query for Search Engine}
\label{apx:query_for_search_engine}
The queries for searching can be found in Table~\ref{tab:exact_match_query_wimbd},~\ref{tab:order_causal_query_wimbd}.
\input{tables/exact_match_query_wimbd}
\input{tables/order_causal_query_wimbd}

\subsection{Synthetic Causal Relations}
\label{apx:synthetic_causal_relations}
\input{tables/imaginary_causal_relation_templates}
Table~\ref{tab:imaginary_causal_relation_templates} demonstrates templates for creating mentions of synthetic causal relations and anti-causal relations.

\subsection{Training Details}
\label{apx:training_details}
We fine-tuning \olmo using LoRA on synthetic datasets, utilizing the official code from the OLMo repository \footnote{We employed the official OLMo code available at \url{https://github.com/allenai/open-instruct}.}. The model was trained on two NVIDIA A100 GPUs with a batch size of 2 per GPU, and a total batch size of 128. We set the LoRA rank and alpha to 256, with a dropout rate of 0.1. The learning rate was configured to 1e-4, employing a linear scheduler for rate adjustments. The training was conducted over one epoch.

\subsection{Human Evaluation for Causal Relation with Contexts}
\label{sec:human_eval_causal_relation_contexts}
\input{tables/prompt_generate_context}
The prompt of generation contexts of causal relations is shown in Table~\ref{tab:prompt_generate_context}.
In this task, we require annotators to evaluate causal relations with different contexts. 
Below we show detailed task instruction to annotators. 

\paragraph{Task Objective.}
You are provided with a series of scenarios and corresponding questions. Your task is to assess the likelihood of a causal relation based on the given scenario and give a reason for your choice. Use only the information provided in the scenario and apply common sense to make your judgment. At the begining of each evaluation, there are 10 relations without any scenarios. In these cases, we can make your judgment based on your common sense. Please review the annotation examples provided below before beginning the actual annotation task. The actual annotation tasks are performed on Google sheet. Please note that each annotator is required to complete at least one evaluation sheet.

\paragraph{Annotation Steps.}
Below is suggested annotation steps to annotators.
\begin{enumerate}
    \item 1. Read the Scenario Carefully: Each scenario provides a specific context. Understand the details and implications of the scenario.
    \item 2. Review the Question: Each question asks you to assess the likelihood of a causal relation occurring, given the provided scenario.
    \item 3. Select the Appropriate Answer: Based on your understanding of the scenario, select the probability range that best represents the likelihood of the stated causal relation occurring.
\end{enumerate}
For each question, we have below options
\begin{itemize}
    \item 100\%: The causal relation definitely occurs.
    \item 81-99\%: The causal relation almost certainly occurs.
    \item 51-80\%: The causal relation is likely to occur.
    \item 50\%: The causal relation has 50
    \item 20-49\%: The causal relation somewhat likely to occur.
    \item 1-19\%: The causal relation rarely occurs.
    \item 0\%: The causal relation never occurs.
    \item The scenario does not make sense. If the scenario contradicts common sense or could not occur in the real world or it is not a scenario at all, please select this option.
\end{itemize}

\paragraph{Annotation Examples.}
\input{tables/annotation_examples}
In Table~\ref{tab:annotation_examples}, we show some annotation examples to help annotators have a better understanding of this task.

\paragraph{Acceptance Policy.}
We will only reject a job if there is clear evidence of malicious behavior, such as random clicking, which suggests non-compliance with task guidelines.

\paragraph{Privacy Policy.}
Our primary objective is to process and publish only anonymized data. We will not publish your name, email address, or any other personal information. If you have concerns about how we handle your personal data, please contact the project manager.

\section{More Experiment Results}
\subsection{Evaluating both open- and closed-source LLMs on causal discovery tasks.}

Causal questions indicate both causal direction identification task and causal discovery task.
\citet{kıcıman2023causal,ze2023causal, feng2024imogreedylayerwisesparse, jiralerspong2024efficient} have reported that closed-source LLMs (e.g., \gptThreeFive, \gptFour) achieve state-of-the-art performance in causal direction identification task and causal discovery tasks. However, their analyses predominantly focus on specific closed-source models and offer a limited examination of open-source LLMs. In this section, we employ closed-source and open-source LLMs to conduct causal relation identification and causal discovery tasks. We aim to compare and analyze the performance disparities when utilizing different models. Table~\ref{tab:arctic_sea_ice_causal_discvoery},~\ref{tab:insurance_causal_discvoery},~\ref{tab:alcohol_causal_discvoery},~\ref{tab:cancer_causal_discvoery},~\ref{tab:diabetes_causal_discvoery} and ~\ref{tab:obesity_causal_discvoery} show the results of causal discovery experiments on the Arctic Sea Ice, Insurance, Alcohol, Cancer, Diabetes, and Obesity causal graphs. Table~\ref{tab:conceptnet_causal_direction_identification} and ~\ref{tab:causenet_causal_direction_identification} show the results of causal direction identification tasks on the \conceptnet and \causenet datasets.

We employ the Normalized Hamming Distance (NHD) as one metric for full causal discovery. A notable issue with NHD is that due to the typically sparse nature of causal graphs, models that predict no edges can still achieve a low NHD. This setup inadvertently penalizes models that predict a larger number of edges, even true edges may be predicted. To address this, following the methodologies outlined by \citet{kıcıman2023causal} and \citet{jiralerspong2024efficient}, we calculate the ratio between the NHD and the baseline NHD of a model that outputs the same number of edges but with all of them being incorrect. The lower the ratio, the better the model performs compared to the worst baseline that outputs the same number of edges. Therefore, we report NHD ratio (\ie NHD / baseline NHD), along with the number of predicted edges, to provide a more comprehensive evaluation of model performance in the full causal discovery task.

Due to the transparency of \olmo and the robust capabilities of its search tool, \olmo serves as our primary analysis model. Therefore,  we explored various numbers of in-context learning examples to identify the optimal example number. In seven out of eight datasets, \olmo with three demonstration examples achieves the highest F1, compared to other numbers of demonstration examples tested. Therefore, to ensure a fair comparison, other LLMs also utilized three demonstration examples for in-context learning.

Considering all LLMs, \gptFouro outperforms others in six of the eight datasets evaluated, specifically Arctic Sea Ice, Insurance, Alcohol, Obesity, \conceptnet, and \causenet. In the remaining two datasets, Cancer and Diabetes, \gptFouro ranks as the second-best model, with only a slight performance differential from the top model. These experiment results show that \gptFouro is the most effective model for causal discovery and causal direction identification tasks in both closed- and open-source models. Among open-source models exclusively, \llamaThree excels, achieving the highest F1 scores in six datasets: Insurance, Alcohol, Cancer, Diabetes, Obesity, and \causenet. Meanwhile, \llamaTwo achieves the highest F1 in two datasets, Arctic Sea Ice and Obesity. In the \conceptnet dataset, \olmo, configured with three in-context learning examples, records the best F1 score.

\input{tables/arctic_sea_ice_causal_discvoery}
\input{tables/insurance_causal_discvoery}
\input{tables/alcohol_causal_discvoery}
\input{tables/cancer_causal_discvoery}
\input{tables/diabetes_causal_discvoery}
\input{tables/obesity_causal_discvoery}
\input{tables/conceptnet_causal_direction_identification}
\input{tables/causenet_causal_direction_identification}

\subsection{Do pre-training corpora contain more correct causal relations?}
\label{apx:causal_relations_more_frequent}
\input{tables/total_occur_short}
Given the effective performance of LLMs on causal discovery tasks, a pertinent research question arises: Why can LLMs perform so well? We posit that a significant factor is the nature of the pre-training data, which contains more correct causal relations than incorrect ones, leading LLMs to primarily memorize correct causal relations.

\begin{RQ}
\label{rq:more_correct_causal_corpora}
Do pre-training corpora contain more correct causal relations than incorrect ones?
\end{RQ}

Humans fundamentally rely on causal relations to understand and generate text. Therefore, it is reasonable that pre-training corpora, which are collected from human-generated texts, are likely to inherently contain a higher proportion of correct causal relations.

\noindent \underline{\textit{Observation}} We count the total occurrence of correct and incorrect causal relations in \dolma and \roots corpora. The results are shown in Table~\ref{tab:total_occur_short}. We use exact matching to count correct and incorrect causal relations. We observe that the occurrence of causal relations is, on average, 12 times higher than that of incorrect causal relations in \dolma and \roots corpora. From our observation, most incorrect causal relations do not exist in an affirmation context. They are usually in a question or negation context. For example, "Which option is correct? A. smoking causes cancer B. cancer causes smoking" or "Which means that either smoking causes cancer or cancer causes smoking. " 

\noindent \underline{\textit{Discussion}} In conclusion, these experimental results show that correct causal relations are more frequently represented than incorrect ones in pre-training corpora. This also explain why LLMs can identify many causal relations in causal discovery tasks.


\subsection{Influence of Model Size on LLMs' Performance in Causal Discovery Tasks}

\begin{RQ}
\label{rq:larger_model_better}
Do larger models perform better on causal discovery tasks?
\end{RQ}

We assume that within the same architectural framework, increasing the model size (i.e., the number of parameters) leads to improved performance on causal discovery tasks. The rationale is that larger models can memorize more information from the pre-training data than their smaller models.

\noindent \underline{\textit{Observation}} We select models from the Llama2 and Llama3 series, each varying in size. These models are evaluated on causal discovery and causal direction identification tasks, with results documented in Table~\ref{tab:causal_discovery_model_different_sizes} and ~\ref{tab:causal_direction_model_different_sizes}. The findings indicate that for both the Llama2 and Llama3 models, there is a positive correlation between the number of parameters and performance. However, discrepancies arise when comparing across architectures. For example, a small Llama3 model (\eg \llamaThree) can outperform a significantly larger Llama3 model (\eg \llamaTwoXX). Notably, across most datasets, \llamaThreeXX either matches or surpasses the performance of the currently leading closed-source LLM, \gptFouro.

\noindent \underline{\textit{Discussion}} The experiment results lead to a critical consideration of the 'bigger is better' paradigm in LLM research.  Future research should thus not only focus on scaling up the size but also on refining the architecture and learning algorithms to better leverage increased model capacity.

\input{tables/causal_discovery_model_different_sizes}
\input{tables/causal_direction_model_different_sizes}

%% file: tables/icl_prompt.tex
\begin{table*}[h]
\centering
\resizebox{\textwidth}{!}{%
\begin{tabular}{ll}
\hline
\multicolumn{2}{l}{\textbf{Demonstrations for in-context learning}} \\ \hline
User: & \begin{tabular}[c]{@{}l@{}}This task is to determine the cause-and-effect relationship between two events based on commonsense knowledge. \\ We are interested in the causal relationship between 'it is raining' and 'carrying an umbrella'. \\ Does 'it is raining' cause 'carrying an umbrella'? \\ Let's provide a step-by-step explanation, then give your final answer using yes or no.\end{tabular} \\
Assistant: & \begin{tabular}[c]{@{}l@{}}Step-by-Step Explanation: \\ 1. Observation: The presence of rain often results in individuals using umbrellas. \\ 2. Rationale: People use umbrellas to protect themselves from getting wet during rain. \\ 3. Conclusion: Rain acts as a cause for the use of an umbrella. The final answer is Yes. \\ Final Answer: Yes\end{tabular} \\
User: & \begin{tabular}[c]{@{}l@{}}This task is to determine the cause-and-effect relationship between two events based on commonsense knowledge. \\ We are interested in the causal relationship between 'it is raining' and 'carrying an umbrella'. \\ Does 'carrying an umbrella' causes 'it is raining'? \\ Let's provide a step-by-step explanation, then give your final answer using yes or no.\end{tabular} \\
Assistant: & \begin{tabular}[c]{@{}l@{}}Step-by-Step Explanation:\\ 1. Observation: The act of opening an umbrella does not influence weather patterns or conditions. \\ 2. Rationale: Rain is a meteorological event determined by atmospheric conditions. The use of an umbrella does not have the capability to influence weather conditions. \\ 3. Conclusion: There is no causal relationship where the umbrella causes rain. The answer is No. \\ Final Answer: No\end{tabular} \\
User: & \begin{tabular}[c]{@{}l@{}}This task is to determine the cause-and-effect relationship between two events based on commonsense knowledge. \\ We are interested in the causal relationship between 'Eating food' and 'feeling Hunger'. \\ Does 'Eating food' causes 'feeling Hunger'? \\ Let's provide a step-by-step explanation, then give your final answer using yes or no.\end{tabular} \\
Assistant: & \begin{tabular}[c]{@{}l@{}}Step-by-Step Explanation: \\ 1. Observation: Eating involves consuming food, which typically reduces the feeling of hunger. \\ 2. Rationale: The act of eating is intended to satiate hunger, not induce it. \\ 3. Conclusion: Eating does not cause hunger. Instead, it alleviates it. The answer is No. \\ Final Answer: No\end{tabular} \\
User: & \begin{tabular}[c]{@{}l@{}}This task is to determine the cause-and-effect relationship between two events based on commonsense knowledge. \\ We are interested in the causal relationship between 'Eating food' and 'feeling Hunger'. \\ Does 'Feeling hunger' causes 'Eating food'? \\ Let's provide a step-by-step explanation, then give your final answer using yes or no.\end{tabular} \\
Assistant: & \begin{tabular}[c]{@{}l@{}}Step-by-Step Explanation: \\ 1. Observation: The sensation of hunger typically motivates individuals to eat. \\ 2. Rationale: Hunger is a biological signal that the body needs nutrients. This sensation prompts the behavior of eating to replenish energy and nutrients. \\ 3. Conclusion: It is reasonable to conclude that hunger causes eating. The answer is Yes. \\ Final Answer: Yes\end{tabular} \\ \hline
\multicolumn{2}{l}{\textbf{Prompt}} \\ \hline
User: & \begin{tabular}[c]{@{}l@{}}This task is to determine the cause-and-effect relationship between two events based on commonsense knowledge. \\ We are interested in the causal relationship between \{cause\}' and '\{effect\}'. \\ Does '\{cause\}' cause '\{effect\}'?\\ Let's provide a step-by-step explanation, then give your final answer using yes or no.\end{tabular} \\ \hline
\end{tabular}%
}
\caption{Demonstrations for in-context learning and the prompt for new input.}
\label{tab:icl_prompt}
\end{table*}

%% file: tables/exact_match_query_wimbd.tex
\begin{table*}[h]
\centering
\resizebox{\textwidth}{!}{%
\begin{tabular}{l}
\hline
\textbf{Exact match for "event A causes event B"}                                                                                                                                                                                                                                                                                                                                                                                                                                                                                                                                                                                                                                                                                                                                                                                                                                                                                                                                                                                                                                                                                                                                                                                                                                                                                                                                                                                                                                                                                             \\ \hline
\begin{tabular}[c]{@{}l@{}}templates = {[}f"\{cause\} causes \{effect\}", f"\{effect\} is caused by \{cause\}", f"\{cause\} leads to \{effect\}",\\                  f"\{cause\} results in \{effect\}", f"\{cause\} triggers \{effect\}", f"\{effect\} is triggered by \{cause\}",\\                  f"\{cause\} induces \{effect\}", f"\{cause\} influences \{effect\}", f"\{effect\} is influenced by \{cause\}",\\                  f"\{cause\} affects \{effect\}", f"\{effect\} is affected by \{cause\}", f"\{cause\} impacts \{effect\}",\\                  f"\{cause\} is impacted by \{effect\}", f"\{cause\} is responsible for \{effect\}",\\                  f"\{cause\} is the reason for \{effect\}", f"The effect of \{cause\} is \{effect\}",\\                  f"The result of \{cause\} is \{effect\}", f"The consequence of \{cause\} is \{effect\}",\\                  f"\{effect\} is a consequence of \{cause\}", f"\{effect\} is a result of \{cause\}", f"\{effect\} is an effect of \{cause\}"{]}\\ \\ \# create match\_phrase query for each template\\ should\_list = {[}{]}\\ for phrase in templates:\\     match\_phrase = \{\\         "match\_phrase": \{\\             "text": \{\\                 "query": phrase,\\                 "slop": int(len(phrase.split())*0.25), \\             \}\\         \}\\     \}\\ should\_list.append(match\_phrase)\\ query = \{\\         "bool": \{\\             "should": should\_list,\\             "minimum\_should\_match": 1\\         \}\end{tabular} \\ \hline
\end{tabular}%
}
\caption{Exact match query for WIMBD.}
\label{tab:exact_match_query_wimbd}
\end{table*}

%% file: tables/order_causal_query_wimbd.tex
\begin{table*}[h]
\centering
\resizebox{\textwidth}{!}{%
\begin{tabular}{l}
\hline
\textbf{Ordered phrase search for "event A" $\Rightarrow$ "causes" $\Rightarrow$ "event B"} \\ \hline
\begin{tabular}[c]{@{}l@{}}causal\_mentions = {[}"causes", "leads to", "results in", "triggers", "induces", "influences", "affects", "impacts",\\                        "is responsible for", "is the reason for", "cause", "lead to", "result in", "trigger", "induce",\\                        "influence", "affect", "impact", "are responsible for", "are the reason for"{]}\\ \\ \# create cause clause in span term format\\ cause\_clauses = {[}{]}\\ for item in cause.split():\\     cause\_clauses.append(\{"span\_term": \{"text": item\}\})\\ \\ \# create effect clause in span term format\\ effect\_clauses = {[}{]}\\ for item in effect.split():\\     effect\_clauses.append(\{"span\_term": \{"text": item\}\})\\ \\ \# create causal relation clause in span term format\\ all\_relation\_clauses = {[}{]}\\ for rel in causal\_mentions:\\     relation\_clauses = {[}{]}\\     for term in rel.split():\\         relation\_clauses.append(\{"span\_term": \{"text": term\}\})\\     all\_relation\_clauses.append(relation\_clauses)\\ \\ \# for each causal relation clause, create a query\\ for relation\_clauses in all\_relation\_clauses:\\     query = \{\\         "span\_near": \{\\             "clauses": {[}\\                 \{\\                     "span\_near": \{\\                         "clauses": cause\_clauses,\\                         "slop": 0,\\                         "in\_order": True\\                     \}\\                 \},\\                 \{\\                     "span\_near": \{\\                         "clauses": relation\_clauses,\\                         "slop": 0,\\                         "in\_order": True\\                     \}\\                 \},\\                 \{\\                     "span\_near": \{\\                         "clauses": effect\_clauses,\\                         "slop": 0,\\                         "in\_order": True\\                     \}\\                 \}\\             {]},\\             "slop": 32, \# window size\\             "in\_order": True\\         \}\\     \}\end{tabular} \\ \hline
\end{tabular}%
}
\caption{"event A" $\Rightarrow$ "causes" $\Rightarrow$ "event B" query for WIMBD.}
\label{tab:order_causal_query_wimbd}
\end{table*}

%% file: tables/imaginary_causal_relation_templates.tex
\begin{table*}[th]
\centering
\resizebox{\textwidth}{!}{%
\begin{tabular}{lll}
\hline
Correct causal relations                                                                                                                                                                                                                                                                                                                                                                                                                                                                                                                                                                                                                                                                                                                                                                                                                                                                                                                                                                                                                                                                            & Reverse causal relations                                                                                                                                                                                                                                                                                                                                                                                                                                                                                                                                                                                                                                                                                                                                                                                                                                                                                                                                                                                                                                                                            & Negation of causal relations                                                                                                                                                                                                                                                                                                                                                                                                                                                                                                                                                                                                                                                                                                                                                                                                                                                                                                                                                                                                                                                                                                                                                                                                                                                                                                         \\
\begin{tabular}[c]{@{}l@{}}templates = [f"{cause} causes {effect}.", \\ f"{effect} is caused by {cause}.", \\ f"{cause} leads to {effect}.",\\                          f"{cause} results in {effect}.", \\ f"{cause} triggers {effect}.", \\ f"{effect} is triggered by {cause}.",\\                          f"{cause} induces {effect}.", \\ f"{cause} influences {effect}.", \\ f"{effect} is influenced by {cause}.",\\                          f"{cause} affects {effect}.", \\ f"{effect} is affected by {cause}.", \\ f"{cause} impacts {effect}.",\\                          f"{cause} is impacted by {effect}.", \\ f"{cause} is responsible for {effect}.",\\                          f"{cause} is the reason for {effect}.", \\ f"The effect of {cause} is {effect}.",\\                          f"The result of {cause} is {effect}.", \\ f"The consequence of {cause} is {effect}.",\\                          f"{effect} is a consequence of {cause}.", \\ f"{effect} is a result of {cause}.",\\                          f"{effect} is an effect of {cause}.", ]\end{tabular} & \begin{tabular}[c]{@{}l@{}}templates = [f"{effect} causes {cause}.", \\ f"{cause} is caused by {effect}.", \\ f"{effect} leads to {cause}.",\\                          f"{effect} results in {cause}.", \\ f"{effect} triggers {cause}.", \\ f"{cause} is triggered by {effect}.",\\                          f"{effect} induces {cause}.", \\ f"{effect} influences {cause}.", \\ f"{cause} is influenced by {effect}.",\\                          f"{effect} affects {cause}.", \\ f"{cause} is affected by {effect}.", \\ f"{effect} impacts {cause}.",\\                          f"{effect} is impacted by {cause}.", \\ f"{effect} is responsible for {cause}.",\\                          f"{effect} is the reason for {cause}.", \\ f"The effect of {effect} is {cause}.",\\                          f"The result of {effect} is {cause}.", \\ f"The consequence of {effect} is {cause}.",\\                          f"{cause} is a consequence of {effect}.", \\ f"{cause} is a result of {effect}.",\\                          f"{cause} is an effect of {effect}.", ]\end{tabular} & \begin{tabular}[c]{@{}l@{}}templates = [  f"{cause} does not cause {effect}.",\\                 f"{effect} is not caused by {cause}.",\\                 f"{cause} does not lead to {effect}.",\\                 f"{cause} does not result in {effect}.",\\                 f"{cause} does not trigger {effect}.",\\                 f"{effect} is not triggered by {cause}.",\\                 f"{cause} does not induce {effect}.",\\                 f"{cause} does not influence {effect}.",\\                 f"{effect} is not influenced by {cause}.",\\                 f"{cause} does not affect {effect}.",\\                 f"{effect} is not affected by {cause}.",\\                 f"{cause} does not impact {effect}.",\\                 f"{cause} is not impacted by {effect}.",\\                 f"{cause} is not responsible for {effect}.",\\                 f"{cause} is not the reason for {effect}.",\\                 f"The effect of {cause} is not {effect}.",\\                 f"The result of {cause} is not {effect}.",\\                 f"The consequence of {cause} is not {effect}.",\\                 f"{effect} is not a consequence of {cause}.",\\                 f"{effect} is not a result of {cause}.",\\                 f"{effect} is not an effect of {cause}.",]\end{tabular} \\ \hline
\end{tabular}%
}
\caption{Templates for creating mentions of imaginary causal relations and anti-causal relations.}
\label{tab:imaginary_causal_relation_templates}
\end{table*}

%% file: tables/prompt_generate_context.tex
\begin{table}[ht]
\centering
\resizebox{\textwidth}{!}{%
\begin{tabular}{l}
\hline
Prompt for generating contexts of causal relations                                                                                                                                                                                                                                                                                                                                                                                                                                      \\ \hline
\begin{tabular}[c]{@{}l@{}}List five scenarios where '{cause}' causes '{effect}'. Then, list five scenarios where '{cause}' does not cause '{effect}'.\\ Each scenario should be distinctly and clearly described, categorized under the respective headings.\\ Response Format:\\ Scenarios where '{cause}' causes '{effect}':\\ Heading:\\ Description:\\ Reason:\\ ...\\ Scenarios where '{cause}' does not cause '{effect}':\\ Heading:\\ Description:\\ Reason:\\ ...\end{tabular} \\ \hline
\end{tabular}%
}
\caption{Prompt for generating contexts of causal relations.}
\label{tab:prompt_generate_context}
\end{table}

%% file: tables/annotation_examples.tex
\begin{table*}[h]
\centering
\resizebox{\textwidth}{!}{%
\begin{tabular}{l}
\hline
\textbf{Annotation examples}                                                                                                                                                                                                                                                    \\ \hline
\#\#\#relation--91\#\#\#                                                                                                                                                                                                                                                        \\
Question: to what extent do you think 'soaking in a hotspring' causes 'relaxation'?                                                                                                                                                                                             \\
81-99\%                                                                                                                                                                                                                                                                         \\
\begin{tabular}[c]{@{}l@{}}Reason: The warm water of a hot spring helps to raise the body's temperature, \\ which can relax muscle tension and soothe aches and pains in the joints and muscles. \\ This physical relaxation naturally leads to mental relaxation.\end{tabular} \\ \hline
\#\#\#relation--91\#\#\#                                                                                                                                                                                                                                                        \\
Scenario --- High water temperature:                                                                                                                                                                                                                                            \\
The water temperature in the hotspring is excessively hot, making the individual feel uncomfortable.                                                                                                                                                                            \\
                                                                                                                                                                                                                                                                                \\
Answer the following question ONLY based on information described in above scenario and your common sense.                                                                                                                                                                      \\
Question: under above scenario, to what extent do you think 'soaking in a hotspring' causes 'relaxation'?                                                                                                                                                                       \\
1-19\%                                                                                                                                                                                                                                                                          \\
Reason: Uncomfortably high temperatures can cause overheating, dizziness, or discomfort, preventing relaxation.                                                                                                                                                                 \\ \hline
\#\#\#relation--96\#\#\#                                                                                                                                                                                                                                                        \\
Scenario --- Entertaining Friends:                                                                                                                                                                                                                                              \\
During a casual get-together with friends, you crack jokes and everyone bursts into laughter.                                                                                                                                                                                   \\
                                                                                                                                                                                                                                                                                \\
Answer the following question ONLY based on information described in above scenario and your common sense.                                                                                                                                                                      \\
Question: under above scenario, to what extent do you think 'making people laugh' causes 'you have fun too'?                                                                                                                                                                    \\
81-99\%                                                                                                                                                                                                                                                                         \\
Reason: The shared joy and camaraderie among friends create a fun and enjoyable atmosphere.                                                                                                                                                                                     \\ \hline
\end{tabular}%
}
\caption{Examples of causal relation evaluation under different contexts.}
\label{tab:annotation_examples}
\end{table*}

%% file: tables/arctic_sea_ice_causal_discvoery.tex
\begin{table*}[h]
\centering
\resizebox{\textwidth}{!}{%
\begin{tabular}{lllllllll}
\hline
                       & \textbf{Precision↑} & \textbf{Recall↑} & \textbf{F1↑}    & \textbf{Accuracy↑} & \textbf{Predict edges} & \textbf{NHD↓} & \textbf{Baseline NHD} & \textbf{Ratio (NHD/Baseline NHD)↓} \\ \hline
\olmo (0 ICL)          & 0.4259              & 0.5              & 0.46            & 0.625              & 54                          & 0.375         & 0.6944                & 0.54                                   \\
\olmo (1 ICL)          & 0.3928              & 0.4782           & 0.4314          & 0.5972             & 56                          & 0.4027        & 0.7083                & 0.5686                                 \\
\olmo (2 ICL)          & 0.4615              & 0.1304           & 0.2034          & 0.6736             & 13                          & 0.3263        & 0.4097                & 0.7966                                 \\
\olmo (3 ICL)          & 0.5555              & 0.1087           & 0.1818          & 0.6875             & 9                           & 0.3125        & 0.3819                & 0.8181                                 \\
\olmo (4 ICL)          & 0.5417              & 0.2826           & 0.3714          & 0.6944             & 24                          & 0.3055        & 0.4861                & 0.6285                                 \\
\bloom (3 ICL)         & 0.3934              & 0.5217           & 0.4485          & 0.5902             & 61                          & 0.4097        & 0.7430                & 0.5514                                 \\
\llamaTwo (3 ICL)      & 0.4444              & 0.5217           & {\ul 0.48}      & 0.6388             & 54                          & 0.3611        & 0.6944                & {\ul 0.52}                             \\
\llamaThree (3 ICL)    & 1.0                 & 0.1956           & 0.3272          & 0.7430             & 9                           & 0.2569        & 0.3819                & 0.6727                                 \\
\gptThreeFive  (3 ICL) & 0.7647              & 0.2826           & 0.4126          & 0.7431             & 17                          & 0.2569        & 0.4375                & 0.5873                                 \\
\gptFouro  (3 ICL)     & 0.5178              & 0.6304           & \textbf{0.5686} & 0.6944             & 56                          & 0.3055        & 0.7083                & \textbf{0.4313}                        \\ \hline
\end{tabular}%
}
\caption{Causal discovery results for the Arctic Sea Ice causal graph, with 12 nodes and 46 edges. \gptFouro surpasses all competing models, achieving an F1 score of 0.5686 and an NHD ratio of 0.4313. The second-best performing model is an open-source LLM, \llamaTwo. (\# ICL) indicates the number of demonstration examples for in-context learning.}
\label{tab:arctic_sea_ice_causal_discvoery}
\end{table*}

%% file: tables/insurance_causal_discvoery.tex
\begin{table*}[h]
\centering
\resizebox{\textwidth}{!}{%
\begin{tabular}{lllllllll}
\hline
                       & \textbf{Precision↑} & \textbf{Recall↑} & \textbf{F1↑}    & \textbf{Accuracy↑} & \textbf{Predict edges} & \textbf{NHD↓} & \textbf{Baseline NHD} & \textbf{Ratio (NHD/Baseline NHD)↓} \\ \hline
\olmo (0 ICL)          & 0.0873              & 0.7692           & 0.1568          & 0.4101             & 458                         & 0.5898        & 0.6995                & 0.8431                             \\
\olmo (1 ICL)          & 0.0963              & 0.9038           & 0.1740          & 0.3882             & 488                         & 0.6117        & 0.7407                & 0.8259                             \\
\olmo (2 ICL)          & 0.0901              & 0.5961           & 0.1565          & 0.5418             & 344                         & 0.4581        & 0.5432                & 0.8434                             \\
\olmo (3 ICL)          & 0.1254              & 0.6731           & 0.2114          & 0.6419             & 279                         & 0.3580        & 0.4540                & 0.7885                             \\
\olmo (4 ICL)          & 0.1093              & 0.7884           & 0.1920          & 0.5267             & 375                         & 0.4732        & 0.5857                & 0.8079                             \\
\bloom (3 ICL)         & 0.0710              & 0.7115           & 0.1291          & 0.3155             & 521                         & 0.6844        & 0.7860                & 0.8708                             \\
\llamaTwo (3 ICL)      & 0.1245              & 0.7115           & 0.2120          & 0.6227             & 297                         & 0.3772        & 0.4787                & 0.7879                             \\
\llamaThree (3 ICL)    & 0.2656              & 0.3269           & {\ul 0.2931}    & 0.8875             & 64                          & 0.1124        & 0.1591                & {\ul 0.7069}                       \\
\gptThreeFive  (3 ICL) & 0.1575              & 0.5              & 0.2396          & 0.7736             & 165                         & 0.2263        & 0.2976                & 0.7603                             \\
\gptFouro  (3 ICL)     & 0.2287              & 0.6730           & \textbf{0.3414} & 0.8148             & 153                         & 0.1851        & 0.2812                & \textbf{0.6585}                    \\ \hline
\end{tabular}%
}
\caption{Causal discovery results for the Insurance causal graph, with 27 nodes and 52 edges. \gptFouro surpasses all competing models, achieving an F1 score of 0.3414 and an NHD ratio of 0.6585. The second-best performing model is an open-source LLM, \llamaThree. }
\label{tab:insurance_causal_discvoery}
\end{table*}

%% file: tables/alcohol_causal_discvoery.tex
\begin{table*}[h]
\centering
\resizebox{\textwidth}{!}{%
\begin{tabular}{lllllllll}
\hline
                       & \textbf{Precision↑} & \textbf{Recall↑} & \textbf{F1↑} & \textbf{Accuracy↑} & \textbf{Predict edges} & \textbf{NHD↓} & \textbf{Baseline NHD} & \textbf{Ratio (NHD/Baseline NHD)↓} \\ \hline
\olmo (0 ICL)          & 0.5                 & 1.0              & 0.6667       & 0.6667             & 6                           & 0.3333        & 1.0                   & 0.3333                             \\
\olmo (1 ICL)          & 0.6                 & 1.0              & 0.75         & 0.7778             & 5                           & 0.2222        & 0.8889                & 0.25                               \\
\olmo (2 ICL)          & 0.5                 & 1.0              & 0.6667       & 0.6667             & 6                           & 0.3333        & 1.0                   & 0.3333                             \\
\olmo (3 ICL)          & 0.6                 & 1.0              & 0.75         & 0.7778             & 5                           & 0.2222        & 0.8889                & 0.25                               \\
\olmo (4 ICL)          & 0.6                 & 1.0              & 0.75         & 0.7778             & 5                           & 0.2222        & 0.8889                & 0.25                               \\
\bloom (3 ICL)         & 0.5                 & 1.0              & 0.6667       & 0.6667             & 6                           & 0.3333        & 1.0                   & 0.3333                             \\
\llamaTwo (3 ICL)      & 0.75                & 1.0              & {\ul 0.8571} & 0.8889             & 4                           & 0.1111        & 0.7778                & {\ul 0.1429}                       \\
\llamaThree (3 ICL)    & 1.0                 & 1.0              & \textbf{1.0} & 1.0                & 3                           & 0             & 0.6667                & \textbf{0}                         \\
\gptThreeFive  (3 ICL) & 1.0                 & 1.0              & \textbf{1.0} & 1.0                & 3                           & 0             & 0.6667                & \textbf{0}                         \\
\gptFouro  (3 ICL)     & 1.0                 & 1.0              & \textbf{1.0} & 1.0                & 3                           & 0             & 0.6667                & \textbf{0}                         \\ \hline
\end{tabular}%
}
\caption{Causal discovery results for the Alcohol causal graph, with 3 nodes and 3 edges. \llamaThree, \gptThreeFive, and \gptFour accurately predict the ground-truth causal graph. The second-best performing model is \llamaTwo. }
\label{tab:alcohol_causal_discvoery}
\end{table*}

%% file: tables/cancer_causal_discvoery.tex
\begin{table*}[h]
\centering
\resizebox{\textwidth}{!}{%
\begin{tabular}{lllllllll}
\hline
                       & \textbf{Precision↑} & \textbf{Recall↑} & \textbf{F1↑}    & \textbf{Accuracy↑} & \textbf{Predict edges} & \textbf{NHD↓} & \textbf{Baseline NHD} & \textbf{Ratio (NHD/Baseline NHD)↓} \\ \hline
\olmo (0 ICL)          & 0.4166              & 1.0              & 0.5882          & 0.5625             & 12                     & 0.4375        & 1.0                   & 0.4375                             \\
\olmo (1 ICL)          & 0.4                 & 0.8              & 0.5333          & 0.5625             & 10                     & 0.4375        & 0.9375                & 0.4667                             \\
\olmo (2 ICL)          & 0.5                 & 0.8              & 0.6153          & 0.6875             & 8                      & 0.3125        & 0.8125                & 0.3846                             \\
\olmo (3 ICL)          & 0.5714              & 0.8              & 0.6667          & 0.75               & 7                      & 0.3125        & 0.9375                & 0.3333                             \\
\olmo (4 ICL)          & 0.5                 & 1.0              & 0.6667          & 0.6875             & 10                     & 0.3125        & 0.9375                & 0.3333                             \\
\bloom (3 ICL)         & 0.4                 & 0.4              & 0.4             & 0.625              & 5                      & 0.375         & 0.625                 & 0.6                                \\
\llamaTwo (3 ICL)      & 0.4166              & 1.0              & 0.5882          & 0.5625             & 12                     & 0.4375        & 1.0                   & 0.4375                             \\
\llamaThree (3 ICL)    & 1.0                 & 0.8              & \textbf{0.8889} & 0.9375             & 4                      & 0.0625        & 0.5625                & \textbf{0.1111}                    \\
\gptThreeFive  (3 ICL) & 1.0                 & 0.8              & \textbf{0.8889} & 0.9375             & 4                      & 0.0625        & 0.5625                & \textbf{0.1111}                    \\
\gptFouro  (3 ICL)     & 0.8                 & 0.8              & {\ul 0.8}       & 0.875              & 5                      & 0.125         & 0.625                 & {\ul 0.2}                          \\ \hline
\end{tabular}%
}
\caption{Causal discovery results for the Cancer causal graph, with 4 nodes and 5 edges. \llamaThree and \gptThreeFive surpass all other models. The second-best performing model is \gptFouro.}
\label{tab:cancer_causal_discvoery}
\end{table*}

%% file: tables/diabetes_causal_discvoery.tex
\begin{table*}[h]
\centering
\resizebox{\textwidth}{!}{%
\begin{tabular}{lllllllll}
\hline
                       & \textbf{Precision↑} & \textbf{Recall↑} & \textbf{F1↑} & \textbf{Accuracy↑} & \textbf{Predict edges} & \textbf{NHD↓} & \textbf{Baseline NHD} & \textbf{Ratio (NHD/Baseline NHD)↓} \\ \hline
\olmo (0 ICL)          & 0.4166              & 1.0              & 0.5882       & 0.5625             & 12                          & 0.4375        & 1.0625                & 0.4117                             \\
\olmo (1 ICL)          & 0.4166              & 1.0              & 0.5882       & 0.5625             & 12                          & 0.4375        & 1.0625                & 0.4117                             \\
\olmo (2 ICL)          & 0.4166              & 1.0              & 0.5882       & 0.5625             & 12                          & 0.4375        & 1.0625                & 0.4117                             \\
\olmo (3 ICL)          & 0.5                 & 1.0              & 0.6666       & 0.6875             & 10                          & 0.3125        & 0.9375                & 0.3333                             \\
\olmo (4 ICL)          & 0.4545              & 1.0              & 0.625        & 0.625              & 11                          & 0.375         & 1.0                   & 0.375                              \\
\bloom (3 ICL)         & 0.4285              & 0.6              & 0.5          & 0.625              & 7                           & 0.375         & 0.75                  & 0.5                                \\
\llamaTwo (3 ICL)      & 0.5556              & 1.0              & 0.7142       & 0.75               & 9                           & 0.25          & 0.875                 & 0.2857                             \\
\llamaThree (3 ICL)    & 1.0                 & 0.8              & 0.8889       & 0.9375             & 4                           & 0.0625        & 0.5625                & 0.1111                             \\
\gptThreeFive  (3 ICL) & 1.0                 & 1.0              & \textbf{1.0} & 1.0                & 5                           & 0             & 0.625                 & \textbf{0}                         \\
\gptFouro  (3 ICL)     & 0.8333              & 1.0              & {\ul 0.9091} & 0.9375             & 6                           & 0.0625        & 0.6875                & {\ul 0.0909}                       \\ \hline
\end{tabular}%
}
\caption{Causal discovery results for the Diabetes causal graph, with 4 nodes and 5 edges. \gptThreeFive accurately predict the ground-truth causal graph. The second-best performing model is \gptFouro.}
\label{tab:diabetes_causal_discvoery}
\end{table*}

%% file: tables/obesity_causal_discvoery.tex
\begin{table*}[h]
\centering
\resizebox{\textwidth}{!}{%
\begin{tabular}{lllllllll}
\hline
                       & \textbf{Precision↑} & \textbf{Recall↑} & \textbf{F1↑}    & \textbf{Accuracy↑} & \textbf{Predict edges (46)} & \textbf{NHD↓} & \textbf{Baseline NHD} & \textbf{Ratio (NHD/Baseline NHD)↓} \\ \hline
\olmo (0 ICL)          & 0.5714              & 0.8              & 0.6666          & 0.75               & 7                           & 0.3125        & 0.9375                & 0.3333                             \\
\olmo (1 ICL)          & 0.5                 & 1.0              & 0.6666          & 0.6875             & 10                          & 0.3125        & 0.9375                & 0.3333                             \\
\olmo (2 ICL)          & 0.5555              & 1.0              & 0.7142          & 0.75               & 9                           & 0.25          & 0.875                 & 0.2857                             \\
\olmo (3 ICL)          & 0.8                 & 0.8              & {\ul 0.8}       & 0.875              & 5                           & 0.125         & 0.625                 & {\ul 0.2}                          \\
\olmo (4 ICL)          & 0.5555              & 1.0              & 0.7142          & 0.75               & 9                           & 0.25          & 0.875                 & 0.2857                             \\
\bloom (3 ICL)         & 0.4444              & 0.8              & 0.5714          & 0.625              & 9                           & 0.375         & 0.875                 & 0.4285                             \\
\llamaTwo (3 ICL)      & 0.8333              & 1.0              & \textbf{0.9091} & 0.9375             & 6                           & 0.0625        & 0.6875                & \textbf{0.0909}                    \\
\llamaThree (3 ICL)    & 0.8333              & 1.0              & \textbf{0.9091} & 0.9375             & 6                           & 0.0625        & 0.6875                & \textbf{0.0909}                    \\
\gptThreeFive  (3 ICL) & 0.8333              & 1.0              & \textbf{0.9091} & 0.9375             & 6                           & 0.0625        & 0.6875                & \textbf{0.0909}                    \\
\gptFouro  (3 ICL)     & 0.8333              & 1.0              & \textbf{0.9091} & 0.9375             & 6                           & 0.0625        & 0.6875                & \textbf{0.0909}                    \\ \hline
\end{tabular}%
}
\caption{Causal discovery results for the Obesity causal graph, with 4 nodes and 5 edges. \llamaTwo, \llamaThree, \gptThreeFive and \gptFouro outperform all other models. The second-best performing method is \olmo (3 ICL).}
\label{tab:obesity_causal_discvoery}
\end{table*}

%% file: tables/conceptnet_causal_direction_identification.tex
\begin{table*}[h]
\centering
\begin{tabular}{lllll}
\hline
                       & \textbf{Precision↑} & \textbf{Recall↑} & \textbf{F1↑}    & \textbf{Accuracy↑} \\ \hline
\olmo (0 ICL)          & 0.5482              & \textbf{0.8831}  & 0.6765          & 0.5778             \\
\olmo (1 ICL)          & 0.5491              & 0.8184           & 0.6573          & 0.5734             \\
\olmo (2 ICL)          & 0.5771              & 0.7825           & 0.6643          & 0.6047             \\
\olmo (3 ICL)          & 0.6612              & 0.8427           & {\ul 0.7410}    & 0.7053             \\
\olmo (4 ICL)          & 0.5294              & {\ul 0.8721}     & 0.6589          & 0.5486             \\
\bloom (3 ICL)         & 0.5027              & 0.7248           & 0.5937          & 0.5041             \\
\llamaTwo (3 ICL)      & 0.6197              & 0.7774           & 0.6897          & 0.6503             \\
\llamaThree (3 ICL)    & {\ul 0.7659}        & 0.6575           & 0.7076          & {\ul 0.7282}       \\
\gptThreeFive  (3 ICL) & 0.6732              & 0.7308           & 0.7008          & 0.6891             \\
\gptFouro  (3 ICL)     & \textbf{0.8141}     & 0.8342           & \textbf{0.8240} & \textbf{0.8224}    \\ \hline
\end{tabular}%
\caption{Causal direction identification results on the \conceptnet dataset, with 1900 causal relations and 1900 reverse causal relations.  \gptFouro outperforms all competing methods, achieving an F1 score of 0.8240. The second-best performing method is \olmo (3 ICL), with an F1 score of 0.7410.}
\label{tab:conceptnet_causal_direction_identification}
\end{table*}

%% file: tables/causenet_causal_direction_identification.tex
\begin{table*}[h]
\centering
\begin{tabular}{lllll}
\hline
                       & \textbf{Precision↑} & \textbf{Recall↑} & \textbf{F1↑}    & \textbf{Accuracy↑} \\ \hline
\olmo (0 ICL)          & 0.5461              & \textbf{0.9657}  & 0.6977          & 0.5815             \\
\olmo (1 ICL)          & 0.5359              & {\ul 0.9606}     & 0.6881          & 0.5644             \\
\olmo (2 ICL)          & 0.5610              & 0.9091           & 0.6938          & 0.5988             \\
\olmo (3 ICL)          & 0.6568              & 0.8771           & 0.7511          & 0.7094             \\
\olmo (4 ICL)          & 0.5860              & 0.9410           & 0.7223          & 0.6382             \\
\bloom (3 ICL)         & 0.5067              & 0.6928           & 0.5853          & 0.5092             \\
\llamaTwo (3 ICL)      & 0.7030              & 0.8931           & 0.7867          & 0.7582             \\
\llamaThree (3 ICL)    & {\ul 0.8838}        & 0.8296           & 0.8558          & 0.8602             \\
\gptThreeFive  (3 ICL) & \textbf{0.8990}     & 0.8857           & {\ul 0.8923}    & {\ul 0.8931}       \\
\gptFouro  (3 ICL)     & 0.8596              & 0.9557           & \textbf{0.9051} & \textbf{0.8998}    \\ \hline
\end{tabular}%
\caption{Causal direction identification results on the \causenet dataset, with 814 causal relations and 814 reverse causal relations.  \gptFouro outperforms all competing methods, achieving an F1 score of 0.9051. The second-best performing method is \gptThreeFive, with an F1 score of 0.8923.}
\label{tab:causenet_causal_direction_identification}
\end{table*}

%% file: tables/total_occur_short.tex
\begin{table*}[ht]
\centering
\resizebox{\textwidth}{!}{%
\begin{tabular}{llcc}
\hline
                                                               &        & \multicolumn{1}{l}{Correct Causal Relations} & \multicolumn{1}{l}{Incorrect Causal Relations} \\ \hline
\multirow{2}{*}{Causal  Discovery (all datasets)}              & \dolma & 28812                                        & 1127                                           \\
                                                               & \roots & 814                                          & 118                                            \\ \hline
\multirow{2}{*}{Causal Direction Identification (\conceptnet)} & \dolma & 41407                                        & 3410                                           \\
                                                               & \roots & 1176                                         & 131                                            \\ \hline
\multirow{2}{*}{Causal Direction Identification (\causenet)}   & \dolma & 949427                                       & 107070                                         \\
                                                               & \roots & 24591                                        & 4236                                           \\ \hline
\end{tabular}%
}
\caption{Occurrences of correct and incorrect causal relations in the \dolma and \roots corpora. }
\label{tab:total_occur_short}
\end{table*}

%% file: tables/causal_discovery_model_different_sizes.tex
\begin{table*}[ht]
\centering
\resizebox{\textwidth}{!}{%
\begin{tabular}{lcccccccc}
\hline
\multicolumn{9}{c}{\textbf{Arctic Sea Ice}}                                                                                                                                                                                                                                                 \\ \hline
                      & \textbf{Precision↑} & \textbf{Recall↑} & \textbf{F1↑}    & \textbf{Accuracy↑} & \multicolumn{1}{l}{\textbf{Predict edges}} & \multicolumn{1}{l}{\textbf{NHD↓}} & \multicolumn{1}{l}{\textbf{Baseline NHD}} & \multicolumn{1}{l}{\textbf{Ratio (NHD/Baseline NHD)↓}} \\ \hline
\llamaTwo (3 ICL)     & 0.4444              & 0.5217           & 0.48            & 0.6388             & 54                                         & 0.3611                            & 0.6944                                    & 0.52                                                   \\
\llamaTwoX (3 ICL)    & 0.4478              & 0.6522           & 0.5309          & 0.6319             & 67                                         & 0.3681                            & 0.7847                                    & 0.4690                                                 \\
\llamaTwoXX (3 ICL)   & 0.3606              & 0.9565           & 0.5238          & 0.4444             & 122                                        & 0.5556                            & 1.0                                       & 0.5556                                                 \\
\llamaThree (3 ICL)   & 1.0                 & 0.1956           & 0.3272          & 0.7430             & 9                                          & 0.2569                            & 0.3819                                    & 0.6727                                                 \\
\llamaThreeXX (3 ICL) & 0.5689              & 0.7174           & \textbf{0.6346} & \textbf{0.7361}    & 58                                         & 0.2639                            & 0.7222                                    & \textbf{0.3653}                                        \\
\gptThreeFive (3 ICL) & 0.7647              & 0.2826           & 0.4126          & 0.7431             & 17                                         & 0.2569                            & 0.4375                                    & 0.5873                                                 \\
\gptFouro (3 ICL)     & 0.5178              & 0.6304           & 0.5686          & 0.6944             & 56                                         & 0.3055                            & 0.7083                                    & 0.4313                                                 \\ \hline
\multicolumn{9}{c}{\textbf{Insurance}}                                                                                                                                                                                                                                                      \\ \hline
\llamaTwo (3 ICL)     & 0.1245              & 0.7115           & 0.2120          & 0.6227             & 297                                        & 0.3772                            & 0.4787                                    & 0.7879                                                 \\
\llamaTwoX (3 ICL)    & 0.1338              & 0.7307           & 0.2262          & 0.6433             & 284                                        & 0.3566                            & 0.4609                                    & 0.7738                                                 \\
\llamaTwoXX (3 ICL)   & 0.1619              & 0.7692           & 0.2675          & 0.6995             & 247                                        & 0.3004                            & 0.4102                                    & 0.7324                                                 \\
\llamaThree (3 ICL)   & 0.2656              & 0.3269           & 0.2931          & 0.8875             & 64                                         & 0.1124                            & 0.1591                                    & 0.7069                                                 \\
\llamaThreeXX (3 ICL) & 0.2183              & 0.5961           & 0.3195          & 0.8189             & 142                                        & 0.1811                            & 0.2661                                    & 0.6804                                                 \\
\gptThreeFive (3 ICL) & 0.1575              & 0.5              & 0.2396          & 0.7736             & 165                                        & 0.2263                            & 0.2976                                    & 0.7603                                                 \\
\gptFouro (3 ICL)     & 0.2287              & 0.6730           & \textbf{0.3414} & 0.8148             & 153                                        & 0.1851                            & 0.2812                                    & \textbf{0.6585}                                        \\ \hline
\multicolumn{9}{c}{Alcohol}                                                                                                                                                                                                                                                                 \\ \hline
\llamaTwo (3 ICL)     & 0.75                & 1.0              & 0.8571          & 0.8889             & 4                                          & 0.1111                            & 0.7778                                    & 0.1429                                                 \\
\llamaTwoX (3 ICL)    & 0.75                & 1.0              & 0.8571          & 0.8889             & 4                                          & 0.1111                            & 0.7778                                    & 0.1429                                                 \\
\llamaTwoXX (3 ICL)   & 0.75                & 1.0              & 0.8571          & 0.8889             & 4                                          & 0.1111                            & 0.7778                                    & 0.1429                                                 \\
\llamaThree (3 ICL)   & 1.0                 & 1.0              & \textbf{1.0}    & 1.0                & 3                                          & 0                                 & 0.6667                                    & \textbf{0}                                             \\
\llamaThreeXX (3 ICL) & 1.0                 & 1.0              & \textbf{1.0}    & 1.0                & 3                                          & 0                                 & 0.6667                                    & \textbf{0}                                             \\
\gptThreeFive (3 ICL) & 1.0                 & 1.0              & \textbf{1.0}    & 1.0                & 3                                          & 0                                 & 0.6667                                    & \textbf{0}                                             \\
\gptFouro (3 ICL)     & 1.0                 & 1.0              & \textbf{1.0}    & 1.0                & 3                                          & 0                                 & 0.6667                                    & \textbf{0}                                             \\ \hline
\multicolumn{9}{c}{\textbf{Cancer}}                                                                                                                                                                                                                                                         \\ \hline
\llamaTwo (3 ICL)     & 0.4166              & 1.0              & 0.5882          & 0.5625             & 12                                         & 0.4375                            & 1.0                                       & 0.4375                                                 \\
\llamaTwoX (3 ICL)    & 0.5556              & 1.0              & 0.7143          & 0.75               & 9                                          & 0.25                              & 0.875                                     & 0.2857                                                 \\
\llamaTwoXX (3 ICL)   & 0.5556              & 1.0              & 0.7143          & 0.75               & 9                                          & 0.25                              & 0.875                                     & 0.2857                                                 \\
\llamaThree (3 ICL)   & 1.0                 & 0.8              & \textbf{0.8889} & 0.9375             & 4                                          & 0.0625                            & 0.5625                                    & \textbf{0.1111}                                        \\
\llamaThreeXX (3 ICL) & 1.0                 & 0.8              & \textbf{0.8889} & 0.9375             & 4                                          & 0.0625                            & 0.5625                                    & \textbf{0.1111}                                        \\
\gptThreeFive (3 ICL) & 1.0                 & 0.8              & \textbf{0.8889} & 0.9375             & 4                                          & 0.0625                            & 0.5625                                    & \textbf{0.1111}                                        \\
\gptFouro (3 ICL)     & 0.8                 & 0.8              & 0.8             & 0.875              & 5                                          & 0.125                             & 0.625                                     & 0.2                                                    \\ \hline
\multicolumn{9}{c}{\textbf{Diabetes}}                                                                                                                                                                                                                                                       \\ \hline
\llamaTwo (3 ICL)     & 0.5556              & 1.0              & 0.7142          & 0.75               & 9                                          & 0.25                              & 0.875                                     & 0.2857                                                 \\
\llamaTwoX (3 ICL)    & 0.625               & 1.0              & 0.7692          & 0.8125             & 8                                          & 0.1875                            & 0.8125                                    & 0.2307                                                 \\
\llamaTwoXX (3 ICL)   & 0.625               & 1.0              & 0.7692          & 0.8125             & 8                                          & 0.1875                            & 0.8125                                    & 0.2307                                                 \\
\llamaThree (3 ICL)   & 1.0                 & 0.8              & 0.8889          & 0.9375             & 4                                          & 0.0625                            & 0.5625                                    & 0.1111                                                 \\
\llamaThreeXX (3 ICL) & 1.0                 & 1.0              & \textbf{1.0}    & 1.0                & 5                                          & 0                                 & 0.625                                     & \textbf{0}                                             \\
\gptThreeFive (3 ICL) & 1.0                 & 1.0              & \textbf{1.0}    & 1.0                & 5                                          & 0                                 & 0.625                                     & \textbf{0}                                             \\
\gptFouro (3 ICL)     & 0.8333              & 1.0              & 0.9091          & 0.9375             & 6                                          & 0.0625                            & 0.6875                                    & 0.0909                                                 \\ \hline
\multicolumn{9}{c}{\textbf{Obesity}}                                                                                                                                                                                                                                                        \\ \hline
\llamaTwo (3 ICL)     & 0.8333              & 1.0              & \textbf{0.9091} & 0.9375             & 6                                          & 0.0625                            & 0.6875                                    & \textbf{0.0909}                                        \\
\llamaTwoX (3 ICL)    & 0.8333              & 1.0              & \textbf{0.9091} & 0.9375             & 6                                          & 0.0625                            & 0.6875                                    & \textbf{0.0909}                                        \\
\llamaTwoXX (3 ICL)   & 0.8333              & 1.0              & \textbf{0.9091} & 0.9375             & 6                                          & 0.0625                            & 0.6875                                    & \textbf{0.0909}                                        \\
\llamaThree (3 ICL)   & 0.8333              & 1.0              & \textbf{0.9091} & 0.9375             & 6                                          & 0.0625                            & 0.6875                                    & \textbf{0.0909}                                        \\
\llamaThreeXX (3 ICL) & 0.8333              & 1.0              & \textbf{0.9091} & 0.9375             & 6                                          & 0.0625                            & 0.6875                                    & \textbf{0.0909}                                        \\
\gptThreeFive (3 ICL) & 0.8333              & 1.0              & \textbf{0.9091} & 0.9375             & 6                                          & 0.0625                            & 0.6875                                    & \textbf{0.0909}                                        \\
\gptFouro (3 ICL)     & 0.8333              & 1.0              & \textbf{0.9091} & 0.9375             & 6                                          & 0.0625                            & 0.6875                                    & \textbf{0.0909}                                        \\ \hline
\end{tabular}%
}
\caption{Performance on causal discovery task using Llama2 and Llama3 models of different sizes.}
\label{tab:causal_discovery_model_different_sizes}
\end{table*}

%% file: tables/causal_direction_model_different_sizes.tex
\begin{table*}[ht]
\centering
\begin{tabular}{lcccc}
\hline
\multicolumn{5}{c}{\conceptnet}                                                                       \\ \hline
                      & \textbf{Precision↑} & \textbf{Recall↑} & \textbf{F1↑}    & \textbf{Accuracy↑} \\ \hline
\llamaTwo (3 ICL)     & 0.6197              & 0.7774           & 0.6897          & 0.6503             \\
\llamaTwoX (3 ICL)    & 0.6010              & 0.8605           & 0.7077          & 0.6647             \\
\llamaTwoXX (3 ICL)   & 0.6384              & 0.8742           & 0.7380          & 0.6897             \\
\llamaThree (3 ICL)   & 0.7659              & 0.6575           & 0.7076          & 0.7283             \\
\llamaThreeXX (3 ICL) & 0.8555              & 0.8253           & \textbf{0.8401} & \textbf{0.8430}    \\
\gptThreeFive (3 ICL) & 0.6732              & 0.7308           & 0.7008          & 0.6891             \\
\gptFouro (3 ICL)     & 0.8141              & 0.8342           & 0.8240          & 0.8224             \\ \hline
\multicolumn{5}{c}{\causenet}                                                                         \\ \hline
                      & \textbf{Precision↑} & \textbf{Recall↑} & \textbf{F1↑}    & \textbf{Accuracy↑} \\ \hline
\llamaTwo (3 ICL)     & 0.7030              & 0.8931           & 0.7867          & 0.7582             \\
\llamaTwoX (3 ICL)    & 0.6625              & 0.9213           & 0.7708          & 0.7260             \\
\llamaTwoXX (3 ICL)   & 0.7359              & 0.9521           & 0.8302          & 0.8053             \\
\llamaThree (3 ICL)   & 0.8838              & 0.8296           & 0.8558          & 0.8602             \\
\llamaThreeXX (3 ICL) & 0.8939              & 0.9423           & \textbf{0.9175} & \textbf{0.9152}    \\
\gptThreeFive (3 ICL) & 0.8990              & 0.8857           & 0.8923          & 0.8931             \\
\gptFouro (3 ICL)     & 0.8596              & 0.9557           & 0.9051          & 0.8998             \\ \hline
\end{tabular}%
\caption{Performance on causal direction identification task using Llama2 and Llama3 models of different sizes.}
\label{tab:causal_direction_model_different_sizes}
\end{table*}